\documentclass{article}

\usepackage[accepted]{icml2024}

\usepackage{graphicx}
\usepackage[backref=page]{hyperref}
\usepackage{xurl} %

\newcommand{\updownarrows}{\mathbin\uparrow\hspace{-.5em}\downarrow}

\usepackage{amsfonts}
\usepackage{amsmath}
\usepackage{amssymb}
\usepackage{mathtools}
\usepackage{amsthm}
\usepackage{mathrsfs}
\usepackage{bm}
\usepackage{dsfont}

\def\1{\bm{1}}

\def\eps{{\epsilon}}
\def\epsilon{{\varepsilon}}

\def\vone{{\bm{1}}}

\def\vomega{{\bm{\omega}}}
\def\vgamma{{\bm{\gamma}}}

\def\vb{{\bm{b}}}

\def\vd{{\bm{d}}}

\def\vg{{\bm{g}}}

\def\vm{{\bm{m}}}
\def\vn{{\bm{n}}}

\def\vp{{\bm{p}}}

\def\vu{{\bm{u}}}
\def\vv{{\bm{v}}}
\def\vw{{\bm{w}}}
\def\vx{{\bm{x}}}
\def\vy{{\bm{y}}}

\def\evomega{{\omega}}

\def\evd{{d}}

\def\evg{{g}}

\def\evn{{n}}

\def\evp{{p}}

\def\evu{{u}}
\def\evv{{v}}

\def\evx{{x}}

\def\mW{{\bm{W}}}
\def\mX{{\bm{X}}}

\def\mZ{{\bm{Z}}}

\DeclareMathAlphabet{\mathsfit}{\encodingdefault}{\sfdefault}{m}{sl}
\SetMathAlphabet{\mathsfit}{bold}{\encodingdefault}{\sfdefault}{bx}{n}

\newcommand{\E}{\mathbb{E}}
\newcommand{\Ls}{\mathscr{L}} %
\newcommand{\R}{\mathbb{R}}

\DeclareMathOperator{\sign}{sign}

\usepackage[utf8]{inputenc} %
\usepackage[T1]{fontenc}    %
\usepackage{booktabs}       %
\usepackage{nicefrac}       %
\usepackage{microtype}      %
\usepackage{xcolor}         %
\usepackage{multirow}       %
\usepackage[labelfont=bf]{caption}  %
\usepackage{cleveref}       %
\crefname{appendix}{appx.}{appx.}
\crefname{algorithm}{algo.}{algo.}
\crefformat{section}{\S#2#1#3}

\usepackage{tikz}
\usetikzlibrary{arrows.meta, angles, quotes}
\definecolor{tabblue}{RGB}{0,114,178} %
\definecolor{tabgreen}{RGB}{0,158,115} %
\definecolor{tabred}{RGB}{204,121,167} %
\definecolor{taborange}{RGB}{222,143,5} %
\usepackage{pgfplots}
\pgfplotsset{compat=1.18}

\newcommand{\semph}[1]{\textcolor{blue!50!black}{{#1}}}
\newcommand{\term}[1]{\textbf{\textit{#1}}}

\usepackage{wrapfig}
\usepackage{subcaption}
\usepackage{enumitem}
\usepackage{placeins}  %

\usepackage{algorithm}
\usepackage[commentColor=gray,beginComment=\#~,beginLComment=\#~,endLComment={}]{algpseudocodex}
\makeatletter
\renewcommand{\ALG@beginalgorithmic}{\small}
\makeatother
\algrenewcommand\alglinenumber[1]{\small #1:}

\allowdisplaybreaks

\newenvironment{talign}
 {\align}
 {\endalign}
\newenvironment{talign*}
 {\csname align*\endcsname}
 {\endalign}

\newcommand{\scaledwidehat}[1]{\scalebox{0.8}{\(\widehat{#1}\)}}
\newcommand{\inlinevomegaeqnorm}{\smash{\scaledwidehat{\|\vomega\|}}}

\icmltitlerunning{Rotational Equilibrium: How Weight Decay Balances Learning Across Neural Networks}

\begin{document}

\twocolumn[
\icmltitle{Rotational Equilibrium: How Weight Decay \texorpdfstring{\\}{} Balances Learning Across Neural Networks}

\icmlsetsymbol{equal}{*}

\begin{icmlauthorlist}
\icmlauthor{Atli Kosson}{EPFL,equal} %
\icmlauthor{Bettina Messmer}{EPFL,equal} %
\icmlauthor{Martin Jaggi}{EPFL}
\end{icmlauthorlist}

\icmlaffiliation{EPFL}{EPFL, Switzerland}
\icmlcorrespondingauthor{}{atli.kosson@epfl.ch}

\icmlkeywords{
    Machine Learning,
    ICML
}

\vskip 0.3in
]

\printAffiliationsAndNotice{\icmlEqualContribution}

\begin{abstract}
This study investigates how weight decay affects the update behavior of individual neurons in deep neural networks through a combination of applied analysis and experimentation.
Weight decay can cause the expected magnitude and angular updates of a neuron's weight vector to converge to a steady state we call \textit{rotational equilibrium}.
These states can be highly homogeneous, effectively balancing the average rotation---a proxy for the effective learning rate---across different layers and neurons.
Our work analyzes these dynamics across optimizers like Adam, Lion, and SGD with momentum, offering a new simple perspective on training that elucidates the efficacy of widely used but poorly understood methods in deep learning.
We demonstrate how balanced rotation plays a key role in the effectiveness of normalization like Weight Standardization, as well as that of AdamW over Adam with $\ell_2$-regularization.
Finally, we show that explicitly controlling the rotation provides the benefits of weight decay while substantially reducing the need for learning rate warmup.

\end{abstract}
\vspace{-10pt}
\section{Introduction}

The use of weight decay or $\ell_2$-regularization has become ubiquitous in deep learning optimization.
Although originally proposed as an explicit regularization method, \citet{van2017l2} showed that this interpretation does not hold for modern networks with normalization layers.
This is because normalization can make a weight vector scale-invariant, meaning that the network output is unaffected by the magnitude of the vector (see \cref{sec:preliminaries}).
For scale-invariant vectors, weight decay instead acts as a scaling factor for some notation of an ``effective'' learning rate with varying definitions \citep{van2017l2,zhang2018three,li2020exponential,wan2021spherical}.
We explore this idea further, aiming to describe the effects of weight decay on the optimization dynamics of modern neural network (NN) training through applied analysis and experimentation.

\begin{figure}
\resizebox{0.7\width}{!}{
\pgfplotsset{width=2.6in}
\begin{tikzpicture}
\begin{axis}[
    axis lines = left,
    xlabel = $t$,
    ylabel = {$\E[\|\vomega_t\|]$},
    ymin=0.0, ymax=1,
    xticklabels={}, yticklabels={},
    tick style={draw=none},
    label style={font=\large}, %
    ylabel shift = -10 pt,
    xlabel shift = -10 pt,
    line width = 1.25pt,
]
\addplot [
    domain=0:17, 
    samples=100, 
    color=tabred,
    line width=1.25pt %
]
{0.3+0.5*(exp(-0.5*(x+1.5))-exp(-0.5*1.5))};
\addplot [
    domain=0:17, 
    samples=100, 
    color=taborange,
    line width=1.25pt %
]
{0.3+0.5*(exp(-0.5*(x+3))-exp(-0.5*3))};
\addplot [
    domain=0:17, 
    samples=100, 
    color=tabblue,
    line width=1.25pt %
]
{0.3+0.5*(1-exp(-0.5*x))};
\addplot [
    domain=0:17, 
    samples=100, 
    color=tabgreen,
    line width=1.25pt %
]
{0.3+0.5*(exp(-0.5*5)-exp(-0.5*(x+5)))};

\end{axis}
\end{tikzpicture}
\pgfplotsset{width=2.6in}
\begin{tikzpicture}
\begin{axis}[
    axis lines = left,
    xlabel = $t$,
    ylabel = {$\E[\angle (\vomega_t, \vomega_{t+1})]$},
    ymin=0.0, ymax=1,
    xticklabels={}, yticklabels={},
    tick style={draw=none},
    label style={font=\large}, %
    ylabel shift = -10 pt,
    xlabel shift = -10 pt,
    line width = 1.25pt,
]
\addplot [
    domain=0:17, 
    samples=100, 
    color=tabred,
    line width=1.25pt %
]
{0.5-exp(-0.5*(x+1.5))};
\addplot [
    domain=0:17, 
    samples=100, 
    color=taborange,
    line width=1.25pt %
]
{0.5-exp(-0.5*(x+3))};
\addplot [
    domain=0:17, 
    samples=100, 
    color=tabblue,
    line width=1.25pt %
]
{exp(-0.5*x)+0.5};
\addplot [
    domain=0:17, 
    samples=100, 
    color=tabgreen,
    line width=1.25pt %
]
{exp(-0.5*(x+5))+0.5};
\addplot [
    dashed,
    domain=0:17, 
    samples=100, 
    color=black,
    line width=1.25pt %
]
{0.5};
\node[above right, font=\normalsize] at (axis cs:1.25,0) {transient phase};

\node[above, font=\normalsize] at (axis cs:13,0) {equilibrium};

\node[above, font=\normalsize] at (axis cs:12.9,0.5) {balanced rotation};

\node[above, color=tabblue, font=\normalsize] (fast) at (axis cs:8,0.9) {fast rotation};
\draw[-{Latex[length=1mm]}, color=tabblue, line width=0.4pt] (fast) .. controls (axis cs:3.5,0.95) and (axis cs:3,0.92) .. (axis cs:2.5,0.9);

\node[below, color=tabred, font=\normalsize] (slow) at (axis cs:7,0.25) {slow rotation};
\draw[-{Latex[length=1mm]}, color=tabred, line width=0.4pt, bend right] (slow) .. controls (axis cs:3.5,0.20) and (axis cs:3.0,0.20) .. (axis cs:2.0,0.25);

\end{axis}
\end{tikzpicture}
}
\caption{Conceptual figure of the norm (left) and angular updates (right) of the weight vector $\vomega_t$ for different neurons (each line color) over time $t$ with a constant learning rate. Weight decay modulates and stabilizes both metrics.}
\label{fig:intro-conceptual}
\vspace{-10pt}
\end{figure}
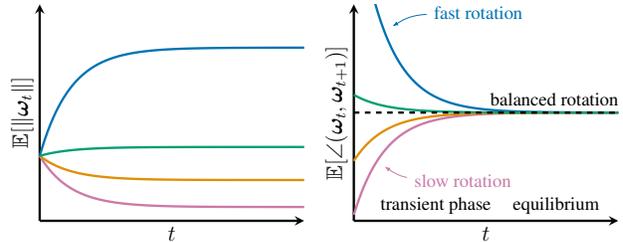

We specifically examine the \term{update dynamics} of individual neurons. A \term{neuron} computes a scalar feature by comparing a learnable weight vector $\vomega_t \in \mathbb{R}^C$ with an incoming activation vector $\vx_t \in \mathbb{R}^C$, followed by a non-linear activation function $\varphi$ and an optional learnable bias $b_t \in \mathbb{R}$:
\begin{equation}\label{eq:neuron}
    \varphi(\langle \vomega_t, \vx_t \rangle + b_t) = \varphi(\|\vomega_t\| \|\vx_t\| \cos(\angle(\vomega_t, \vx_t)) + b_t)
\end{equation}
Here $\langle \cdot \rangle$ denotes a dot product which can be rewritten with a cosine similarity, showing that the direction of $\vomega_t$ determines which ``patterns'' in $\vx_t$ the neuron responds to.
We describe the neuronal update dynamics through the \term{expected weight norm} $\E[\|\vomega_t\|]$, \term{root-mean-square (RMS) update size} $\eta_g$ for the bias, and \term{expected angular update size} $\eta_r$:
\begin{talign}
    \eta_g &:= \sqrt{\E[(b_{t+1}-b_t)^2]} \\
    \eta_r &:= \E[\angle(\vomega_{t},\vomega_{t+1})]=\E\left[\arccos\left(\frac{\langle \vomega_{t}, \vomega_{t+1}\rangle}{\|\vomega_{t}\| \|\vomega_{t+1}\|}\right)\right] 
\end{talign}
where the expectation accounts for noise from randomly sampled minibatches (e.g., data shuffling).
We assume weight decay is only applied to $\vomega$, not $b$ or other parameters, which is common and good practice~\cite{jia2018highly}.

\Cref{fig:intro-conceptual} shows how the weight norm and angular updates ($\eta_r$) of different neurons could behave over time in a typical case.
This behavior is caused by Spherical Motion Dynamics~\citep{wan2021spherical} which arise from the interaction of weight decay and stochastic gradient updates (described further in \cref{sec:analysis}).
Over time, the weight norm reaches a stable \term{equilibrium magnitude} in expectation, at which point the opposing effects of gradient updates (which increase the norm) and weight decay (which reduces it) cancel each other out.
Interestingly, this simply stems from the geometry of the stochastic optimizer updates (especially with normalization layers), not the convergence of the underlying loss function, and can therefore be analyzed for a random walk.

The angular updates shown in \cref{fig:intro-conceptual}R have an inverse dependency on the weight magnitude.
During an initial \term{transient phase} the rotation is somewhat arbitrary, depending on the initial weight magnitude and gradient norm (for some optimizers).
The convergence of the weight norm results in a \term{steady-state} we call \term{rotational equilibrium}, characterized by the average angular update having a specific, stable magnitude.
For some setups the rotational equilibrium is identical for different layers and neurons (even if the weight norms differ), resulting in \term{balanced rotation} which we empirically observe aids optimization.
We discuss our intuition for why this helps in \cref{appx:why_balanced_rotation}.

This study significantly expands upon prior investigations into the interactions between weight decay and normalization, demystifying the effects of weight decay and other common tricks in deep learning.
While we touch upon certain previous works throughout, please refer to \cref{appx:related_work} for an extended discussion.
Earlier research has primarily focused on the general mechanisms and properties of weight decay and normalization, especially for plain SGD.
We focus on two main new directions, rotational equilibrium in other optimizers like AdamW~\citep{loshchilov2018decoupled}, and the importance of balanced rotation in the optimization of neural networks.
Compared to prior work, our analysis also targets more fine-grained update dynamics at the neuron level, applies to networks without scale-invariance from perfectly placed normalization layers, investigates additional aspects of the dynamics (e.g.\ transient phase, bias behavior), and relates standard optimizers to the LARS-style optimizers~\cite{you2017lars}.
Our key contributions are:
\begin{itemize}[leftmargin=15pt,itemsep=0pt,topsep=0pt]
    \item Deriving the steady-state neuronal update dynamics of AdamW, Adam with $\ell_2$-regularization, Lion and SGD with momentum for a random walk. We experimentally validate that the results hold for NN training in practice.
    \item Showing how the interaction of weight decay and learning rate shapes the rotation in the initial transient phase and results in two distinct ``effective step sizes'' in the steady state, $\eta_g$ for biases and $\eta_r$ for weights.
    \item Demonstrating how explicitly controlling the angular updates via Rotational Optimizer Variants provides an alternative way of achieving the benefits of weight decay and normalization, while also simplifying the update dynamics and reducing the need for learning rate warmup.
    \item Revealing how balanced rotation contributes to the performance benefit of AdamW vs Adam+$\ell_2$, and certain normalization layers like Weight Standardization.
\end{itemize}

\section{Preliminaries}\label{sec:preliminaries}
\paragraph{Scale-Invariance:} \citet{li2020exponential,wan2021spherical} describe how properly placed normalization can make the weight vector $\vomega$ of a neuron/layer scale invariant w.r.t.\ a loss $\Ls$ and the resulting properties of the gradient $\nabla_\vomega \Ls(\vomega,\cdot)$:
\begin{talign}
    \text{Scale Invariance:~} & \Ls(r\vomega,\cdot)=\Ls(\vomega,\cdot), \forall r\!>\!0 \\
    \text{Gradient orthogonality:~} & \nabla_\vomega \Ls(\vomega,\cdot) \perp \vomega \label{eq:grad_perp} \\
    \text{Inverse proportionality:~} & \|\nabla_\vomega \Ls(\vomega,\cdot)\| \propto \|\vomega\|^{-1} \label{eq:grad_inv}
\end{talign}
See \cref{appx:norm_scale_invariance} for an overview.
\semph{Note that different normalization operations can result in scale-invariance at a different granularity}, for example individual neurons for Batch Normalization~\cite{ioffe2015batch} and Weight Standardization~\cite{qiao2019ws,huang2017centered} but only whole layers for Layer Normalization~\cite{ba2016layer}.

\paragraph{The Effective Learning Rate:} In related literature there are multiple definitions of ``effective'' learning rates \citep{van2017l2,chiley2019online,wan2021spherical} which aim to describe how fast the neural network is being updated based on some metric.
We use the average angular change $\eta_r$ for this purpose, but will refer to it as an \term{effective update size}.
\semph{Note that the direction of $\vomega$ in \cref{eq:neuron} controls which patterns in the inputs $\vx$ the neuron detects, and $\eta_r$ thus directly captures how fast this important aspect in the underlying functional representation of the neuron changes.}
This applies to all neurons, but particularly when the weight vector is scale-invariant and the direction thus fully determines its effect.
Simple alternatives like $\|\vomega_{t+1}-\vomega_t\|$ are scale dependent and do not directly measure changes in the encoded function.
For other parameters we use the RMS change $\eta_g$ as a measure of the update size.
This is generally not a ``functional'' update measure (which would vary based on the architecture), but still informative and easy to analyze.
Finally we note that an update size only measures the size of individual updates.
Momentum affects the correlation of the updates over time, which also influences long-term ``learning speed'', see \cref{appx:update_size_vs_lr}.

\section{Analysis}\label{sec:analysis}
In this section we analyze the rotational equilibrium of a weight vector $\vomega$ to obtain simple expressions for the equilibrium magnitude~$\inlinevomegaeqnorm$ and the expected angular update in equilibrium,~$\widehat{\eta_r}$.
We focus on a simplified setting where updates are dominated by noise, resulting in a type of random walk.
The equilibrium dynamics derived in this analytically tractable setting are predictive of the behavior we empirically observe in neural networks (see experiments), but this is not meant to be a formal theoretical analysis that fully captures all the intricacies of neural network optimization.

Specifically, we assume the loss is in the form of empirical risk, i.e. ${\Ls(\vomega,\mathbb{X})=\frac{1}{|\mathbb{X}|}\sum_{\vx \in \mathbb{X}} \Ls(\vomega,\vx)}$ where $\mathbb{X}$ is our training dataset, $\vomega$ are our weights and $\vx$ is a data point.
The true noiseless gradient is then ${\vg_\mathbb{X}=\nabla_\vomega \Ls(\vomega,\mathbb{X})}$ and the gradient for a minibatch $\mathbb{B}$ is ${\vg_\mathbb{B}=\nabla_\vomega \Ls(\vomega,\mathbb{B})}$.
We can define the noise in the gradient as ${\vg_N = \vg_\mathbb{B}-\vg_\mathbb{X}}$ with $\E_\mathbb{B}[\vg_N]=0$, because $\E_\mathbb{B}[\vg_\mathbb{B}]=\vg_\mathbb{X}$ for a randomly sampled $\mathbb{B}$.
Our simplifying assumption of the noise dominating can be stated ${\vg_\mathbb{B} = \vg_\mathbb{X} + \vg_N \approx \vg_N}$, resulting in a \textit{random walk} for the neural network parameters.
Analogous assumptions have been successfully used elsewhere, e.g.\ with stochastic differential equations to derive the batch size scaling behavior of optimizers \citep{li2021validity,malladi2022on}.
In our experiments we find that the final predictions hold very well for a variety of networks despite being derived for this simplified setting.
\Cref{appx:real_vs_random_walk} further describes the analytical setting and explores how the differences between a random walk and real neural network optimization affect the predictions.

\subsection{Geometric Model for Equilibrium}\label{sec:equilibrium_geometric_model}

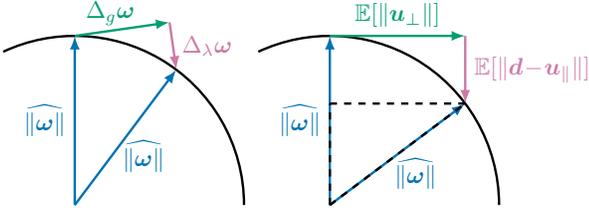
\begin{figure}
\resizebox{0.9\width}{!}{
\begin{tikzpicture}[
  arrow_blue/.style={tabblue, -{Latex[length=2mm]}, line width=1.0pt, opacity=1.0},
  arrow_green/.style={tabgreen, -{Latex[length=2mm]}, line width=1.0pt, opacity=1.0},
  arrow_red/.style={tabred, -{Latex[length=2mm]}, line width=1.0pt, opacity=1.0},
  triangle/.style={black, line width=1.0pt, dashed, opacity=1.0},
  scale=0.5, transform shape, every node/.style={scale=2.0}
]
  \draw[line width=1.0pt] (5,0cm) arc (0:115:5cm);

  \draw[arrow_blue] (0,0) -- (0,5cm) node[midway,left] {$\widehat{\|\vomega\|}$};
  \draw[arrow_blue] (0,0) -- (3,4cm) node[midway,right,xshift=-5pt,yshift=-8pt] {$\widehat{\|\vomega\|}$};

  \draw[arrow_green] (0,5cm) -- (2.8,5.4cm) node[midway,above,xshift=-5pt,yshift=-2pt] {$\Delta_g \vomega$};

  \draw[arrow_red] (2.8,5.4cm) -- (3,4cm) node[midway,right] {$\Delta_\lambda \vomega$};
\end{tikzpicture}
\begin{tikzpicture}[
  arrow_blue/.style={tabblue, -{Latex[length=2mm]}, line width=1.0pt, opacity=1.0},
  arrow_green/.style={tabgreen, -{Latex[length=2mm]}, line width=1.0pt, opacity=1.0},
  arrow_red/.style={tabred, -{Latex[length=2mm]}, line width=1.0pt, opacity=1.0},
  triangle/.style={black, line width=1.0pt, dashed, opacity=1.0},
  scale=0.5, transform shape, every node/.style={scale=2.0}
]
  \draw[line width=1.0pt] (5,0cm) arc (0:115:5cm);

  \draw[arrow_blue] (0,0) -- (0,5cm) node[midway,left] {$\widehat{\|\vomega\|}$};
  \draw[arrow_blue] (0,0) -- (4,3cm) node[midway,right,xshift=-5pt,yshift=-8pt] {$\widehat{\|\vomega\|}$};

  \draw[arrow_green] (0,5cm) -- ++(0:4cm) node[midway,above] {$\E[\|\vu_\perp\|]$};

  \draw[arrow_red] (90:5cm)+(0:4cm) -- (4,3cm) node[midway,right] {$\E[\|\vd\!-\!\vu_\parallel\|]$};

  \draw[triangle] (0,3cm) -- (4,3cm);
  \draw[triangle] (0,0cm) -- (4,3cm);
  \draw[triangle] (0,0cm) -- (0,3cm);

\end{tikzpicture}
}
\caption{
Two views of equilibrium where the weight norm $\inlinevomegaeqnorm$ is preserved because the gradient and weight decay components balance out on average. \textbf{Left:} Standard optimizer update \cref{eq:abstract_update}. \textbf{Right:} The total update contributions over the course of training, $\vu$ and $\vd$, derived from the gradient and weight decay of a given timestep, respectfully.
}
\label{fig:smd_components}
\end{figure}

\begin{table*}[tb]
\centering
\caption{
Analytical predictions for the steady state neuronal update dynamics of different optimizers.
The $\widehat{\eta_g}$ values apply to any parameter $\vp \in \R^{C}$ with gradient $\vg$.
The $\widehat{\eta_r}$ and $\inlinevomegaeqnorm$ values apply to scale-invariant weights $\vomega \in \R^{C}$ in equilibrium.
Expressions with $\tilde{\vg}:=\|\vp\|\vg$ use the inverse proportionality from \cref{eq:grad_inv} and would therefore differ without scale-invariance.
}
\label{tab:equilibrium_summary}
\begin{tabular}{r|cccc}
\toprule
& SGDM \eqref{eq:sgdm_update} & AdamW \eqref{eq:adamw_update} & Adam+$\ell_2$ \eqref{eq:adam_update} & Lion \eqref{eq:lion_update} \\
\midrule
RMS update size $\widehat{\eta_g}$ & $\eta \sqrt{\frac{\E[\|\vg\|^2]}{1-\alpha^2}}$ & $\eta \sqrt{C\frac{1-\beta_1}{1+\beta_1}}$ & $\eta \sqrt{C\frac{1-\beta_1}{1+\beta_1}}$ & $\eta \sqrt{C}$ \\
Expected rotation $\widehat{\eta_r}$ & $\sqrt{\frac{2 \eta \lambda}{1 + \alpha}}$ & $\sqrt{2 \eta \lambda \frac{1-\beta_1}{1 + \beta_1}}$ & $\sqrt[3]{\frac{2 \eta^2 \lambda}{\langle \vone, \sqrt{\E[\tilde{\vg}^2]} \rangle}} \sqrt{\frac{1-\beta_1}{1+\beta_1}C}$ & $\sqrt{\pi \eta \lambda} \left((1\!-\!\beta_1)^2 + \beta_1^2 \frac{1-\beta_2}{1+\beta_2}\right)^{\frac{1}{2}}$ \\
Equilibrium norm $\widehat{\|\vomega\|}$ & $\sqrt[4]{\frac{\eta \E[\|\tilde{\vg}\|^2]}{2\lambda \cdot (1-\alpha)}}$ & $\sqrt{\frac{\eta C}{2\lambda}}$ & $\sqrt[3]{\frac{\eta}{2 \lambda}\!\cdot\! 
    \langle \vone, \sqrt{\E[\tilde{\vg}^2]} \rangle}$ & $\sqrt{\frac{\eta C}{\pi \lambda}} \left((1\!-\!\beta_1)^2 + \beta_1^2 \frac{1-\beta_2}{1+\beta_2}\right)^{-\frac{1}{2}}$ \\
\bottomrule
\end{tabular}
\end{table*}

In this section we present a simple geometric derivation of the equilibrium norm $\inlinevomegaeqnorm$ for different optimizers inspired in part by the analysis in Online Normalization~\citep{chiley2019online}.
We divide a parameter update $\vomega_{t} \rightarrow \vomega_{t+1}$ into:
\begin{equation}\label{eq:abstract_update}
    \vomega_{t+1}-\vomega_t=\Delta_g \vomega + \Delta_\lambda \vomega
\end{equation}
where $\Delta_g \vomega$ comes from the gradients and $\Delta_\lambda \vomega$ from the weight decay.
Equilibrium is an abstract state where the effects of $\Delta_g \vomega$ and $\Delta_\lambda \vomega$ on the expected weight magnitude balance out on average.
\semph{These components typically have different monotonic dependencies on the weight magnitude, with weight decay being proportional while the gradient component is either constant or inversely proportional, depending on the setting.
As a result, the effects of these components can balance out in expectation at the equilibrium magnitude}~$\inlinevomegaeqnorm$.
As shown in \cref{fig:smd_components}L, the geometry of this is not necessarily simple.
Due to the averaging effects of momentum over time, $\Delta_g \vomega$ is not necessarily orthogonal to the weights even in cases where individual gradients are (e.g.\ for scale-invariant weights).
Similarly, the weight decay (or $\ell_2$-regularization) component $\Delta_\lambda \vomega$ may not be perfectly anti-parallel to the weights with momentum.

\vspace{10pt}
To simplify the effects of momentum, we instead consider a different view of equilibrium shown in \cref{fig:smd_components}R.
Here we consider the total weight change throughout training arising from the weight decay term and gradient from a given time step, instead of the update that is applied in that iteration.
The \term{Total Update Contribution (TUC)} of the gradient at time step~$t$, denoted $\vu$, is the sum of the contributions of $\nabla_\vomega \Ls(\vomega_t,\cdot)$ to subsequent updates ${\vomega_t \rightarrow \vomega_{t+1}}$, ${\vomega_{t+1} \rightarrow \vomega_{t+2}}$, and so on.
Analogously, the TUC of the weight decay, denoted $\vd$, is the total change due to the weight decay or $\ell_2$-regularization of the weights $\vomega_t$ at iteration~$t$.
Note that without momentum $\vu=\Delta_g \vomega$, $\vd=\Delta_\lambda \vomega$ and that if $\Delta_g \vomega$ and $\Delta_\lambda \vomega$ balance out on average, then so must $\vu$ and $\vd$.

In many cases $\vu$ is orthogonal to the weights on average due to scale-invariance or randomness.
Otherwise, we can split it into an orthogonal $\vu_\perp$ and a radial (outwards) $\vu_\parallel$ component.
The $\vu_\parallel$ term then has a similar effect as the weight decay term $\vd$ which is anti-parallel to the weights in the cases we consider. %
If we can obtain an expression for the orthogonal $\|\vu_\perp\|$ and radial ${\|\vd-\vu_\parallel\|}$ total update contributions, we can apply the Pythagorean theorem to the dashed triangle in \cref{fig:smd_components}R:
\vspace{-4pt}
\begin{equation} \label{eq:smd_pythagorean}
     (\widehat{\|\vomega\|}-\E[\|\vd-\vu_\parallel\|])^2 + \E[\|\vu_\perp\|]^2 = \smash{\widehat{\|\vomega\|}}^2
\end{equation}
We can then solve for $\inlinevomegaeqnorm$, making sure to account for the dependency of $\vu$ and $\vd$ on the weight norm.

Once we have an expression for $\inlinevomegaeqnorm$, we can compute a prediction for the corresponding RMS update size ${\widehat{\eta_g}=\sqrt{E[\|\Delta_g \vomega\|^2]}}$.
This gives a prediction for the expected relative update size $\widehat{\eta_g} / \inlinevomegaeqnorm := \widehat{\eta_r}$, which closely approximates the angular update $\eta_r$ in equilibrium.
We do this for AdamW in the next subsection and for SGDM, Lion~\citep{chen2023symbolic} and Adam with $\ell_2$-regularization in \cref{appx:sgdm_equilibrium}, \ref{appx:lion_equilibrium} and \ref{appx:adam_l2}.
The resulting predictions for the update dynamics of each optimizer are summarized in \cref{tab:equilibrium_summary}.

\subsection{AdamW Equilibrium}\label{sec:adamw_equilibrium}
We write AdamW~\citep{loshchilov2018decoupled} updates as:
\begin{talign}
    \vm_t &= \beta_1 \vm_{t-1} + (1 - \beta_1) \vg_t \\
    \vv_t &= \beta_2 \vv_{t-1} + (1 - \beta_2) \vg_t^2 \\
    \vp_t &= \vp_{t-1} - \eta \cdot \Big( \frac{\vm_t/(1-\beta_1^t)}{\sqrt{\vv_t/(1-\beta_2^t)} + \epsilon} + \lambda \vp_{t-1} \Big) \label{eq:adamw_update}
\end{talign}
where $\vp_t \in \R^{C}$ is a parameter vector at iteration $t$, ${\vg_t=\nabla_{\vp} \Ls(\vp_t,\ldots)}$ is the gradient, $\vm$ is the first moment and $\vv$ is the second moment.
The learning rate $\eta \ge 0$, weight decay $\lambda \ge 0$ (zero expect for weight vectors), moment coefficients $\beta_1,\beta_2 \in (0, 1)$ and $\epsilon \ge 0$ are hyperparameters.
For simplicity we assume that $\epsilon$ and the bias correction can be ignored, i.e.\ that $\epsilon$, $\beta_1^t$ and $\beta_2^t$ are all $\approx 0$.

\textbf{Equilibrium Magnitude}: When applying AdamW to a weight vector $\vomega \in \R^{C}$, the total update contributions are:
\begin{equation}
    \textstyle
    \vu = - \eta  \sum_{k=t}^{\infty} \beta_1^{k-t} (1 - \beta_1) \frac{\vg_t}{\sqrt{\vv_k}}, \quad \vd = -\eta\lambda\vomega
\end{equation}
We note that due to symmetry, each coordinate of $\vu$ has a zero-mean distribution in the random walk setup.
Since $\vu$ is independent from $\vomega$, this makes them orthogonal in expectation i.e.\ $\E[\langle \vu, \vomega \rangle]=0$.
It is also reasonable to assume that the variance of each coordinate remains constant when the gradient distribution is not changing over time, resulting in $\forall t,k: \E[\|\vg_t / \sqrt{\vv_k}\|^2]=C$ (the vector dimension) and therefore $\E[\|\vu\|^2]=\eta^2 C$.
Defining $\omega=\|\vomega\|$, $u=\|\vu\|$, $u_\parallel=\langle \vomega, \vu \rangle/\|\vomega\|$, $u_\perp^2=u^2-u_\parallel^2$ and $d=\|\vd\|$ we can write a recurrence relation based on \cref{eq:smd_pythagorean}:
\begin{align}
    \E[\omega_{i+1}^2] &= \E[(\omega_{i}-d+u_\parallel)^2 + u_\perp^2] \\
    &= \E[\omega_{i}^2-2d\omega_i+2u_\parallel \omega_i -2du_\parallel \nonumber \\
    & \quad +d^2+u_\parallel^2+(u^2-u_\parallel^2)] \\
    &= \E[\omega_i^2](1-2\eta\lambda+\eta^2\lambda^2) + \eta^2 C
\end{align}
where we used independence, $\E[u_\parallel]=0$, and ${\E[u]=\eta^2 C}$. The solution is:
\begin{talign}
    \textstyle \E[\omega_i^2] &= \E[\omega_0^2] a^i + \frac{\eta^2 C}{2\eta\lambda-\eta^2\lambda^2} (1 - a^i) \label{eq:adamw_norm_convergence}\\
    a & = 1-2\eta\lambda+\eta^2\lambda^2
\end{talign}
The recurrence relation is written in terms of the $\vu$ and $\vd$ instead of $\Delta_g \vomega$ and $\Delta_\lambda \vomega$.
This is thus only an approximation of how the real system converges to equilibrium over time, but still informative.
It may be a good approximation if $\|\vomega\|$ changes slowly compared to how fast $\vu$ is applied and $\vv$ is updated, i.e.\ when $\beta_1, \beta_2$ are low compared to $a$.
The limit ${i \rightarrow \infty}$ gives us the equilibrium norm listed in \cref{tab:equilibrium_summary}:
\begin{equation}\label{eq:adamw_equilibrium_norm}
    \textstyle
    \smash{\widehat{\|\vomega\|}} = \sqrt{\frac{\eta C}{2\lambda - \eta\lambda^2}} \approx \sqrt{\frac{\eta C}{2\lambda}}\qquad \text{(for }\lambda\eta \ll 2\text{)}
\end{equation}
\textbf{Expected Update Size:} We can estimate the RMS update size $\eta_g$ of a parameter $\vp \in \R^{C}$ with $\Delta_g \vp = \frac{\eta\vm_t}{\sqrt{\vv_t}}$ as follows:
\begin{talign}
     & \E[\|\Delta_g \vp\|^2]  = \E[\| 
\frac{\eta}{\sqrt{\vv_t}} (1-\beta_1) \sum_{k=0}^{t-1} \beta_1^{t-k} \vg_{t-k} \|^2] \label{eq:rms_update1}\\
    &\quad = \eta^2 (1-\beta_1)^2 \sum_{k=0}^{t-1} \beta_1^{2t-2k} \E[\|\frac{\vg_{t-k}}{\sqrt{\vv_t}}\|^2] \label{eq:rms_update2}\\
    &\quad \approx \eta^2 \frac{1-\beta_1}{1+\beta_1} C \label{eq:rms_update3}
\end{talign}
where we have approximated the geometric sum with its limit $t \rightarrow \infty$, used the fact that for the random walk $\forall j \ne k: \E[\langle \vg_j,\vg_k \rangle]=0$ as well as our previous assumption ${\forall t,k: \E[\|\vg_t / \sqrt{\vv_k}\|^2]=C}$.
This gives us the prediction $\widehat{\eta_g} \approx \E[\sqrt{\|\Delta_g \vp\|^2}]$ listed in \cref{tab:equilibrium_summary}.
Approximating the equilibrium angular update size with the expected relative update size $\sqrt{\E[\|\Delta_g \vomega \|^2]}/\smash{\widehat{\|\vomega\|}}$ gives the $\widehat{\eta_r}$ value.
This approximation is good for small relative updates and a relatively small radial component in $\Delta_g \vomega$.

\subsection{AdamW vs Adam with \texorpdfstring{$\ell_2$}{L2}-Regularization}\label{sec:analysis_adam_vs_adamw}
\citet{loshchilov2018decoupled} proposed the use of decoupled weight decay instead of $\ell_2$-regularization in Adam~\citep{kingma15adam}.
In their experiments they find that Adam with decoupled weight decay (i.e.\ AdamW, see \cref{eq:adamw_update}) outperforms the $\ell_2$-regularized form (i.e.\ Adam+$\ell_2$, \cref{eq:adam_update}) across a wide range of settings.
Since then AdamW has been widely adopted, but as far as we know the reason for its effectiveness over Adam+$\ell_2$ is not well understood.

Our analysis of Adam+$\ell_2$ in \cref{appx:adam_l2} reveals that both the equilibrium norm and angular update size depend on the gradient magnitude (through $\tilde{\vg}$) unlike AdamW, see \cref{tab:equilibrium_summary}.
When the gradient norm varies between neurons or layers, this results in imbalanced rotation.
\semph{We believe the balanced vs imbalanced equilibrium rotation is a key difference between AdamW and Adam+$\ell_2$, which may explain why decoupled weight decay is more effective for Adam-like methods.}
In our experiments (\S\ref{sec:experiments_balanced_rotation}) we explore this further along with the general impact of imbalanced rotation.

\subsection{Rotational Dynamics of Scale-Sensitive Parameters}\label{sec:scale_sensitive_dynamics}
Prior work has primarily focused on the dynamics of scale-invariant weights.
Note that any weight vector can be made scale-invariant by simply applying normalization to it, e.g.\ Weight Standardization~\citep{qiao2019ws}.
For a random walk the gradient component is always orthogonal in expectation, but for real tasks scale-sensitive weights can have an average radial gradient component $\E[\vu_\parallel]\!\ne\!0$ (\cref{fig:smd_components}R).
In \cref{appx:scale_sensitive_smd} we explore how this affects the rotational dynamics of these weights (for SGDM).
\semph{We find that a radial gradient component acts like an additional weight decay $\lambda_u = -\E[\vu_\parallel]/(\eta \|\vomega\|)$ giving a new ``effective'' weight decay of $\lambda_e=\lambda + \lambda_u$, resulting in update dynamics similar to scale-invariant weights with the adjusted value.}

\renewcommand{\algorithmicthen}{\textbf{:}}
\renewcommand{\algorithmicdo}{\textbf{:}}
\renewcommand{\algorithmicelse}{\textbf{else:}}
\algnewcommand\LeftComment[2]{
\hspace{#1\algindent}$\triangleright$ \eqparbox{COMMENT}{#2} \hfill %
}
\begin{algorithm}[tb]
\caption{Rotational Wrapper for constrained dynamics.}\label{alg:rotational_wrapper}
\begin{algorithmic}[1]
\Require Inner optimizer $\text{F}$, decay factor ${0 \le \beta < 1}$, $\epsilon \ge 0$ for numerical stability, iteration count $T$, rotational set $\Omega$
\For{$\vp$ in $\Omega$} \Comment{For each neuronal weight vector}
    \State $\nu_\vp \gets 0$ \Comment{Initialize the update RMS tracker}
    \State $n_\vp \gets \|\vp\|$ \Comment{Save the initial magnitude}
    \State ${\vp \gets n_\vp \cdot \frac{\vp - \bar{\vp}}{\|\vp - \bar{\vp}\|}}$ \Comment{Remove the mean component of $\vp$}
\EndFor

\For{$t \in \{1,...,T\}$}
    \State Perform backpropagation, obtain gradients for all params
    \ForAll{$\vp$} \Comment{For each parameter}
        \LComment{Get update components (\S~\ref{sec:equilibrium_geometric_model}):}
        \State $\Delta_g \vp, \Delta_\lambda \vp \gets \text{F.get\_update}\!\left(\vp, \nabla_\vp \Ls(\vp,\ldots)\right)$
    \If{$\vp \in \Omega$} \Comment{If $\vp$ is a neuronal weight vector}
        \State ${\Delta_g \vp \gets \Delta_g \vp / \eta}$ \Comment{Undo $\eta$ used in F}
        \LComment{Remove projections, s.t. $\Delta_g \vp \perp \vp$, $\Delta_g \vp \perp \vone$:}
        \State $\Delta_g \vp \gets \Delta_g \vp - \frac{\langle \Delta_g \vp, \vp \rangle}{\|\vp\|^2}\vp - \frac{\langle \Delta_g \vp, \vone \rangle}{\|\vone\|^2}\vone$ \label{line:remove_projection}
        \LComment{Update RMS tracker:}
        \State $\nu_\vp \gets \beta \cdot \nu_\vp + (1 - \beta) \cdot \|\Delta_g \vp\|^2 $\label{line:rms_update}
        \LComment{Rotate $\vp$ by $\widehat{\eta_r}$ from \cref{tab:equilibrium_summary} on average:}
        \State ${\vp \gets \vp + \widehat{\eta_r} \cdot n_\vp \cdot \frac{\Delta_g \vp}{\sqrt{\nu_\vp/(1-\beta^t)} + \epsilon}}$ \label{line:rotational_update}
        \LComment{Normalize $\vp$ to the initial magnitude:}
        \State ${\vp \gets n_\vp \cdot \frac{\vp}{\|\vp\|}}$
    \Else{} \Comment{$\vp$ is not a neuronal weight vector}
        \State $\vp \gets \vp + \Delta_g \vp + \Delta_\lambda \vp$  \Comment{Standard update}
    \EndIf
    \EndFor
\EndFor
\end{algorithmic}
\end{algorithm}

\section{Rotational Variants of Optimizers (RVs)}\label{sec:rotational_optimizers}
Our analysis shows how weight decay causes standard optimizers to transition towards equilibrium over time.
In the steady state, weight decay balances the rotation across weight vectors and scales $\eta_r$ relative to $\eta_g$.
\semph{The same effect can be achieved without weight decay by directly controlling the average angular update} as shown in \cref{alg:rotational_wrapper}.
We keep the weight magnitude constant and optionally introduce a learnable gain to compensate, which can matter for scale-sensitive weights and avoids numerical issues.
These \term{Rotational Variants (RVs)} of existing optimizers serve as valuable research tools that allow us to verify our understanding of weight decay and perform ablation studies.
Since RVs mimic the steady-state behavior of standard optimizers we expect similar performance.
However, the update dynamics also differ in certain ways:
\begin{itemize}[leftmargin=10pt,itemsep=-5pt,topsep=0pt]
    \item The RVs can perfectly balance the average rotation without relying on scale-invariance from e.g.\ normalization.
    \item The RVs are always in equilibrium so there is no transient-phase. This means that the specified learning rate schedule directly controls $\eta_r$ unlike in standard optimizers.
\end{itemize}
This simplifies the optimization dynamics and makes them more robust to architectural choices such as normalization.
\semph{Note that the RVs closely resemble relative optimizers like LARS~\cite{you2017lars} and LAMB~\cite{you2019lamb}, revealing how they relate to and differ from standard optimizers.}
The RVs are further described in \cref{appx:rotational_wrapper}.

\begin{figure}[tb]
    \centering
    \vspace{-9pt} %
    \includegraphics[width=\columnwidth]{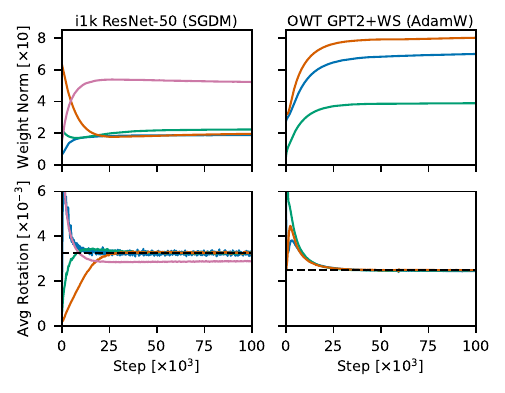}
    \vspace{-25pt} %
    \caption{
    Measured weight norms and average rotation for different layers (solid colors) in two real neural network training tasks, ResNet-50 on ImageNet-1k (SGDM) and Weight Standardized GPT2-124M on OpenWebText (AdamW). The predicted equilibrium rotation (dashed black) from \cref{tab:equilibrium_summary} holds very well for all scale-invariant layers. The final fully-connected layer in RN-50 (pink) is not scale-invariant with a radial gradient component that decreases the effective weight decay, slowing the rotation (see \cref{sec:scale_sensitive_dynamics}). The learning rate is constant for easier comparison.
    }
    \label{fig:measurements_rn50_gpt2ws}
    \vspace{-10pt}
\end{figure}

\section{Experiments \& Discussion}\label{sec:experiments}
In this section we experimentally validate our analysis of the neuronal update dynamics and explore their impact on training. See \cref{appx:experimental_details} for experimental setup and details.

\setlength{\tabcolsep}{4pt}
\begin{table*}[tb]
  \centering
  \caption{Test set performance (mean$\pm$std) over three seeds for the baseline optimizer, AdamW, and its rotational variant (RV). We use the baseline hyperparameters directly for the no weight decay ($\lambda=0$) and zero-shot results, but do minor tuning for the few-shot RV results where needed. The performance parity of the RV suggests that the benefits of weight decay, which are clear from the baseline degradation without it, can be achieved by directly controlling angular updates.}
  \vspace{-5pt}
  \label{tab:constraining_smd}
  \resizebox{1.0\textwidth}{!}{
  \begin{tabular}{@{}llll|c|c|c|c@{}}
    \toprule
    & & & & \multicolumn{2}{c|}{\textbf{AdamW}} & \multicolumn{2}{c}{\textbf{RV-AdamW}} \\
    \textbf{Dataset} & \textbf{Model} & \textbf{Batch Size} & \textbf{Metric ($\updownarrows$)} & \multicolumn{1}{c}{\textbf{Baseline}} & \multicolumn{1}{c|}{$\boldsymbol{\lambda=0}$} & \multicolumn{1}{c}{\textbf{(zero-shot)}} & \multicolumn{1}{c}{\textbf{(few-shot)}} \\
    \midrule
    CIFAR-10 & ResNet-20 & 128 & Top-1 Acc. [\%] ($\uparrow$) & 92.2$\pm$0.2 & 91.1$\pm$0.3 & 92.4$\pm$0.3 & N/A \\ 
    CIFAR-10 & ResNet-20 & 2048 & Top-1 Acc. [\%] ($\uparrow$) & 91.5$\pm$0.3 & 90.4$\pm$0.3 & 91.4$\pm$0.3 & 92.2$\pm$0.3 \\ 
    Imagenet-1k & DeiT tiny & 1024 & Top-1 Acc. [\%] ($\uparrow$) & 72.1 & 69.3 & 71.3 & 72.5 \\
    IWSLT2014 de-en & Transformer-S & 4096 & Bleu ($\uparrow$) & 34.6$\pm$0.1 & 34.6$\pm$0.1 & 20.0$\pm$0.1 & 34.7$\pm$0.1 \\
    Wikitext & GPT2-124M & 165$\times$512 & Perplexity ($\downarrow$) & 19.6$\pm$0.1 & \textit{unstable} & 19.0$\pm$0.2 & N/A \\
    OpenWebText & GPT2-124M & 480$\times$1024 & Loss ($\downarrow$) & 3.18$\pm$0.02 & 3.21$\pm$0.01 & 3.18$\pm$0.01 & N/A \\
    OpenWebText (25k) & GPT2-124M & 480$\times$1024 & Loss ($\downarrow$) & 2.94$\pm$0.01 & \textit{unstable} & 2.93$\pm$0.01 & N/A \\
    \bottomrule
  \end{tabular}
  }
\end{table*}

\subsection{Measuring \& Constraining the Update Dynamics}\label{sec:experiments_measure_and_constrain}
In \S\ref{sec:analysis} we derived how weight decay affects the neuronal update dynamics of a network undergoing a carefully constructed analytically tractable random walk.
\semph{Here we show that these dynamics occur in practical problems and that they suffice for obtaining the benefits of weight decay}.

\textbf{Measurements:} \Cref{fig:measurements_rn50_gpt2ws} shows the weight norm and average angular updates over time for several layers in RN-50 and a GPT2 variant.
For scale-invariant layers, the average rotation converges to the predicted equilibrium rotation $\widehat{\eta_r}$ over time as anticipated.
This value depends solely on the hyperparameters with no dependency on other unknown or time-varying quantities such as the gradient magnitude.

For layers that are not scale-invariant, the equilibrium rotation is affected by the average radial gradient component as expected (\S\ref{sec:scale_sensitive_dynamics}), but this deviation is often not very significant as seen for the final FC layer in \cref{fig:measurements_rn50_gpt2ws}.
Note that transformers are not scale-invariant without additional tricks like Weight Standardization (WS), but we find the equilibrium rotation is still often close to the scale-invariant value (see \cref{appx:sec:additional_experiments}, \cref{fig:gpt_dynamics_scale_sensitive} for GPT2 without WS).

For \cref{fig:measurements_rn50_gpt2ws} the length of the transient phase is small compared to the length of typical training.
ResNet-50 was originally trained for 90 epochs ($\sim$450k steps) and GPT2 for $\sim$600k steps.
However, this need not be the case, and will also depend on the hyperparameters as predicted by \cref{eq:adamw_norm_convergence}.

\textbf{Constraining the Rotational Dynamics:}
\Cref{tab:constraining_smd} shows the impact of replacing weight decay in AdamW via \cref{alg:rotational_wrapper} (RV-AdamW) for various network architectures and tasks.
To quantify the effect of weight decay on the baselines, we repeat them with the same hyperparameters aside from disabling weight decay, i.e.\ with\ $\lambda=0$.
Training without weight decay is similar to using a different learning rate schedule \citep{li2020exponential}.
In particular, \semph{for scale-invariant weights, disabling weight decay is similar to multiplying the learning rate schedule for $\eta_r$ with an exponentially decaying function.}
An example of this can be seen in \cref{appx:fig:adamw_vs_adam_dynamics} in \cref{appx:sec:additional_experiments}, which also shows the accompanying increase in the weight norms. %
The effects of this will vary with training length, the weight decay value used, the initial weight norms etc.
Without weight decay GPT training eventually diverges (at least in bfloat16), which was also observed before by \citet{andriushchenko2023wd}.

\Cref{tab:constraining_smd} shows that in most cases the RV perform similarly to the original baseline using exactly the same hyperparameters (zero-shot).
This is expected if the baseline training is dominated by the steady-state equilibrium phase that the RV is designed to model.
This is not always the case, for example in the IWSLT2014 the weight decay of the baseline is too low to have a significant impact over the course of training.
The resulting training never reaches the equilibrium the RV models, where the average rotation would be very low (remember that the weight decay value affects the convergence rate towards equilibrium, see \cref{eq:adamw_norm_convergence}).
In such cases we also perform a ``few-shot'' experiment where we do light tuning of the weight decay parameter.
We can roughly predict the modified value based on measurements of the observed rotation during the baseline training using \cref{tab:equilibrium_summary}.

\Cref{appx:sec:additional_experiments} shows experiments for other base optimizers.
In all cases we are able to recover the baseline performance with the RV.
\semph{This suggests that controlling the average angular update size is sufficient to obtain the benefits of weight decay.
Note that our intention here is not to outperform the baselines}, although we believe the insights from the RVs could help with this in the future.

\begin{figure*}[tb]
    \centering
    \vspace{-6pt} %
    \includegraphics[width=\textwidth]{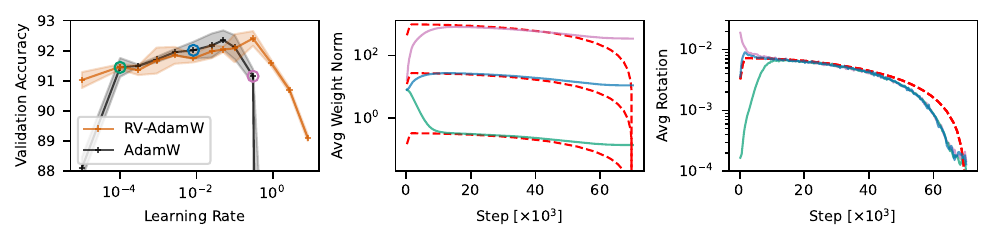}
    \vspace{-25pt} %
    \caption{
    Weight decay influences transient behavior and how fast weights are updated relative to biases. \textbf{Left:}~Validation accuracy for ResNet-20 on CIFAR-10 for learning rate, weight decay pairs with a constant product ($\eta\lambda=5\!\cdot\!10^{-4}$) resulting in a specific $\widehat{\eta_r}$ but different $\widehat{\eta_g}$ and \inlinevomegaeqnorm~(\cref{tab:equilibrium_summary}).
    \textbf{Middle/Right:}~The weight norm $\|\vomega\|$ and angular update size $\eta_r$ over time for three $(\eta,\lambda)$ pairs corresponding to the colored circles on the left with equilibrium predictions in dashed red.
    }
    \label{fig:lr-wd-trade-off}
    \vspace{-5pt}
\end{figure*}

\subsection{Transient Effects \& The Interaction of \texorpdfstring{$\eta$}{lr} and \texorpdfstring{$\lambda$}{wd}}\label{sec:experiments_scheduling}
In our analysis we describe two different update sizes, the angular update size $\eta_r$ (for neuronal weight vectors) and the RMS update size $\eta_g$ (for biases, gains etc).
The steady state value of $\eta_r$ depends on both the learning rate $\eta$ and weight decay $\lambda$ hyperparameters, while $\eta_g$ is not directly affected by weight decay (which is not applied to biases etc).
The initial $\eta_r$ also only depends on $\eta$, but $\lambda$ modulates it over time creating an induced ``effective'' learning rate schedule $\eta_r(t)$ that can deviate from the specified learning rate schedule $\eta(t)$.
Here we explore these effects experimentally.

\begin{figure*}[tb]
    \centering
    \begin{subfigure}[b]{2.166in}
        \includegraphics[width=\textwidth]{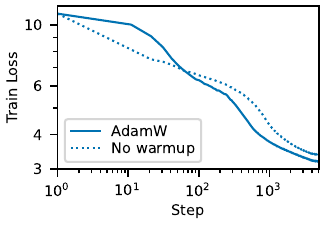}
    \end{subfigure}
    \hfill
    \begin{subfigure}[b]{2.166in}
        \includegraphics[width=\textwidth]{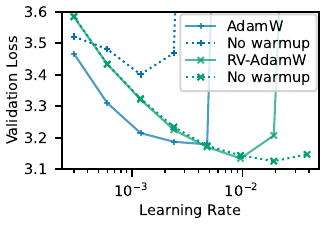}
    \end{subfigure}
    \hfill
    \begin{subfigure}[b]{2.166in}
        \includegraphics[width=\textwidth]{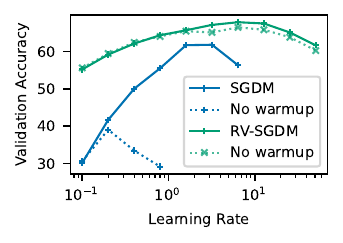}
    \end{subfigure}
    \vspace{-8pt}
    \caption{The RVs display a reduced need for learning rate warmup compared to standard optimizers, offering insights into the utility of warmups.  \textbf{Left:} GPT2-124M OWT loss curve with/without learning rate warmup. \textbf{Middle:} GPT2-124M OWT final validation loss for  different learning rates with AdamW/RV-AdamW and with/without warmup. \textbf{Right:} ResNet-50 i1k validation accuracy (short, large batch training) for different learning rates with SGDM/RV-SGDM, with/without warmup.}
    \label{fig:warmup}
    \vspace{-5pt}
\end{figure*}

\newpage
\textbf{Learning Rate vs Weight Decay:}
\Cref{fig:lr-wd-trade-off}L shows the final network performance obtained for different ($\eta$, $\lambda$) pairs with a constant $\eta\lambda$ product and therefore the same equilibrium rotation value.
With the RV, this is equivalent to changing the size of the bias updates via $\eta_g$, while keeping the rotation $\eta_r$ constant.
The performance varies quite a bit, the leftmost values at $\eta_g\!\approx\!0$ roughly match the results with frozen gains/biases (91.1\%), while the rightmost values have rapidly changing biases degrading performance.
Although we don't show it here, the $\eta_r$ value also matters significantly so $\eta$ and $\lambda$ jointly determine two distinct effective step sizes, $\eta_r$ and $\eta_g$, both of which affect performance.
\semph{Weight decay can therefore be seen as a scaling factor for the effective (equilibrium) update size of the weights ($\eta_r$) relative to other parameters such as biases ($\eta_g$).}
As a result we need to keep the $\lambda$ hyperparameter in the RVs even if we don't actually decay the weights.
The baseline optimizer (black in \cref{fig:lr-wd-trade-off}L) shows a similar trend but is also affected by changes in the transient phase which we will look at next.

\textbf{Transient Effects:}
Although the weight decay and learning rate have an identical effect on the  equilibrium rotation $\widehat{\eta_r}$ for AdamW, they affect the norm differently (see \cref{tab:equilibrium_summary}).
In \cref{fig:lr-wd-trade-off}M we plot the weight norm and equilibrium prediction for three different $(\eta, \lambda)$ pairs corresponding to the colored circles in \cref{fig:lr-wd-trade-off}L.
These experiments use a cosine decay learning rate schedule and a five epoch warmup.
Note how \semph{configurations with a higher learning rate (and lower weight decay) converge towards a higher equilibrium norm after the initial transient phase.}
Towards the later stages of training the weights fall back out of equilibrium as the weights can not decay fast enough to keep up with the shifting equilibrium norm (which decreases with the learning rate schedule).
We expect longer training would keep the weights in equilibrium for proportionally longer.

\Cref{fig:lr-wd-trade-off}R shows the measured angular updates corresponding to the previous $(\eta, \lambda)$ pairs.
We note how the rotational behavior out of equilibrium is inverted compared to that of the norm.
Weight norms below the equilibrium magnitude result in rotation that is faster than in equilibrium, and vice versa.
\semph{This means that depending on the initialization magnitude of the weights compared to the equilibrium magnitude, we either observe overly fast or slow rotation during the initial transient phase.}
The purple example shows that even though we use a learning rate warmup, the induced ``effective'' update schedule of $\eta_r$ can still include overly fast rotation.
The RVs eliminate these transient phases causing differences at both the start and end of training which can be significant, but could be replicated by scheduling $\eta$ or $\lambda$ appropriately if desired.

\begin{figure*}[tb]
    \vspace{-9pt}
    \centering
    \begin{subfigure}[b]{0.66\textwidth}
        \includegraphics[width=\textwidth]{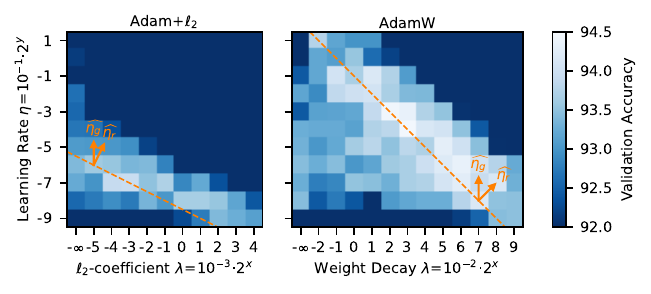}
    \end{subfigure}
    \hfill
    \begin{subfigure}[b]{0.33\textwidth}
        \includegraphics[width=\textwidth]{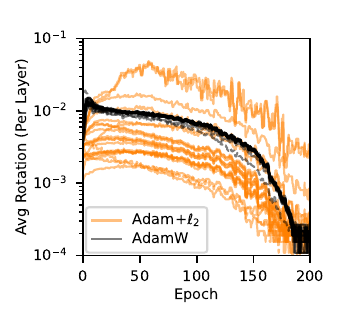}
    \end{subfigure}
    \vspace{-9pt}
    \caption{Balanced vs imbalanced rotation appears to be a key difference between AdamW and Adam+$\ell_2$. \textbf{Left:} A sweep of the learning rate and $\ell_2$-regularization / weight decay of Adam+$\ell_2$ and AdamW on CIFAR-10 ResNet-18. Adam+$\ell_2$ is unable to match the performance of AdamW. The orange dashed lines represent contour lines for $\widehat{\eta_r}$ based on \cref{tab:equilibrium_summary} along which $\widehat{\eta_g}$ varies, demonstrating how both update sizes matter. This dashed line is analogous to the sweep depicted in \cref{fig:lr-wd-trade-off}. \textbf{Right:}~The observed average rotation of each layer during training with the best setting for each optimizer. Here the rotation is affected by a cosine learning rate schedule (no warmup).}
    \label{fig:adam_vs_adamw_rn18}
\end{figure*}

\vspace{10pt}
\textbf{Need for Learning Rate Warmup:} We conjecture that the irregular update dynamics observed in the initial transient phase can hinder optimization.
In \cref{fig:warmup}L we show training curves for GPT2-124M OWT with and without a 5\% learning rate warmup.
The run without warmup is stable and makes faster initial progress but eventually falls behind, creating a loss gap that is not closed throughout training.
\Cref{fig:warmup}M compares the final validation loss with and without warmup for different learning rates for both AdamW and its rotational variant.
The AdamW runs show a significant gap between the two, but the difference is minimal with the RV.
We even observe the no-warmup runs being stable at slightly higher learning rates with the RV, but believe this is noise due to non-deterministic divergence.

In \cref{fig:warmup}R we do a similar experiment with 10 epoch ResNet-50 i1k training with a large batch size of 8k.
We again observe that the baseline benefits significantly from warmups, but the RV only marginally (which could be due to the dynamics of other parameters like biases which the RV does not stabilize).
We observed the same trend when extending the best runs to full 90 epochs (see \cref{appx:sec:additional_experiments}).

\semph{This suggests learning rate warmup may aid training in part by stabilizing the transient phase, and that explicitly controlling the angular updates could offer similar benefits.}

\subsection{The Importance of Balanced Rotation}\label{sec:experiments_balanced_rotation}
The previous sections have mostly glossed over the fact that equilibrium forms independently for various neurons and layers.
In certain settings this leads to different (imbalanced) equilibrium rotation between network components.
Here we explore the performance impact of this experimentally.

\vspace{10pt}
\textbf{Adam vs AdamW:}
Our analysis revealed the the equilibrium rotation of Adam+$\ell_2$ has a dependency on the gradient magnitude unlike AdamW.
When the gradient norm differs between neurons/layers this should lead to imbalanced rotation (even for scale-invariant layers).
\Cref{fig:adam_vs_adamw_rn18}L reproduces the performance gap between AdamW and Adam+$\ell_2$ observed by \citet{loshchilov2018decoupled}, around 0.5\% on the validation set.
Note how the structure of these heatmaps corresponds to changes in the two equilibrium update sizes $\widehat{\eta_r}$ and $\widehat{\eta_g}$, demonstrating how both matter in practice as suggested in \S\ref{sec:experiments_scheduling}.
The original decoupled version of AdamW described by \citet{loshchilov2018decoupled} differs from the one used in PyTorch and described here, replacing $\eta\lambda$ in the weight decay term of \cref{eq:adamw_update} with just $\lambda$, but scaling both $\eta$ and $\lambda$ over time according to the learning rate schedule.
In this case $\widehat{\eta_r}$ depends only on $\lambda$ and $\widehat{\eta_g}$ on $\eta$, explaining the separability of the hyperparameter space they observe.

\Cref{fig:adam_vs_adamw_rn18}R shows the rotation of different layers over training for the best configuration of each optimizer.
\semph{In AdamW all layers behave similarly, even the scale-sensitive final FC layer (dashed) does not deviate significantly, unlike Adam+$\ell_2$ where the rotation differs drastically.}
The $\sim$30$\times$ range in observed rotation corresponds to a factor of $\sim$1000$\times$ in the learning rate (\cref{tab:equilibrium_summary}).
In \cref{appx:sec:additional_experiments} we show that enforcing balanced rotation in Adam+$\ell_2$ roughly closes the gap.
We also observe imbalanced rotation impeding training in other settings (see later), so
\semph{we hypothesize that balanced equilibrium rotation is the main benefit of AdamW over Adam+$\ell_2$.}

\begin{figure*}[tb]
    \centering
    \begin{subfigure}[b]{2.166in}
        \includegraphics[width=\textwidth]{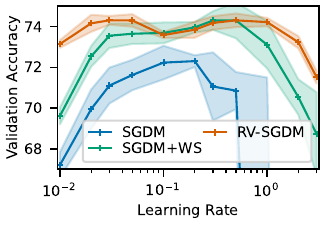}
    \end{subfigure}
    \hfill
    \begin{subfigure}[b]{4.333in}
        \includegraphics[width=\textwidth]{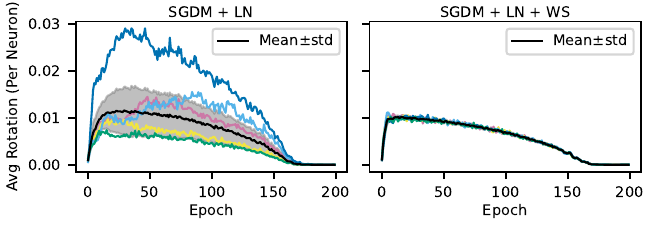}
    \end{subfigure}
    \vspace{-5pt}
    \caption{Balanced neuronal rotation obtained through e.g.\ RVs or Weight Standardization seems to aid training. \textbf{Left:} Final accuracy for a Layer-Normalized (LN) ResNet-18 on CIFAR-100 for different peak learning rates (cos+warmup schedule). Baseline compared with additional Weight Standardization (WS) and RV training. \textbf{Right:} The baseline has imbalanced rotation across neurons, WS balances them. Example neurons (each color) and mean$\pm$std (black/gray) shown for one layer.}
    \label{fig:layernorm_balanced}
    \vspace{-10pt} %
\end{figure*}

\newpage
\textbf{The Benefit of Weight Standardization:}
Layer Normalization can make a whole layer scale-invariant but not individual neurons (unlike e.g.\ Batch Normalization).
This means that the gradient is orthogonal to the weights of the whole layer (see \cref{eq:grad_perp}) but individual neurons may have a non-zero radial component.
In fact, the normalized output magnitude of one neuron can be increased by decreasing the output magnitude of the other neurons in the layer and vice versa, potentially encouraging radial gradient components on the neuron level.
This changes the ``effective'' weight decay as discussed in \S\ref{sec:scale_sensitive_dynamics}, which can lead to imbalanced rotation between neurons.
Weight Standardization (WS) makes each neuron scale invariant eliminating this effect.
Note that WS is fully independent of the network inputs and can thus be seen as a reparameterization or an optimization trick.
Other types of normalization like Batch Normalization can have additional effects like affecting signal propagation \citep{brock2021signalpropagation} or causing a dropout-like regularization effect~\cite{kosson2024ghost}.

\Cref{fig:layernorm_balanced}L shows the final accuracy achieved when training a Layer-Normalized ResNet-18 on CIFAR-100 for different learning rates.
Training using either Weight Standardization or an RV significantly outperforms standard training.
\cref{fig:layernorm_balanced}R shows imbalanced neuronal rotation in the baseline which the WS eliminates as expected (the RV does as well).
\semph{Weight Standardization thus appears to facilitate optimization by balancing the equilibrium dynamics across neurons}, especially when applied on top of Layer Normalization or Group Normalization~\cite{wu2018group}.

\textbf{Imbalanced Rotation:}
We can modify the RVs to directly test the impact of imbalanced rotation without potential confounding effects.
In \cref{appx:sec:additional_experiments} (\cref{fig:rotation_speed_control}) we show that an RV with artificially imbalanced neuronal rotation, where some neurons are intentionally rotated slower or faster, suffers from a performance degradation compared to a standard balanced version.
This gap persists despite tuning the hyperparameters for each configuration.
\semph{We observe that even small imbalances in the rotational speed ($\eta_r$) can significantly degrade performance.}

\section{Conclusion}
In this work we have described how weight decay modulates the update dynamics of individual neurons and explored how their variance across components (balanced vs imbalanced) and time (transient vs equilibrium phases) impacts training.
Despite their simplicity, neural update dynamics offer a surprisingly insightful perspective of neural network optimization and demystify the effectiveness of widely adopted techniques.
We believe this area presents numerous opportunities for further theoretical and practical research.

\section*{Impact Statement}
This paper presents work whose goal is to advance the field of Machine Learning. There are many potential societal consequences of our work, none which we feel must be specifically highlighted here.

\section*{Acknowledgements}
We thank Maksym Andriushchenko for insightful conversations related to this study. We also appreciate Alex Hagele, Dongyang Fan and Nikita Doikov for their feedback on the manuscript which enhanced its quality and clarity.

\bibliography{main}

\begin{thebibliography}{54}
\providecommand{\natexlab}[1]{#1}
\providecommand{\url}[1]{\texttt{#1}}
\expandafter\ifx\csname urlstyle\endcsname\relax
  \providecommand{\doi}[1]{doi: #1}\else
  \providecommand{\doi}{doi: \begingroup \urlstyle{rm}\Url}\fi

\bibitem[Andriushchenko et~al.(2023)Andriushchenko, D'Angelo, Varre, and
  Flammarion]{andriushchenko2023wd}
Andriushchenko, M., D'Angelo, F., Varre, A., and Flammarion, N.
\newblock Why do we need weight decay in modern deep learning?
\newblock \emph{arXiv preprint arXiv:2310.04415}, 2023.
\newblock URL \url{https://arxiv.org/abs/2310.04415}.

\bibitem[Ba et~al.(2016)Ba, Kiros, and Hinton]{ba2016layer}
Ba, J.~L., Kiros, J.~R., and Hinton, G.~E.
\newblock Layer normalization.
\newblock \emph{arXiv preprint arXiv:1607.06450}, 2016.
\newblock URL \url{https://arxiv.org/abs/1607.06450}.

\bibitem[Brock et~al.(2021{\natexlab{a}})Brock, De, and
  Smith]{brock2021signalpropagation}
Brock, A., De, S., and Smith, S.~L.
\newblock Characterizing signal propagation to close the performance gap in
  unnormalized resnets.
\newblock In \emph{9th International Conference on Learning Representations,
  {ICLR}}, 2021{\natexlab{a}}.
\newblock \href{https://arxiv.org/abs/2101.08692}{arXiv:2101.08692}.

\bibitem[Brock et~al.(2021{\natexlab{b}})Brock, De, Smith, and
  Simonyan]{brock2021high}
Brock, A., De, S., Smith, S.~L., and Simonyan, K.
\newblock High-performance large-scale image recognition without normalization.
\newblock In \emph{International Conference on Machine Learning}, pp.\
  1059--1071. PMLR, 2021{\natexlab{b}}.
\newblock \href{https://arxiv.org/abs/2102.06171}{arXiv:2102.06171}.

\bibitem[Cettolo et~al.(2014)Cettolo, Niehues, St{\"u}ker, Bentivogli, and
  Federico]{cettolo14iwslt}
Cettolo, M., Niehues, J., St{\"u}ker, S., Bentivogli, L., and Federico, M.
\newblock Report on the 11th {IWSLT} evaluation campaign.
\newblock In \emph{Proceedings of the 11th International Workshop on Spoken
  Language Translation: Evaluation Campaign}, pp.\  2--17, Lake Tahoe,
  California, December 4-5 2014.
\newblock URL \url{https://aclanthology.org/2014.iwslt-evaluation.1}.

\bibitem[Chen et~al.(2023)Chen, Liang, Huang, Real, Wang, Pham, Dong, Luong,
  Hsieh, Lu, and Le]{chen2023symbolic}
Chen, X., Liang, C., Huang, D., Real, E., Wang, K., Pham, H., Dong, X., Luong,
  T., Hsieh, C.-J., Lu, Y., and Le, Q.~V.
\newblock Symbolic discovery of optimization algorithms.
\newblock In \emph{Thirty-seventh Conference on Neural Information Processing
  Systems}, 2023.
\newblock URL \url{https://openreview.net/forum?id=ne6zeqLFCZ}.
\newblock \href{https://arxiv.org/abs/2302.06675}{arXiv:2302.06675}.

\bibitem[Chiley et~al.(2019)Chiley, Sharapov, Kosson, Koster, Reece, Samaniego
  de~la Fuente, Subbiah, and James]{chiley2019online}
Chiley, V., Sharapov, I., Kosson, A., Koster, U., Reece, R., Samaniego de~la
  Fuente, S., Subbiah, V., and James, M.
\newblock Online normalization for training neural networks.
\newblock \emph{Advances in Neural Information Processing Systems}, 32, 2019.
\newblock \href{https://arxiv.org/abs/1905.05894}{arXiv:1905.05894}.

\bibitem[Fu et~al.(2023)Fu, Wang, Zhang, Zhang, Chen, and
  Zheng]{fu2023momentum}
Fu, J., Wang, B., Zhang, H., Zhang, Z., Chen, W., and Zheng, N.
\newblock When and why momentum accelerates sgd: An empirical study.
\newblock \emph{arXiv preprint arXiv:2306.09000}, 2023.
\newblock URL \url{https://arxiv.org/abs/2306.09000}.

\bibitem[Goodfellow et~al.(2016)Goodfellow, Bengio, and
  Courville]{Goodfellow-et-al-2016}
Goodfellow, I., Bengio, Y., and Courville, A.
\newblock \emph{Deep Learning}.
\newblock MIT Press, 2016.
\newblock \url{http://www.deeplearningbook.org}.

\bibitem[He et~al.(2016)He, Zhang, Ren, and Sun]{he2016deep}
He, K., Zhang, X., Ren, S., and Sun, J.
\newblock Deep residual learning for image recognition.
\newblock In \emph{Proceedings of the IEEE Conference on Computer Vision and
  Pattern Recognition (CVPR)}, June 2016.
\newblock \href{https://arxiv.org/abs/1512.03385}{arXiv:1512.03385}.

\bibitem[Heo et~al.(2021)Heo, Chun, Oh, Han, Yun, Kim, Uh, and
  Ha]{heo2021adamp}
Heo, B., Chun, S., Oh, S.~J., Han, D., Yun, S., Kim, G., Uh, Y., and Ha, J.-W.
\newblock Adamp: Slowing down the slowdown for momentum optimizers on
  scale-invariant weights.
\newblock In \emph{International Conference on Learning Representations}, 2021.
\newblock URL \url{https://openreview.net/forum?id=Iz3zU3M316D}.
\newblock \href{https://arxiv.org/abs/2006.08217}{arXiv:2006.08217}.

\bibitem[Hoffer et~al.(2018)Hoffer, Banner, Golan, and Soudry]{hoffer2018norm}
Hoffer, E., Banner, R., Golan, I., and Soudry, D.
\newblock Norm matters: efficient and accurate normalization schemes in deep
  networks.
\newblock \emph{Advances in Neural Information Processing Systems}, 31, 2018.
\newblock \href{https://arxiv.org/abs/1803.01814}{arXiv:1803.01814}.

\bibitem[Hoffmann et~al.(2022)Hoffmann, Borgeaud, Mensch, Buchatskaya, Cai,
  Rutherford, Casas, Hendricks, Welbl, Clark, et~al.]{hoffmann2022training}
Hoffmann, J., Borgeaud, S., Mensch, A., Buchatskaya, E., Cai, T., Rutherford,
  E., Casas, D. d.~L., Hendricks, L.~A., Welbl, J., Clark, A., et~al.
\newblock Training compute-optimal large language models.
\newblock \emph{arXiv preprint arXiv:2203.15556}, 2022.
\newblock URL \url{https://arxiv.org/abs/2203.15556}.

\bibitem[Huang et~al.(2017{\natexlab{a}})Huang, Liu, Lang, and
  Li]{huang2017projection}
Huang, L., Liu, X., Lang, B., and Li, B.
\newblock Projection based weight normalization for deep neural networks.
\newblock \emph{arXiv preprint arXiv:1710.02338}, 2017{\natexlab{a}}.
\newblock URL \url{https://arxiv.org/abs/1710.02338}.

\bibitem[Huang et~al.(2017{\natexlab{b}})Huang, Liu, Liu, Lang, and
  Tao]{huang2017centered}
Huang, L., Liu, X., Liu, Y., Lang, B., and Tao, D.
\newblock Centered weight normalization in accelerating training of deep neural
  networks.
\newblock In \emph{Proceedings of the IEEE International Conference on Computer
  Vision (ICCV)}, pp.\  2803--2811, 2017{\natexlab{b}}.
\newblock URL
  \url{https://openaccess.thecvf.com/content_iccv_2017/html/Huang_Centered_Weight_Normalization_ICCV_2017_paper.html}.

\bibitem[Ioffe \& Szegedy(2015)Ioffe and Szegedy]{ioffe2015batch}
Ioffe, S. and Szegedy, C.
\newblock Batch normalization: Accelerating deep network training by reducing
  internal covariate shift.
\newblock In \emph{International conference on machine learning}, pp.\
  448--456. pmlr, 2015.
\newblock \href{https://arxiv.org/abs/1502.03167}{arXiv:1502.03167}.

\bibitem[Jia et~al.(2018)Jia, Song, He, Wang, Rong, Zhou, Xie, Guo, Yang, Yu,
  et~al.]{jia2018highly}
Jia, X., Song, S., He, W., Wang, Y., Rong, H., Zhou, F., Xie, L., Guo, Z.,
  Yang, Y., Yu, L., et~al.
\newblock Highly scalable deep learning training system with mixed-precision:
  Training imagenet in four minutes.
\newblock \emph{arXiv preprint arXiv:1807.11205}, 2018.
\newblock \href{https://arxiv.org/abs/1807.11205}{arXiv:1807.11205}.

\bibitem[Karpathy(2023)]{nanoGPT}
Karpathy, A.
\newblock nanogpt.
\newblock \url{https://github.com/karpathy/nanoGPT/}, 2023.

\bibitem[Karras et~al.(2023)Karras, Aittala, Lehtinen, Hellsten, Aila, and
  Laine]{karras2023analyzing}
Karras, T., Aittala, M., Lehtinen, J., Hellsten, J., Aila, T., and Laine, S.
\newblock Analyzing and improving the training dynamics of diffusion models.
\newblock \emph{arXiv preprint
  \href{https://arxiv.org/abs/2312.02696}{arXiv:2312.02696}}, 2023.

\bibitem[Kingma \& Ba(2015)Kingma and Ba]{kingma15adam}
Kingma, D. and Ba, J.
\newblock Adam: A method for stochastic optimization.
\newblock In \emph{International Conference on Learning Representations
  (ICLR)}, San Diega, CA, USA, 2015.
\newblock \href{https://arxiv.org/abs/1412.6980}{arXiv:1412.6980}.

\bibitem[Kodryan et~al.(2022)Kodryan, Lobacheva, Nakhodnov, and
  Vetrov]{kodryan2022training}
Kodryan, M., Lobacheva, E., Nakhodnov, M., and Vetrov, D.~P.
\newblock Training scale-invariant neural networks on the sphere can happen in
  three regimes.
\newblock \emph{Advances in Neural Information Processing Systems},
  35:\penalty0 14058--14070, 2022.
\newblock \href{https://arxiv.org/abs/2209.03695}{arXiv:2209.03695}.

\bibitem[Kosson et~al.(2024)Kosson, Fan, and Jaggi]{kosson2024ghost}
Kosson, A., Fan, D., and Jaggi, M.
\newblock Ghost noise for regularizing deep neural networks.
\newblock \emph{Proceedings of the AAAI Conference on Artificial Intelligence},
  38\penalty0 (12):\penalty0 13274--13282, Mar. 2024.
\newblock \doi{10.1609/aaai.v38i12.29228}.
\newblock URL \url{https://ojs.aaai.org/index.php/AAAI/article/view/29228}.
\newblock \href{https://arxiv.org/abs/2305.17205}{arXiv:2305.17205}.

\bibitem[Krizhevsky(2009)]{Krizhevsky2009LearningML}
Krizhevsky, A.
\newblock Learning multiple layers of features from tiny images.
\newblock \emph{self-published}, 2009.
\newblock URL
  \url{https://www.cs.toronto.edu/~kriz/learning-features-2009-TR.pdf}.

\bibitem[Li \& Arora(2020)Li and Arora]{li2020exponential}
Li, Z. and Arora, S.
\newblock An exponential learning rate schedule for deep learning.
\newblock In \emph{International Conference on Learning Representations}, 2020.
\newblock URL \url{https://openreview.net/forum?id=rJg8TeSFDH}.
\newblock \href{https://arxiv.org/abs/1910.07454}{arXiv:1910.07454}.

\bibitem[Li et~al.(2020)Li, Lyu, and Arora]{li2020reconciling}
Li, Z., Lyu, K., and Arora, S.
\newblock Reconciling modern deep learning with traditional optimization
  analyses: The intrinsic learning rate.
\newblock \emph{Advances in Neural Information Processing Systems},
  33:\penalty0 14544--14555, 2020.
\newblock \href{https://arxiv.org/abs/2010.02916}{arXiv:2010.02916}.

\bibitem[Li et~al.(2021)Li, Malladi, and Arora]{li2021validity}
Li, Z., Malladi, S., and Arora, S.
\newblock On the validity of modeling {SGD} with stochastic differential
  equations ({SDE}s).
\newblock In Beygelzimer, A., Dauphin, Y., Liang, P., and Vaughan, J.~W.
  (eds.), \emph{Advances in Neural Information Processing Systems}, 2021.
\newblock URL \url{https://openreview.net/forum?id=goEdyJ_nVQI}.
\newblock \href{https://arxiv.org/abs/2102.12470}{arXiv:2102.12470}.

\bibitem[Li et~al.(2022{\natexlab{a}})Li, Bhojanapalli, Zaheer, Reddi, and
  Kumar]{li2022robust}
Li, Z., Bhojanapalli, S., Zaheer, M., Reddi, S., and Kumar, S.
\newblock Robust training of neural networks using scale invariant
  architectures.
\newblock In \emph{International Conference on Machine Learning}, pp.\
  12656--12684. PMLR, 2022{\natexlab{a}}.
\newblock \href{https://arxiv.org/abs/2202.00980}{arXiv:2202.00980}.

\bibitem[Li et~al.(2022{\natexlab{b}})Li, Wang, and Yu]{li2022fast}
Li, Z., Wang, T., and Yu, D.
\newblock Fast mixing of stochastic gradient descent with normalization and
  weight decay.
\newblock \emph{Advances in Neural Information Processing Systems},
  35:\penalty0 9233--9248, 2022{\natexlab{b}}.
\newblock URL
  \url{https://proceedings.neurips.cc/paper_files/paper/2022/hash/3c215225324f9988858602dc92219615-Abstract-Conference.html}.

\bibitem[Liu et~al.(2021)Liu, Bernstein, Meister, and Yue]{liu2021nero}
Liu, Y., Bernstein, J., Meister, M., and Yue, Y.
\newblock Learning by turning: Neural architecture aware optimisation.
\newblock In \emph{International Conference on Machine Learning}, pp.\
  6748--6758. PMLR, 2021.
\newblock \href{https://arxiv.org/abs/2102.07227}{arXiv:2102.07227}.

\bibitem[Loshchilov \& Hutter(2019)Loshchilov and
  Hutter]{loshchilov2018decoupled}
Loshchilov, I. and Hutter, F.
\newblock Decoupled weight decay regularization.
\newblock In \emph{International Conference on Learning Representations}, 2019.
\newblock URL \url{https://openreview.net/forum?id=Bkg6RiCqY7}.
\newblock \href{https://arxiv.org/abs/1711.05101}{arXiv:1711.05101}.

\bibitem[Malladi et~al.(2022)Malladi, Lyu, Panigrahi, and Arora]{malladi2022on}
Malladi, S., Lyu, K., Panigrahi, A., and Arora, S.
\newblock On the {SDE}s and scaling rules for adaptive gradient algorithms.
\newblock In Oh, A.~H., Agarwal, A., Belgrave, D., and Cho, K. (eds.),
  \emph{Advances in Neural Information Processing Systems}, 2022.
\newblock URL \url{https://openreview.net/forum?id=F2mhzjHkQP}.
\newblock \href{https://arxiv.org/abs/2205.10287}{arXiv:2205.10287}.

\bibitem[McCandlish et~al.(2018)McCandlish, Kaplan, Amodei, and
  Team]{mccandlish2018empirical}
McCandlish, S., Kaplan, J., Amodei, D., and Team, O.~D.
\newblock An empirical model of large-batch training.
\newblock \emph{arXiv preprint
  \href{https://arxiv.org/abs/1812.06162}{arXiv:1812.06162}}, 2018.

\bibitem[Merity et~al.(2017)Merity, Xiong, Bradbury, and
  Socher]{merity2017pointer}
Merity, S., Xiong, C., Bradbury, J., and Socher, R.
\newblock Pointer sentinel mixture models.
\newblock In \emph{International Conference on Learning Representations}, 2017.
\newblock URL \url{https://openreview.net/forum?id=Byj72udxe}.
\newblock \href{https://arxiv.org/abs/1609.07843}{arXiv:1609.07843}.

\bibitem[Neyshabur et~al.(2015)Neyshabur, Salakhutdinov, and
  Srebro]{neyshabur2015path}
Neyshabur, B., Salakhutdinov, R.~R., and Srebro, N.
\newblock Path-sgd: Path-normalized optimization in deep neural networks.
\newblock \emph{Advances in neural information processing systems}, 28, 2015.
\newblock \href{https://arxiv.org/abs/1506.02617}{arXiv:1506.02617}.

\bibitem[Ott et~al.(2019)Ott, Edunov, Baevski, Fan, Gross, Ng, Grangier, and
  Auli]{ott2019fairseq}
Ott, M., Edunov, S., Baevski, A., Fan, A., Gross, S., Ng, N., Grangier, D., and
  Auli, M.
\newblock fairseq: A fast, extensible toolkit for sequence modeling.
\newblock In \emph{Proceedings of NAACL-HLT 2019: Demonstrations}, 2019.
\newblock \href{https://arxiv.org/abs/1904.01038}{arXiv:1904.01038}.

\bibitem[Pagliardini(2023)]{llm_baseline}
Pagliardini, M.
\newblock llm-baseline.
\newblock \url{https://github.com/epfml/llm-baselines}, 2023.

\bibitem[Paszke et~al.(2019)Paszke, Gross, Massa, Lerer, Bradbury, Chanan,
  Killeen, Lin, Gimelshein, Antiga, et~al.]{paszke2019pytorch}
Paszke, A., Gross, S., Massa, F., Lerer, A., Bradbury, J., Chanan, G., Killeen,
  T., Lin, Z., Gimelshein, N., Antiga, L., et~al.
\newblock Pytorch: An imperative style, high-performance deep learning library.
\newblock \emph{Advances in neural information processing systems}, 32, 2019.
\newblock \href{https://arxiv.org/abs/1912.01703}{arXiv:1912.01703}.

\bibitem[Qiao et~al.(2019)Qiao, Wang, Liu, Shen, and Yuille]{qiao2019ws}
Qiao, S., Wang, H., Liu, C., Shen, W., and Yuille, A.~L.
\newblock Weight standardization.
\newblock \emph{CoRR}, abs/1903.10520, 2019.
\newblock URL \url{http://arxiv.org/abs/1903.10520}.

\bibitem[Radford et~al.(2019)Radford, Wu, Child, Luan, Amodei, and
  Sutskever]{radford2019language}
Radford, A., Wu, J., Child, R., Luan, D., Amodei, D., and Sutskever, I.
\newblock Language models are unsupervised multitask learners.
\newblock \emph{self-published}, 2019.
\newblock URL
  \url{https://d4mucfpksywv.cloudfront.net/better-language-models/language_models_are_unsupervised_multitask_learners.pdf}.

\bibitem[Roburin et~al.(2020)Roburin, de~Mont-Marin, Bursuc, Marlet, Perez, and
  Aubry]{roburin2020spherical}
Roburin, S., de~Mont-Marin, Y., Bursuc, A., Marlet, R., Perez, P., and Aubry,
  M.
\newblock A spherical analysis of adam with batch normalization.
\newblock \emph{arXiv preprint arXiv:2006.13382}, 2020.
\newblock URL \url{https://arxiv.org/abs/2006.13382}.

\bibitem[Russakovsky et~al.(2015)Russakovsky, Deng, Su, Krause, Satheesh, Ma,
  Huang, Karpathy, Khosla, Bernstein, Berg, and Fei-Fei]{imagenet15russakovsky}
Russakovsky, O., Deng, J., Su, H., Krause, J., Satheesh, S., Ma, S., Huang, Z.,
  Karpathy, A., Khosla, A., Bernstein, M., Berg, A.~C., and Fei-Fei, L.
\newblock {ImageNet Large Scale Visual Recognition Challenge}.
\newblock \emph{International Journal of Computer Vision (IJCV)}, 115\penalty0
  (3):\penalty0 211--252, 2015.
\newblock \doi{10.1007/s11263-015-0816-y}.
\newblock \href{https://arxiv.org/abs/1409.0575}{arXiv:1409.0575}.

\bibitem[Salimans \& Kingma(2016)Salimans and Kingma]{salimans2016weight}
Salimans, T. and Kingma, D.~P.
\newblock Weight normalization: A simple reparameterization to accelerate
  training of deep neural networks.
\newblock \emph{Advances in neural information processing systems}, 29, 2016.
\newblock \href{https://arxiv.org/abs/1602.07868}{arXiv:1602.07868}.

\bibitem[Shallue et~al.(2019)Shallue, Lee, Antognini, Sohl-Dickstein, Frostig,
  and Dahl]{shallue2019measuring}
Shallue, C.~J., Lee, J., Antognini, J., Sohl-Dickstein, J., Frostig, R., and
  Dahl, G.~E.
\newblock Measuring the effects of data parallelism on neural network training.
\newblock \emph{Journal of Machine Learning Research}, 20\penalty0
  (112):\penalty0 1--49, 2019.
\newblock \href{https://arxiv.org/abs/1811.03600}{arXiv:1811.03600}.

\bibitem[Simonyan \& Zisserman(2015)Simonyan and Zisserman]{simonyan2014very}
Simonyan, K. and Zisserman, A.
\newblock Very deep convolutional networks for large-scale image recognition.
\newblock In \emph{3rd International Conference on Learning Representations,
  {ICLR} 2015, San Diego, CA, USA, May 7-9, 2015, Conference Track
  Proceedings}, 2015.
\newblock URL \url{http://arxiv.org/abs/1409.1556}.

\bibitem[Touvron et~al.(2021)Touvron, Cord, Douze, Massa, Sablayrolles, and
  Jegou]{deit_tiny}
Touvron, H., Cord, M., Douze, M., Massa, F., Sablayrolles, A., and Jegou, H.
\newblock Training data-efficient image transformers \& distillation through
  attention.
\newblock In Meila, M. and Zhang, T. (eds.), \emph{Proceedings of the 38th
  International Conference on Machine Learning}, volume 139 of
  \emph{Proceedings of Machine Learning Research}, pp.\  10347--10357. PMLR,
  18--24 Jul 2021.
\newblock URL \url{https://proceedings.mlr.press/v139/touvron21a.html}.
\newblock \href{https://arxiv.org/abs/2012.12877}{arXiv:2012.12877}.

\bibitem[Van~Laarhoven(2017)]{van2017l2}
Van~Laarhoven, T.
\newblock L2 regularization versus batch and weight normalization.
\newblock \emph{arXiv preprint arXiv:1706.05350}, 2017.
\newblock URL \url{https://arxiv.org/abs/1706.05350}.

\bibitem[Wan et~al.(2021)Wan, Zhu, Zhang, and Sun]{wan2021spherical}
Wan, R., Zhu, Z., Zhang, X., and Sun, J.
\newblock Spherical motion dynamics: Learning dynamics of normalized neural
  network using sgd and weight decay.
\newblock In Ranzato, M., Beygelzimer, A., Dauphin, Y., Liang, P., and Vaughan,
  J.~W. (eds.), \emph{Advances in Neural Information Processing Systems},
  volume~34, pp.\  6380--6391. Curran Associates, Inc., 2021.
\newblock URL
  \url{https://proceedings.neurips.cc/paper/2021/file/326a8c055c0d04f5b06544665d8bb3ea-Paper.pdf}.
\newblock \href{https://arxiv.org/abs/2006.08419}{arXiv:2006.08419}.

\bibitem[Wightman(2019)]{rw2019timm}
Wightman, R.
\newblock Pytorch image models.
\newblock \url{https://github.com/rwightman/pytorch-image-models}, 2019.

\bibitem[Wu \& He(2018)Wu and He]{wu2018group}
Wu, Y. and He, K.
\newblock Group normalization.
\newblock In \emph{Proceedings of the European conference on computer vision
  (ECCV)}, pp.\  3--19, 2018.
\newblock \href{https://arxiv.org/abs/1803.08494}{arXiv:1803.08494}.

\bibitem[Xie et~al.(2023)Xie, zhiqiang xu, Zhang, Sato, and
  Sugiyama]{xie2023on}
Xie, Z., zhiqiang xu, Zhang, J., Sato, I., and Sugiyama, M.
\newblock On the overlooked pitfalls of weight decay and how to mitigate them:
  A gradient-norm perspective.
\newblock In \emph{Thirty-seventh Conference on Neural Information Processing
  Systems}, 2023.
\newblock URL \url{https://openreview.net/forum?id=vnGcubtzR1}.
\newblock \href{https://arxiv.org/abs/2011.11152}{arXiv:2011.11152}.

\bibitem[You et~al.(2017)You, Gitman, and Ginsburg]{you2017lars}
You, Y., Gitman, I., and Ginsburg, B.
\newblock Large batch training of convolutional networks.
\newblock \emph{arXiv preprint arXiv:1708.03888}, 2017.
\newblock URL \url{https://arxiv.org/abs/1708.03888}.

\bibitem[You et~al.(2020)You, Li, Reddi, Hseu, Kumar, Bhojanapalli, Song,
  Demmel, Keutzer, and Hsieh]{you2019lamb}
You, Y., Li, J., Reddi, S., Hseu, J., Kumar, S., Bhojanapalli, S., Song, X.,
  Demmel, J., Keutzer, K., and Hsieh, C.-J.
\newblock Large batch optimization for deep learning: Training bert in 76
  minutes.
\newblock In \emph{International Conference on Learning Representations}, 2020.
\newblock URL \url{https://openreview.net/forum?id=Syx4wnEtvH}.
\newblock \href{https://arxiv.org/abs/1904.00962}{arXiv:1904.00962}.

\bibitem[Zhang et~al.(2019)Zhang, Wang, Xu, and Grosse]{zhang2018three}
Zhang, G., Wang, C., Xu, B., and Grosse, R.
\newblock Three mechanisms of weight decay regularization.
\newblock In \emph{International Conference on Learning Representations}, 2019.
\newblock URL \url{https://openreview.net/forum?id=B1lz-3Rct7}.
\newblock \href{https://arxiv.org/abs/1810.12281}{arXiv:1810.12281}.

\bibitem[Zhou et~al.(2021)Zhou, Sun, and Zhong]{zhou2021fixnorm}
Zhou, Y., Sun, Y., and Zhong, Z.
\newblock Fixnorm: Dissecting weight decay for training deep neural networks.
\newblock \emph{arXiv preprint arXiv:2103.15345}, 2021.
\newblock URL \url{https://arxiv.org/abs/2103.15345}.

\end{thebibliography}
\bibliographystyle{icml2024}

\onecolumn
\newpage
\appendix

\section{Expanded Related Work}\label{appx:related_work}
In this section we discuss more related works divided into five main categories.

\subsection{Scale-Invariance and Effective Learning Rates}
Several works have investigated how the scale-invariance results in a certain ``effective learning rate'' based on the relative change that varies based on the norm of the weights, often in the form of $\eta / \|\vomega\|^2$.
The works in this section do not describe how $\|\vomega\|$ can converge to an equilibrium value that results in a fixed relative or rotational learning rate.
In Weight Normalization, \citet{salimans2016weight} point out how normalization can make parameters scale-invariant and that the gradient magnitude varies based on the weight magnitude.
They describe how the gradient can ``self-stabilize its norm'', with larger gradients becoming smaller over time due to growth in the weight magnitude, but do not consider the effects of weight decay on this process.
\citet{zhang2018three} and \citet{hoffer2018norm} empirically find that the regularization effects of weight decay are primarily caused by increases in the effective learning rate due to decreased weight norms.
\citet{li2020exponential} show that weight decay can be replaced by an exponentially increasing learning rate when optimizing scale-invariant weights with SGDM.

\subsection{Equilibrium}
The works in this section also consider the fact that the weight norm converges to a specific value and they explore the resulting effects on the relative update size.
\citet{van2017l2} points out the scale-invariance property of normalization and how it interacts with $\ell_2$-regularization.
They derive the $\eta / \|\vomega\|^2$ as the effective learning rate and show there exists a fixed point where the weight norms are stable.
Their work does not consider convergence of the weight magnitude as a separate process from the overall convergence of the loss and weights.
In Online Normalization, \citet{chiley2019online} show a simple derivation of the equilibrium condition in SGD and how it results in a relative update size that is identical across layers.
The Spherical Motion Dynamics (SMD)~\citep{wan2021spherical} expands on prior work by deriving the convergence of the weight norm and extending the analysis to include momentum.
They also show plots of the weight norm over the course of training, providing empirical evidence for early convergence of the weight norm and note how it can fall out of equilibrium with sudden learning rate changes or when the learning rate becomes too small.
They also consider the angular updates, empirically and analytically showing that they converge to an equilibrium value.
\citet{li2020reconciling} analyze the convergence of SGD to equilibrium by modelling it as a stochastic differential equation, arriving at similar conclusion as the SMD paper (without momentum).
This is expanded upon by \citet{li2022fast}.

\subsection{Understanding and Improving Weight Decay}
In traditional literature (e.g.~\citet{Goodfellow-et-al-2016}), $\ell_2$-regularization is introduced as a simple and commonly used regularization method achieved by adding $\frac{1}{2}\lambda\|\vomega\|^2$ to the objective function.
This strategy is commonly referred to as weight decay, but this is confusing since it is technically different from directly decaying the weights as pointed out by \citet{loshchilov2018decoupled}.
As previously mentioned, \citet{van2017l2} showed that this interpretation does not hold for modern networks with normalization layers.
Normalization can make a weight vector scale-invariant, meaning that the network output is unaffected by the magnitude of the vector (see \S\ref{sec:preliminaries}).
In this work, we focus on building upon the insight that weight decay serves as a scaling factor for an ``effective'' learning rate, as suggested by previous research \citep{van2017l2,zhang2018three,li2020exponential,wan2021spherical}.
However, other lines of work have explored different facets of understanding and improving weight decay.
\citet{xie2023on} have aimed at mitigating pitfalls of weight decay and improving optimization performance.
They propose a weight decay schedule that ensures better convergence towards the end of training.
Lastly, \citet{andriushchenko2023wd} investigated weight decay's role in regularization.
They specifically, examined how weight decay influences the implicit regularization of SGD noise and the \emph{bias-variance trade-off}.

\subsection{Projected Optimization}
Some existing works remove weight decay and rely on projections of either the updates or weights instead.
AdamP~\citep{heo2021adamp} orthogonalizes the update of the Adam and SGDM optimizers by removing the radial component of $\Delta_g \vomega$.
The main reason for this is to avoid a rapid increase in the weight norm during the initial phases of training.
\citet{zhou2021fixnorm} propose keeping the weight magnitude constant, projecting it onto a sphere after every step and removing the weight decay.
\citet{kodryan2022training} analyze training using projections onto the unit sphere after every optimizer update.
Both of these works consider the union of all scale-invariant weights.
Fixing the norm of this total parameter vector has a similar effect as weight decay and will balance the effective learning rates for each scale-invariant group over time (but does not eliminate the transient phase).
This differs from \citet{huang2017projection} which projects the weights of each neuron individually, which does generally not result in balanced rotation when used with SGD as in their case.
Performing a similar operation with Adam or Lion could result in balanced rotation when the weights of different neurons have identical RMS values. 
\citet{roburin2020spherical} analyzes the spherical projection of the Adam optimization trajectory during standard training.

\subsection{Relative Optimization}
LARS~\citep{you2017lars} and LAMB~\citep{you2019lamb} are variants of SGDM and AdamW that scale the update of a weight to be proportional to its norm (sometimes a clamped version of the weight norm).
They apply this to linear and convolutional layer weights, keeping the original update for weights and biases.
LARS and LAMB were proposed for large batch size training and found to work well there.
Although they are not inspired by the Spherical Motion Dynamics, their form is quite similar to the Rotational Optimizer Variants (\cref{alg:rotational_wrapper}) with a few important differences.
The default form of the RVs is applied filter-wise, centers the weights and allows the update magnitude to vary between steps while keeping the average relative update constant.
The RV also doesn't apply weight decay while LARS and LAMB consider it a part of the update and take into account when scaling the relative update.
Finally, the RVs adjust the learning rate based on the rotational equilibrium value.
This makes it more compatible with the underlying optimizer variants in terms of hyperparameters.
One notable difference is the square root dependency on the relative updates in equilibrium, while LARS and LAMB are directly proportional.
This means that any learning rate schedule for these optimizers is more similar to applying a squared version of this schedule to standard optimizers in equilibrium or the RVs.
This does not fully eliminate the differences however, because changing the schedule also affects gains and biases where the update magnitude is directly proportional to the learning rate for all the optimizers and variants discussed here.

Nero~\citep{liu2021nero} is another optimizer that applies relative updates that are directly proportional to the learning rate and weight magnitude.
Like LARS and LAMB, Nero is not inspired by the neuronal update dynamics of standard optimizers with weight decay, and to the best of our knowledge their relationship has not been pointed out before.
Like the RVs, Nero is applied filter-wise and centers the weights.
Overall, Nero is similar to the SGDM RV without momentum and the hyperparameter mapping, but also applies Adam like updates to the gains and biases, using a separate learning rate.
By making the relative updates directly proportional to the learning rate, it has the same learning rate scheduling differences as LARS and LAMB mentioned above.
Nero lacks momentum which is something that we observed can hurt large batch size training (exploratory experiments not shown).

Instead of controlling the average relative update size, \citet{brock2021high} and \citet{li2022robust} clip the maximum relative update size instead.
The Adaptive Gradient Clipping from \citet{brock2021high} is applied on a per filter basis and is constant throughout training, i.e. does not vary with the learning rate or weight decay.
The clipping introduced in \citet{li2022robust} scales with the learning rate and weight decay in a manner inspired by the equilibrium norm for SGD.
They seem to apply this globally (i.e., not per neuron or layer).

In parallel with this work, \citet{karras2023analyzing} studied diffusion model training and proposed combining neuron-wise Weight Normalization \citep{salimans2016weight}, projections and the Adam algorithm \cite{kingma15adam} for optimization.
They also forgo weight decay, ending up with a training approach that is very similar to our RV-AdamW, showing this significantly improves the optimization of their networks.
The main differences are that they do not use a separate parameter such as the weight decay coefficient to set the relative update size of non-weight parameters like gains and biases (which they remove entirely), do not center the weights, and in their case the rotation is proportional to the learning rate.

\section{Normalization and Scale-Invariance}\label{appx:norm_scale_invariance}
This section provides an overview of normalization and scale-invariance.

\textbf{Setup:} We use Batch Normalization~\citep{ioffe2015batch} as an example of a normalization operation. Let $\vx = \mZ \vomega$ for $\vx \in \R^{B \times 1}$, $\vomega \in \R^{C \times 1}$ and $\mZ \in \R^{B \times C}$ correspond to a single output feature of a linear layer (i.e.\ a neuron).
We can write the batch normalization of this feature as:
\begin{equation}\label{eq:bn-appendix}
    \hat{\vx} = N(\vx) = \frac{\vx - \mu}{\sqrt{\sigma^2 + \varepsilon}},\qquad \mu = \frac{1}{B}\sum_{i=1}^{B} \evx_i, \qquad \sigma^2 = \frac{1}{B}\sum_{i=1}^{B} (\evx_i-\mu)^2
\end{equation}
where $\vx = [\evx_1,\ldots,\evx_B]^\top \in \R^{B}$ is a vector and $\varepsilon \ge 0$ is a small hyperparameter added for numerical stability.
Backpropagation accurately treats $\mu$ and $\sigma$ as functions of $\vx$.
When $\varepsilon$ is sufficiently small to be ignored, the output of the normalization is not affected by a positive scaling of the input:
\begin{equation}
    \textstyle N(r \vx) = (r\vx - r\mu)/\sqrt{r^2\sigma^2 + \varepsilon} \approx (\vx - \mu)/\sqrt{\sigma^2 + \varepsilon} = N(\vx), \qquad r > 0
\end{equation}
If the training loss $\Ls$ does not depend on $\vx$ in other ways than through $N(\vx)$, this makes $\vx$ scale-invariant with respect to the loss, i.e.\ $\Ls(r \vomega)=\Ls(\vomega)$ for $r > 0$.
Note that although we sometimes write $\Ls(\vomega)$ for brevity the loss generally depends on other weights and inputs as well, $\vomega$ is generally only a portion of the parameters used in the network, and could for example be a particular row in the weight matrix of a fully connected layer.
Some normalization operations like Centered Weight Normalization~\citep{huang2017centered} a.k.a.\ Weight Standardization~\citep{qiao2019ws} are performed directly on the weights instead of activations.
This also makes the weight scale-invariant and in case of the aforementioned methods also makes $\nabla_\vomega \Ls(\vomega) \perp \vone$.

\textbf{Properties:} Scale-invariance results in the properties stated in \cref{eq:grad_perp} and \cref{eq:grad_inv}, repeated below:
\begin{talign}
    \text{Gradient orthogonality:} \qquad & \nabla_\vomega \Ls(\vomega) \perp \vomega \label{appx:eq:grad_perp} \\
    \text{Inverse proportionality:}\qquad & \|\nabla_\vomega \Ls(\vomega)\| \propto \|\vomega\|^{-1} \label{appx:eq:grad_inv}
\end{talign}

\textbf{Intuition:} 
The first property is a result of the loss surface being invariant along the direction of $\vomega$.
Hence the directional derivative of $\Ls(\vomega)$ in the direction of $\vomega$ is zero:
\begin{align}\label{eq:grad_perp_intuition}
    \langle \nabla_\vomega \Ls(\vomega), \frac{\vomega}{\|\vomega\|} \rangle &= \lim_{h \to 0} \frac{\Ls(\vomega + h \vomega / \|\vomega\|) - \Ls(\vomega)}{h} \\
    &= \lim_{h \to 0} \frac{\Ls(\vomega) - \Ls(\vomega)}{h} \\
    &= \lim_{h \to 0} \frac{0}{h} = 0
\end{align}
The second property is a result of the backpropagation through $N$, which scales the gradient by the factor used on the forward pass $1/\sqrt{\sigma^2 + \varepsilon} \approx \sigma^{-1}$ as if it were a constant, and the fact that $\sigma \propto \|\vomega\|$.

\textbf{Backpropagation:} The properties can also be shown using expressions for the backpropagation through the normalization layers. For completeness we include the learnable affine transformation that typically follows normalization operations:
\begin{equation}
    \vy = \gamma \hat{\vx} + \beta
\end{equation}

For the backpropagation we have:
\begin{talign}
\nabla_{\gamma} \Ls(\vp) &= \langle \hat{\vx}, \nabla_\vy \Ls(\vp) \rangle \\
\nabla_{\beta} \Ls(\vp) &= \langle \vone_B, \nabla_\vy\Ls(\vp) \rangle \\
\nabla_{\vx} \Ls(\vp) &= \frac{\gamma}{\sqrt{\sigma^2 + \eps}}\cdot\Bigg[\nabla_\vy\Ls(\vp) - \frac{1}{B} \langle \vone_B, \nabla_\vy\Ls(\vp) \rangle \vone_B - \frac{1}{B} \langle \hat{\vx}, \nabla_\vy \Ls(\vp) \rangle \hat{\vx} \Bigg]
\end{talign}
Assuming that $\epsilon$ is small gives:
\begin{equation}
    \textstyle
    \nabla_{\vx} \Ls(\vp) = \frac{\gamma}{\sigma}\Bigg[\nabla_\vy\Ls(\vp) - \frac{1}{B} \langle \vone_B, \nabla_\vy\Ls(\vp) \rangle \vone_B - \frac{1}{B} \langle \hat{\vx}, \nabla_\vy \Ls(\vp) \rangle \hat{\vx} \Bigg] \label{eq:norm_small_eps}
\end{equation}

In this case we have:
\begin{talign}
    \textstyle
    \langle \nabla_{\vx}\Ls(\vp), \vone_B \rangle &=\frac{\gamma}{\sigma} \Bigg[\langle \vone_B, \nabla_\vy\Ls(\vp)\rangle 
    -\frac{1}{B} \langle \vone_B, \nabla_\vy\Ls(\vp) \rangle \underbrace{\langle \vone_B, \vone_B \rangle}_{=B} - \frac{1}{B} \langle \hat{\vx}, \nabla_\vy \Ls(\vp) \rangle \underbrace{\langle \hat{\vx}, \vone_B \rangle}_{=0} \Bigg] \nonumber \\
    & = 0 \label{appx:eq:orthogonality_mu}
\end{talign}
and similarly:
\begin{talign}
    \textstyle
    \langle \nabla_{\vx}\Ls(\vp), \hat{\vx} \rangle &=\frac{\gamma}{\sigma} \Bigg[\langle \hat{\vx}, \nabla_\vy\Ls(\vp)\rangle 
    -\frac{1}{B} \langle \vone_B, \nabla_\vy\Ls(\vp) \rangle \underbrace{\langle \vone_B, \hat{\vx} \rangle}_{=0} - \frac{1}{B} \langle \hat{\vx}, \nabla_\vy \Ls(\vp) \rangle \underbrace{\langle \hat{\vx}, \hat{\vx} \rangle}_{=B} \Bigg] \nonumber \\
    & = 0 \label{appx:eq:orthogonality_nz}
\end{talign}
which gives:
\begin{equation}
    \langle \nabla_{\vx}\Ls(\vp), \vx \rangle = \langle \nabla_{\vx}\Ls(\vp), \sigma \hat{\vx} + \mu \vone_B \rangle = 0
\end{equation}

This allows us to obtain the properties of the weight gradient:
\begin{equation}
    \nabla_\vp \Ls(\vp) = \mZ^\top \nabla_{\vx}\Ls(\vp)
\end{equation}

First we note that:
\begin{equation}
    \|\nabla_\vp \Ls(\vp)\| \propto \| \nabla_{\vx}\Ls(\vp) \| \propto \sigma^{-1} \propto \|\vp\|^{-1}
\end{equation}
where the second proportionality follows from \cref{eq:norm_small_eps} and the final one from \cref{eq:bn-appendix}.
This gives the inverse proportionality listed in \cref{appx:eq:grad_inv}.

We can also derive the gradient orthogonality in \cref{appx:eq:grad_perp} as follows:
\begin{talign}
    \langle \nabla_\vp \Ls(\vp), \vp \rangle &= \langle \mZ^\top \nabla_{\vx}\Ls(\vp), \vp \rangle \\
    &= \vp^\top \mZ^\top \nabla_{\vx}\Ls(\vp) \\
    &= \vx^\top \nabla_{\vx}\Ls(\vp) \\
    &= \langle \nabla_{\vx}\Ls(\vp), \vx \rangle \\
    &= 0
\end{talign}

These properties can also be shown directly from the scale-invariance using calculus theorems as done in \citet{wan2021spherical}.

\section{SGDM Equilibrium}\label{appx:sgdm_equilibrium}
The standard version of SGD with momentum (SGDM) can be written as:
\begin{talign}
    \vm_t &= \alpha \vm_{t-1} + \vg_t + \lambda \vp_{t-1} \\
    \vp_t &= \vp_{t-1} - \eta \cdot \vm_t \label{eq:sgdm_update}
\end{talign}
where $\vp_t \in \R^{C}$ is a parameter vector at time $t$, $\vg_t=\nabla_{\vp} \Ls(\vp_t)$ is the gradient and $\vm$ is the first moment (i.e.\ momentum or velocity).
The learning rate ($\eta \ge 0$), weight decay ($\lambda \ge 0$), and momentum coefficient ($0 < \alpha < 1$) are hyperparameters.

We compute the total update contribution due to $\vg_t$, i.e. $\vu$ in \cref{eq:smd_pythagorean} as:
\begin{equation}\label{eq:sgd_u}
    \textstyle
    \vu = -\eta \sum_{k=t}^{\infty} \alpha^{k-t} \vg_t = \frac{-\eta}{1-\alpha} \vg_t
\end{equation}
Analogously, the total update contribution of the $\ell_2$-regularization of $\vw_{t-1}$, i.e. $\vd$ in \cref{eq:smd_pythagorean}, is:
\begin{equation}\label{eq:sgd_d}
    \textstyle
    \vd = -\eta \sum_{k=t}^{\infty} \alpha^{k-t} \lambda \vp_{t-1} = \frac{-\eta \lambda}{1-\alpha} \vp_{t-1}
\end{equation}

Combining \cref{eq:sgd_u} and \cref{eq:sgd_d}, this allows us to solve \cref{eq:smd_pythagorean} for a scale-invariant weight vector $\vomega$.
Here we assume scale-invariance since it slightly changes the resulting expression due to the dependency of $\|\vu\|$ on $\|\vomega\|$.
It also simplifies the math a bit, with $\vu \perp \vomega$, not just in expectation.
We get:
\begin{talign}
    \smash{\widehat{\|\vomega\|}}^2 &= \E\left[ (\smash{\widehat{\|\vomega\|}} - \|\vd\|)^2 + \|\vu\|^2 \right] \\
    &= \E \left[ \left( \smash{\widehat{\|\vomega\|}} - \frac{\eta\lambda}{1 - \alpha}\smash{\widehat{\|\vomega\|}} \right)^2 + \left( \frac{\eta}{1 - \alpha} \frac{\|\tilde{\vg}\|}{\widehat{\|\vomega\|}} \right)^2 \right]
\end{talign}
Where we define $\tilde{\vg}_t = \vg_t \|\vomega_t\|$ using $\|\vg_t\| \propto \|\vomega_t\|^{-1}$ due to the inverse proportionality of the gradient magnitude, see \cref{eq:grad_inv} or \cref{appx:eq:grad_inv}.
We can interpret $\tilde{\vg}_t$ as the gradient for weights of unit norm $\|\vomega_t\|=1$.

Solving for $\smash{\widehat{\|\vomega\|}}$ and assuming that $\eta \lambda \ll 2 \cdot (1 - \alpha)$ gives:

\begin{equation}
   \smash{\widehat{\|\vomega\|}} = \sqrt[4]{\frac{\eta \E[\|\tilde{\vg}\|^2]}{2\lambda \cdot (1-\alpha)-\eta \lambda^2}} \approx \sqrt[4]{\frac{\eta \E[\|\tilde{\vg}\|^2]}{2\lambda \cdot (1-\alpha)}}
\end{equation}

To obtain the absolute size of an update, we further assume that $\E[\|\vg_t\|^2]$ can be approximated as a constant $\E[\|\vg\|^2]$ when computing the size of $\vm_t$, and that successive gradients are roughly orthogonal giving $\vm_{t-1} \perp \vg_t$ in expectation.
For the random walk setting, the first is reasonable when the norm is stable e.g.\ around equilibrium and the second always holds.
The average square size of an update is then:
\begin{talign}
    \E[\|\Delta_g \vp\|^2] &= \eta^2 \E\left[\left\| \alpha \vm_{t-1} + \vg_t \right\|^2 \right] \label{eq:sgd_deltag_0}\\
    &= \eta^2 \E\left[\left\| \alpha \vm_{t-1}\right\|^2 \right] + \E\left[\left\| \vg_t \right\|^2 \right] \label{eq:sgd_deltag_1}\\
    &= \eta^2 \sum_{k=0}^{t} \E[\left(\alpha^{t-k} \|\vg_k\| \right)^2] \label{eq:sgd_deltag_2}\\ 
    &\approx \eta^2 \frac{\E[\|\vg\|^2]}{1-\alpha^2}\label{eq:sgd_deltag_3}
\end{talign}
where (\ref{eq:sgd_deltag_1}) comes from the orthogonality, (\ref{eq:sgd_deltag_2}) by recursively writing out $\vm$ in terms of $\vg$, and (\ref{eq:sgd_deltag_3}) from assuming that $t$ is high enough to approximate the sum of the geometric series as an infinite sum.

Simplifying, we get the $\eta_g=\sqrt{\E[\|\Delta_g \vp \|^2]}$ and $\eta_r=\sqrt{\E[\|\Delta_g \vomega \|^2]}/\smash{\widehat{\|\vomega\|}}$ in \cref{tab:equilibrium_summary}. We note that the derived rates are consistent with the ones derived in the Spherical Motion Dynamics~\citep{wan2021spherical} and Online Normalization~\citep{chiley2019online} (when $\alpha=0$).

\section{Lion Equilibrium}\label{appx:lion_equilibrium}
The standard version of Lion~\citep{chen2023symbolic} can be written as:
\begin{talign}
    \vv_t &= \sign(\beta_1 \vm_{t-1} + (1 - \beta_1) \vg_t) \\
    \vm_t &= \beta_2 \vm_{t-1} + (1 - \beta_2) \vg_t \\
    \vp_t &= \vp_{t-1} - \eta \cdot (\vv_t + \lambda\vp_{t-1} )\label{eq:lion_update}
\end{talign}
where $\vp_t \in \R^{C}$ is a parameter vector at time $t$, $\vg_t=\nabla_{\vp} \Ls(\vp_t)$ is the gradient, $\vm$ is the first moment and $\vv$ is the update velocity. The learning rate ($\eta \ge 0$), weight decay ($\lambda \ge 0$), and moment coefficients ($0 < \beta_1 < 1, 0 < \beta_2 < 1$) are hyperparameters.

In our analysis we look at the arguments of the sign function which we define as:
\begin{equation}
    \vn_t := \beta_1 \vm_{t-1} + (1 - \beta_1) \vg_t,\qquad \vv_t = \sign(\vn_t)
\end{equation}
To obtain an estimate of the magnitude $\|\vn_t\|$, we assume that the gradient magnitude $\E[\|\vg_t\|^2]$ can be approximated as a constant $\E[\|\vg\|^2]$, and that successive gradients are roughly orthogonal giving $\vm_{t-1} \perp \vg_t$ in expectation.
For the random walk setting, the first is reasonable when the norm is stable e.g.\ around equilibrium and the second always holds.
This gives:
\begin{align}
    \E[\| \vn_t \|^2] &= \E \left[ \| \beta_1 \vm_{t-1} + (1 - \beta_1) \vg_t \|^2 \right] \\
    &= \beta_1^2 \E \left[ \| \beta_2 \vm_{t-1} + (1 - \beta_2) \vg_{t-1} \|^2 \right] + (1 - \beta_1)^2 \E \left[ \|\vg_t\|^2 \right] \\
    &= \beta_1^2 (1 - \beta_2)^2 \sum_{k=0}^{k=t-1} \beta_2^{t-1-k} \E[ \|\vg_k\|^2 ] + (1 - \beta_1)^2 \E \left[ \|\vg_t\|^2 \right] \\
    &\approx \beta_1^2 (1 - \beta_2)^2 \sum_{k=0}^{\infty} \beta_2^{k} \E[ \|\vg\|^2 ] + (1 - \beta_1)^2 \E \left[ \|\vg\|^2 \right] \\
    &= \left( (1 - \beta_1)^2 + \beta_1^2 \frac{1 - \beta_2}{1 + \beta_2} \right) \E[\|\vg\|^2]
\end{align}
where we have used the gradient orthogonality and constant magnitude, and approximated the geometric sum as extending to infinity.

To compute the total update contribution of the gradient, $\|\vu\|$ in \cref{eq:smd_pythagorean}, we first need to model how the sign non-linearity affects the magnitude and direction of the update.
We note that for a parameter of dimension $C$, we have:
\begin{equation}
    \|\vv_t\| = \sqrt{C}
\end{equation}
so the sign function has an average scaling effect:
\begin{equation}
    \frac{\|\vv_t\|}{\|\vn_t\|} = \sqrt{\frac{C}{\left( (1 - \beta_1)^2 + \beta_1^2 \frac{1 - \beta_2}{1 + \beta_2} \right) \E[\|\vg\|^2]}}
\end{equation}
The sign function will also rotate $\vn_t$ resulting in two components, one parallel to $\vn_t$ and the other orthogonal.
We will assume that the orthogonal one cancels out on average without significantly affecting equilibrium and focus on the parallel component.
This component depends on the average angle between $\vn_t$ and $\sign(\vn_t)$ which is determined by the distribution and correlation between the elements.
In the random walk setting, we can assume the components of $\vn_t = [\evn_1,\ldots,\evn_C]$ are normally distributed with mean zero.
However, the expression for the average angle is still complicated unless the components are independent and identically distributed (i.i.d.) so we make this assumption for this step with $\evn_k \sim \mathcal{N}(0, \sigma^2)$ i.i.d.\ for all $k$.
Then we can use the known expected absolute value for a centered normal distribution to get:
\begin{equation}
    \E[ \langle \vn_t, \sign(\vn_t) \rangle ] = C \cdot \E[|\evn_k|] = C \cdot \sqrt{\frac{2 \sigma^2 }{\pi}} = \sqrt{\frac{2 }{\pi}} \cdot \|\vn_t\| \cdot \|\sign(\vn_t)\|
\end{equation}
Note that the angle is still bounded regardless of the distribution but will result in a different factor in the range that $\|\vn\|_1/(\sqrt{C} \|\vn\|_2)$ can take, i.e.\ $[C^{-\frac{1}{2}},1]$ instead of $\sqrt{2/\pi}$.

Based on the preceding analysis we will model the sign function for the computation of $\|\vu\|$ as:
\begin{equation}
    \sign (\vn_t) \approx \sqrt{\frac{2}{\pi}} \frac{\|\vv_t\|}{\|\vn_t\|} \vn_t = \sqrt{\frac{2C}{\E[\|\vg\|^2] \cdot \pi \cdot \left( (1 - \beta_1)^2 + \beta_1^2 \frac{1 - \beta_2}{1 + \beta_2} \right)}} \vn_t
\end{equation}
which gives:
\begin{align}
    \E[\|\vu\|^2] &= \frac{2 \eta C}{\E[\|\vg\|^2] \cdot \pi \cdot \left( (1 - \beta_1)^2 + \beta_1^2 \frac{1 - \beta_2}{1 + \beta_2} \right)} \E\left[ \left\|(1-\beta_1) \vg + \beta_1 (1-\beta_2) \sum_{k=0}^{\infty} \beta_2^k \vg \right\|^2 \right] \\
    &= \frac{2 \eta C}{\pi \cdot \left( (1 - \beta_1)^2 + \beta_1^2 \frac{1 - \beta_2}{1 + \beta_2} \right)}
\end{align}

Combined with the total update contribution for the weight decay, $\vd=-\eta \lambda \vomega_{t-1}$, this allows us to write \cref{eq:smd_pythagorean} for $\vomega \in \R^C$:
\begin{align}
    \smash{\widehat{\|\vomega\|}}^2 &= \E\left[ (\smash{\widehat{\|\vomega\|}} - \|\vd\|)^2 + \|\vu\|^2 \right] \\
    &= (\smash{\widehat{\|\vomega\|}} - \eta \lambda \smash{\widehat{\|\vomega\|}})^2 + \E[\|\vu\|^2] \\
    &= (1 - 2\eta \lambda + \eta^2 \lambda^2) \smash{\widehat{\|\vomega\|}}^2 + \frac{2 \eta C}{\pi} \left( (1 - \beta_1)^2 + \beta_1^2 \frac{1 - \beta_2}{1 + \beta_2} \right)^{-1}
\end{align}
Solving for $\smash{\widehat{\|\vomega\|}}$ and assuming $\eta\lambda \ll 2$ gives:
\begin{equation}
    \textstyle
    \smash{\widehat{\|\vomega\|}} = \sqrt{\frac{2 \eta C}{\pi \cdot (2\lambda - \eta \lambda^2)}} \left( (1 - \beta_1)^2 + \beta_1^2 \frac{1 - \beta_2}{1 + \beta_2} \right)^{-\frac{1}{2}} \approx \sqrt{\frac{\eta C}{\pi \lambda}} \left( (1 - \beta_1)^2 + \beta_1^2 \frac{1 - \beta_2}{1 + \beta_2} \right)^{-\frac{1}{2}}
\end{equation}

Combined with $\|\vv_t\| = \sqrt{C}$ for $\vomega,\vp \in \R^C$ we get the expected equilibrium rotation and RMS update size:
\begin{align}
    \widehat{\eta_g} &= \sqrt{\E[\|\Delta_g \vp\|^2]} = \eta \sqrt{C} \\
    \widehat{\eta_r} &= {\sqrt{\E[\|\Delta_g \vomega \|^2]}}/{\widehat{\|\vomega\|}} = \sqrt{\pi \eta \lambda} \cdot \left((1-\beta_1)^2 + \beta_1^2 \frac{1-\beta_2}{1+\beta_2}\right)^{\frac{1}{2}}
\end{align}

\section{Adam+\texorpdfstring{$\ell_2$}{L2} Equilibrium}\label{appx:adam_l2}
In this section we apply a modified form of the geometric model from \cref{sec:equilibrium_geometric_model} to Adam~\citep{kingma15adam} with $\ell_2$-regularization (Adam+$\ell_2$ for short) to gain insight into how the rotational equilibrium differs from that of Adam with decoupled weight decay (AdamW, see \cref{sec:adamw_equilibrium}).

\subsection{Adam+\texorpdfstring{$\ell_2$}{L2} Formulation}
We will write the Adam+$\ell_2$ update as follows:
\begin{talign}
    \vm_t &= \beta_1 \vm_{t-1} + (1 - \beta_1) (\vg_t + \lambda \vp_{t-1}) \\
    \vv_t &= \beta_2 \vv_{t-1} + (1 - \beta_2) (\vg_t + \lambda \vp_{t-1})^2 \\
    \vp_t &= \vp_{t-1} - \eta \cdot \Big( \frac{\vm_t/(1-\beta_1^t)}{\sqrt{\vv_t/(1-\beta_2^t)} + \epsilon} \Big) \label{eq:adam_update}
\end{talign}
Similar to AdamW, $\vp_t \in \R^{C}$ is a parameter vector at time $t$, $\vg_t=\nabla_{\vp} \Ls(\vp_t)$ is the gradient, and all operations (e.g. division, squaring) are performed elementwise.
In Adam+$\ell_2$, both the first and second moment of the gradient ($\vm$ and $\vv$) include contributions from the $\ell_2$-regularization term.
This differs from AdamW (see \cref{eq:adamw_update}) where the $\ell_2$-regularization (or technically weight decay) does not affect $\vm$ and $\vv$.
The learning rate ($\eta \ge 0$), $\ell_2$-regularization coefficient ($\lambda \ge 0$), moment coefficients ($0 < \beta_1 < 1, 0 < \beta_2 < 1$) and $\epsilon \ge 0$ are hyperparameters similar to AdamW.
Like before, we use $\vomega$ to specifically denote the weight vector of a neuron or a layer that can form rotational equilibrium (as opposed to $\vp$ that we use as a general symbol for any parameter vector, and could denote e.g.\ a bias).

\subsection{Simplifications}

The rotational dynamics of Adam+$\ell_2$ are more complicated than those of AdamW.
The main difference is that the strength of the ``weight decay'' is affected by the gradient norm.
As we will see, this makes the equilibrium norm and angular update depend on the gradient magnitude.
Furthermore, the weight decay can be scaled differently for each coordinate of the weight vector as the gradient distribution may vary between them.
This complicates the analysis, forcing us to treat each coordinate separately.

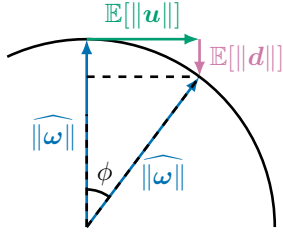
\begin{figure}[tb]
\centering
\begin{tikzpicture}[
  arrow_blue/.style={tabblue, -{Latex[length=2mm]}, line width=1.0pt, opacity=1.0},
  arrow_green/.style={tabgreen, -{Latex[length=2mm]}, line width=1.0pt, opacity=1.0},
  arrow_red/.style={tabred, -{Latex[length=2mm]}, line width=1.0pt, opacity=1.0},
  triangle/.style={black, line width=1.0pt, dashed, opacity=1.0},
  scale=0.5, transform shape, every node/.style={scale=2.0}
]
  \draw[line width=1.0pt] (5,0cm) arc (0:115:5cm);

  \pgfmathsetmacro{\AngleSpan}{90-atan(3/4)} %
  \pgfmathsetmacro{\LabelAngle}{90-atan(3/4)/2} %
  \draw[line width=1.0pt] (0.6,0.8cm) arc (\AngleSpan:90:1cm);

  \node at (\LabelAngle:1.5cm) {$\phi$};

  \draw[arrow_blue] (0,0) -- (0,5cm) node[midway,left] {$\widehat{\|\vomega\|}$};
  \draw[arrow_blue] (0,0) -- (3,4cm) node[midway,right,xshift=-5pt,yshift=-8pt] {$\widehat{\|\vomega\|}$};

  \draw[arrow_green] (0,5cm) -- ++(0:3cm) node[midway,above] {$\E[\|\vu\|]$};

  \draw[arrow_red] (3,5cm) -- (3,4cm) node[midway,right] {$\E[\|\vd\|]$};

  \draw[triangle] (0,4cm) -- (3,4cm);
  \draw[triangle] (0,0cm) -- (3,4cm);
  \draw[triangle] (0,0cm) -- (0,3cm);

\end{tikzpicture}
\caption{Weight norm equilibrium. The total update contribution of the gradient is $\vu$ and the TUC of the weight decay is $\vd$.}
\label{fig:smd_triangle_ud_angle}
\end{figure}

Our analysis is based on the random walk setup introduced in \cref{sec:analysis} and described in \cref{appx:real_vs_random_walk}.
We further make several assumptions and simplifying approximations that allow us to obtain simpler expressions for the cases of interest:
\begin{enumerate}
    \item We assume the rotational equilibrium exists as a steady state where hyperparameters are fixed (not varying over time), the expected weight norm $\sqrt{\E[\|\vomega_t\|^2]} = \inlinevomegaeqnorm$ is constant, and the second moment of the gradient $\E[\vg_t^2]$ is constant over time. For simplicity we will drop the $t$ subscript.
    \item We focus on the case where the weights are scale-invariant, defining $\tilde{\vg} = \|\vomega\| \vg$ as the gradient corresponding to a unit weight norm based on the inverse proportionality from \cref{eq:grad_inv}.
    \item We will assume that $\epsilon$ and the bias correction can be ignored, i.e.\ that $\epsilon$, $\beta_1^t$ and $\beta_2^t$ are all effectively zero.
    \item We will assume the second moment tracker $\vv_t$ is dominated by the gradient component, i.e. that $\vg^2 \gg \lambda^2 \vomega^2$, and that it perfectly tracks the expected value, i.e. that $\E[\vv]=\E[\vg^2]$.
    This is a non-trivial approximation based on the geometry of equilibrium when the angular updates are small. For a small $\phi$ in \cref{fig:smd_triangle_ud_angle} we can can approximate:
    \begin{align}
        \E[\|\vu\|] &= \widehat{\|\vomega\|}\cdot\tan(\phi)\approx\widehat{\|\vomega\|}\cdot \phi \\
        \E[\|\vd\|] &= \widehat{\|\vomega\|} \cdot (1-\cos(\phi)) \approx \widehat{\|\vomega\|} \cdot \frac{\phi^2}{2}
    \end{align}
    As a result $\E[\|\vu\|] \gg \E[\|\vd\|]$.
    For Adam+$\ell_2$ we have ${\vu=-\eta \vg_t/\sqrt{\vv}}$ and ${\vd = -\eta \lambda \vomega_{t-1}/\sqrt{\vv}}$.
    As long as $\vv$ is relatively homogeneous across coordinates, we therefore have ${\E[\|\vg\|] \gg \E[\lambda\|\vomega\|]}$.
    We assume this holds roughly coordinate wise as well, giving ${\vg^2 \gg \lambda^2 \vomega^2}$.
    We note that this fourth assumption is not strictly necessary but significantly simplifies the resulting expressions, giving us an interpretable closed form solution instead a solution expressed as the root of a third-degree polynomial.
\end{enumerate}

\subsection{Equilibrium Norm}

The total update contribution of the gradient $\vg_t$, i.e.\ $\vu$ in \cref{eq:smd_pythagorean}, is given by:
\begin{equation}
    \textstyle
    \vu = - \eta  \sum_{k=t}^{\infty} \beta_1^{k-t} (1 - \beta_1) \frac{\vg_t}{\sqrt{\vv}} = -\eta \frac{\vg_t}{\sqrt{\vv}}
\end{equation}
Similarly, the total update contribution due to the weight decay of $\vomega_{t-1}$ is:
\begin{equation}
    \vd=-\eta \sum_{k=t}^{\infty} \beta_1^{k-t} (1 - \beta_1) \frac{\lambda \vomega_{t-1}}{\sqrt{\vv}} = -\eta \frac{\lambda \vomega_{t-1}}{\sqrt{\vv}}
\end{equation}

Due to the coordinate-wise differences in the weight decay, we analyze a single element $\evomega_k$ at coordinate $k$ in $\vomega$ with corresponding elements $\evd_k$, $\evu_k$, $\evg_k$, $\evv_k$ in $\vd$, $\vu$, $\vg$, $\vv$, respectively.
Although the geometric model is not well defined coordinate-wise, we can still use the concept of orthogonality as defined for random variables.
This gives us:
\begin{talign}
    \E[\evomega_k^2] &= \E[(\evomega_k - \evd_k)^2 + \evu_k^2] \\
    &= (1 - \frac{\eta \lambda}{\sqrt{\evv_k}})^2 \E[\evomega_k^2] + \eta^2 \frac{\E[\evg_k^2]}{\evv_k} \\
    &= (1 - \frac{2 \eta \lambda}{\sqrt{\evv_k}} + \frac{\eta^2 \lambda^2}{\evv_k}) \E[\evomega_k^2] + \eta^2 (1 - \frac{\lambda^2 \E[\evomega_k^2]}{\evv_k}) \\
    &= (1 - \frac{2 \eta \lambda}{\sqrt{\evv_k}}) \E[\evomega_k^2] + \eta^2 \label{eq:adaml2_cw_omega_triangle}
\end{talign}
where we have used the fact that $\evv_k = \E[\evg_k^2] + \lambda^2 \E[\evomega_k^2]$.

Since we are targeting the scale-invariant case (Assumption~2) we can write:
\begin{equation}
    \E[\evg_k^2] = \E[\tilde{\evg}_k^2] \E[\|\vomega\|^2]^{-1}
\end{equation}
where $\tilde{\evg}_k$ corresponds to an element of the unit norm weight gradient $\tilde{\vg}$.
Accordingly we can write:
\begin{equation}
    \evv_k = \E[\tilde{\evg}_k^2] \E[\|\vomega\|^2]^{-1} + \lambda^2 \E[\evomega_k^2] \approx \E[\tilde{\evg}_k^2] \E[\|\vomega\|^2]^{-1}
\end{equation}
where we used Assumption~4.

Plugging this form of $\evv$ into \cref{eq:adaml2_cw_omega_triangle}, squaring and simplifying gives:
\begin{equation}
    \E[\evomega_k^2] = \frac{\eta}{2 \lambda} \sqrt{\frac{\E[\tilde{\evg}_k^2]}{ \E[\|\vomega\|^2]}}
\end{equation}

We can now write an expression for the equilibrium norm:
\begin{equation}
    \widehat{\|\vomega\|}^2 = \E[\|\vomega\|^2] = \sum_{k=1}^{C} \E[\evomega_k^2] = \frac{\eta }{2 \lambda} \sum_{k=1}^{C} \sqrt{\frac{\E[\tilde{\evg}_k^2]}{ \E[\|\vomega\|^2]}}
\end{equation}
which gives:
\begin{equation}
    \widehat{\|\vomega\|} = \sqrt[3]{\frac{\eta}{2 \lambda}\!\cdot\! 
    \langle \vone, \sqrt{\E[\tilde{\vg}^2]} \rangle}
\end{equation}
where $\langle\cdot,\cdot\rangle$ denotes an inner product.
Note that when the elements of $\vg$ have the same second moment, e.g. when they are identically distributed, we can write $\langle \vone, \sqrt{\E[\tilde{\vg}^2]} \rangle = \sqrt{C} \sqrt{\E[\|\tilde{\vg}\|^2]}$.
Also note how this behavior differs from that of AdamW, here the equilibrium norm depends on the gradient magnitude.
Finally we note that without scale-invariance we would get a square root instead of a cube root.

\subsection{Equilibrium Angular Update}
To obtain the absolute size of an update for $\vomega$ in equilibrium, we use the fact that in the random walk successive gradients are orthogonal in expectation.

Similar to AdamW, we can then write the average square size of an update as:
\begin{talign}
    \E[\|\Delta_g \vomega\|^2] &= \eta^2 (1-\beta_1)^2 \sum_{k=0}^{t} \beta_1^{2t - 2k} \E\left[\left\| \frac{\vg_k}{\vv} \right\|^2 \right] \\
    &\approx \eta^2 \frac{1-\beta_1}{1+\beta_1} C \label{eq:adam_deltag}
\end{talign}
where we approximated the geometric sum with its limit and used $\vv \approx \E[\vg^2]$ based on Assumption~4.
Note that the use of Assumption~4 gives the same result as for AdamW.

We can then approximate the expected angular update in equilibrium as:
\begin{equation}
    \widehat{\eta_r} = \frac{\sqrt{\E[\|\Delta_g \vomega\|^2]}}{\inlinevomegaeqnorm} = \sqrt[3]{\frac{2 \eta^2 \lambda}{\langle \vone, \sqrt{\E[\tilde{\vg}^2]} \rangle}} \sqrt{\frac{1-\beta_1}{1+\beta_1}C}
\end{equation}

Note that the average angular update depends on the gradient magnitude unlike for other optimizers.
Also note the different dependency on $\eta$ and $\lambda$, here the angular update depends on the product $\eta^2 \lambda$, not $\eta \lambda$ like for other optimizers.
This pattern is visible in the hyperparameter heatmap seen in \cref{fig:adam_vs_adamw_rn18}, the performance varies faster along the direction of increasing $\widehat{\eta_r}$ than where it is constant.
Finally there is an odd dependency on $C$ that is not present in the other optimizers.
Without scale-invariance, the first cube root would be replaced by a square root and the gradient dependency on $C$ would cancel the $C$ in the second root.

\section{Random Walk Experiments}\label{appx:simple_system}
\subsection{Measurements of a Random Walk}
We can directly perform a random walk to validate the expressions given in \cref{tab:equilibrium_summary}.
\Cref{fig:optimizer_prediction_overview} shows measurements for a random walk in a simple system described below for each of the metrics given in the table.
We show four neurons (colored solid lines), the average over a layer (black) and the predicted equilibrium value (red dashed line) from \cref{tab:equilibrium_summary}.
As we can see the analytically derived expressions accurately describe the neuronal dynamics of the random walk for each optimizer.
We use reasonable hyperparameters values in each case that we have found to work well for either ResNet-18 or ResNet-20 on CIFAR-10 (see details in the next section).

\subsection{Simple System for Random Walks}

\begin{figure*}[tb]
    \centering
    \vspace{-6pt} %
    \includegraphics[width=\textwidth]{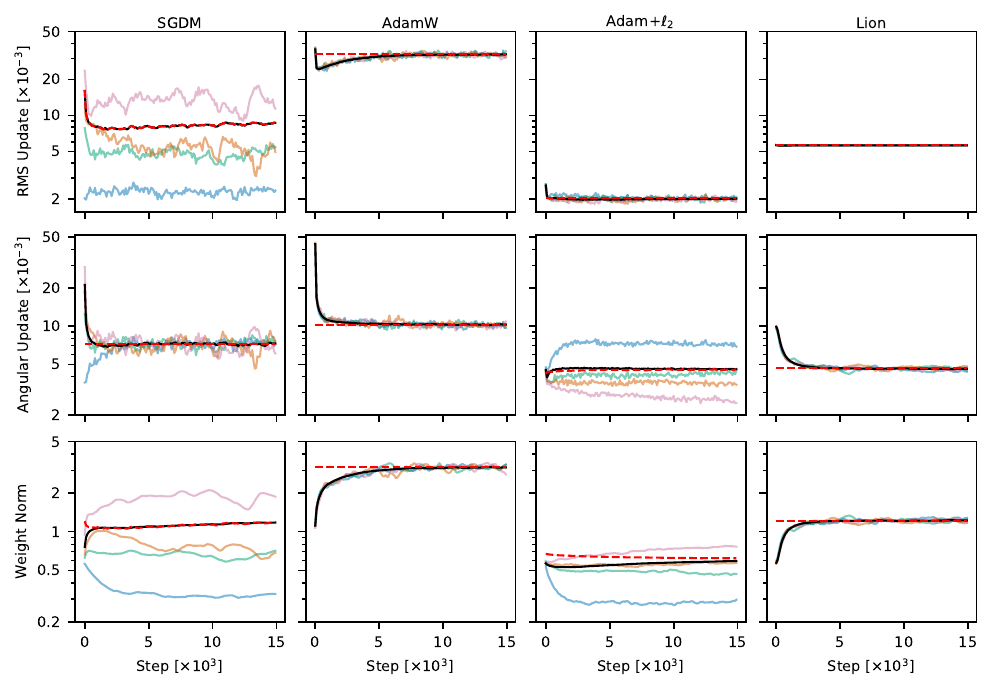}
    \vspace{-6pt} %
    \caption{
    Measurements of the neuronal update dynamics in a random walk, comparing them with the our analytical equilibrium predictions.
    Colors (pink, orange, green, blue) correspond to individual neurons with varying gradient norms from a single layer. Black lines represent averages over the whole layer, and red dashed lines show equilibrium predictions for the layer average from \cref{tab:equilibrium_summary}. The predictions accurately describe the steady state value in each case.
    }
    \label{fig:optimizer_prediction_overview}
    \vspace{-10pt}
\end{figure*}

\textbf{Definition:} We define the simple system as:
\begin{equation}
    f(\mX) = \vgamma_{\mathrm{out}} \odot N\big(\mW (\vgamma_{\mathrm{in}} \odot \mX) \big)
\end{equation}
where $\mX \in \R^{C \times B}$, $\mW \in \R^{K \times C}$, $\vgamma_{\mathrm{in}} \in \R^{C \times 1}$, $\vgamma_{\mathrm{out}} \in \R^{K \times 1}$ and $N$ is a batch normalization function (see \cref{eq:bn-appendix}) applied to each feature independently.
The only learnable parameters are the weights $\mW$, the gammas $\vgamma_{\mathrm{in}}$ and $\vgamma_{\mathrm{out}}$ are kept constant.
We initialize the weights using the default initialization for a linear layer in PyTorch~\citep{paszke2019pytorch} i.e.\ each element is sampled independently and uniformly from the interval $[-\frac{1}{\sqrt{C}}, \frac{1}{\sqrt{C}}]$.
The gammas are initialized with elements independent and identically distributed (i.i.d.) following a standard normal distribution.
The inputs are also sampled i.i.d.\ from a standard normal distribution at each iteration. 
The gradients of $f(\mX)$, which are used to compute other gradients via the chain-rule or backpropagation, are sampled i.i.d.\ from a normal distribution with standard deviation $\frac{1}{KB}$ where the $B$ simulates the typical averaging of the loss over a batch and the $K$ gives a scale more similar to the derivatives of softmax-cross-entropy (the difference of two vectors with an $L_1$-norm of 1 each).
We can also scale the output gradients (that get backpropagated to compute the parameter gradients) further with a loss scale to obtain different gradient norms (especially important for Adam).

\textbf{Rationale:} We use this system to study a random walk in a neural network as described in \cref{sec:analysis}, which serves as a simplified model of a real optimization problem.
The gammas give different variances for each input and output channel, causing the second gradient moment in Adam/AdamW to vary between elements of $\mW$ like they may in real neural network training due to the rest of the network.
The normalization ensures that the system is scale-invariant for each row of $\mW$.
The randomly sampled inputs and output gradients ensure that everything is orthogonal in expectation.
Compared to a real neural network training, the dynamics of this system are simplified with no loss converging over time and steady input / gradient distributions.
Other complicated effects such as dead ReLUs do also not happen in this system.
This makes this simple system a good setting to study the equilibrium dynamics in a controlled manner.

\textbf{Details of \Cref{fig:optimizer_prediction_overview}:} Here we use $B=32, C=K=128$.
We use the CIFAR-10 ResNet-20 hyperparameters from \cref{appx:tab:constraining_smd} for SGDM and Lion (batch size 128), and the optimal configuration from the learning rate, weight decay sweep on CIFAR-10 ResNet-18 for AdamW and Adam+$\ell_2$ (\cref{fig:adam_vs_adamw_rn18}L, see details for that experiment in \cref{appx:experimental_details}).
The learning rate is constant and the experiments run for 15k steps, with the plots downsampled by a factor of 100x using RMS averaging.

\section{Differences between Real Networks and a Random Walk}\label{appx:real_vs_random_walk}
\begin{figure}[bth]
\includegraphics[width=\linewidth]{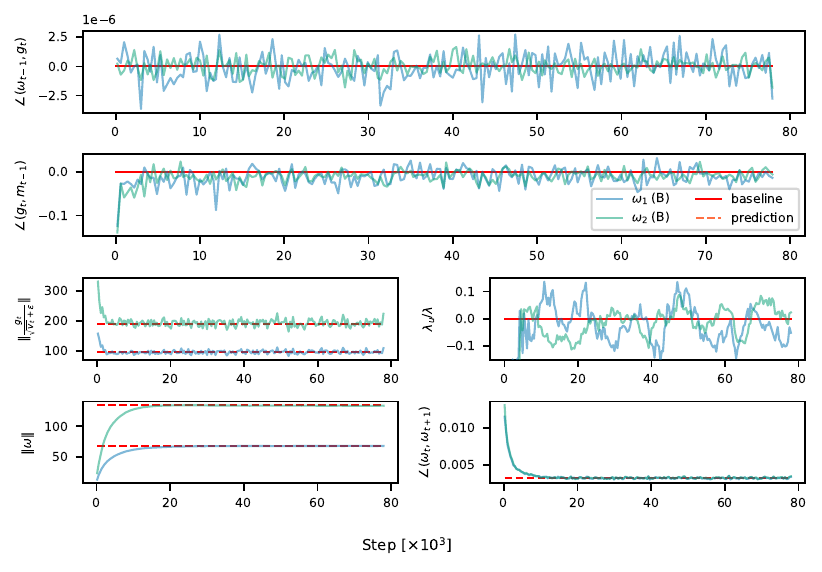}
\caption{
Measurement of how closely the random walk model approximates the training dynamics of two convolutional filters in a ResNet-20 trained on CIFAR-10 using AdamW, with a constant learning rate of $\eta = 0.01$ and a weight decay of $\lambda = 0.01$.
This standard ResNet employs Batch Normalization after each convolutional layer, ensuring scale-invariant convolutional weights.
As expected, $\E[\langle \vu, \vomega \rangle]$ and $\lambda_u = 0$ are close to zero.
Similarly, $\angle(g_{t}, m_{t-1})$ is nearly zero, indicating that the approximation $\forall j \ne k: \E[\langle \vg_j, \vg_k \rangle] = 0$ holds well in practice. Additionally, the approximation $\forall t,k: \E[\|\vg_t / \sqrt{\vv_k}\|] = \sqrt{C}$ appears to hold in this case.
Finally, we observe that the predictions closely match the measurements after the initial transient phase in this setup.}
\label{appx:fig:smd_cifar10}
\end{figure}

\begin{figure}[tbh]
    \centering
    \includegraphics[width=\textwidth,keepaspectratio]{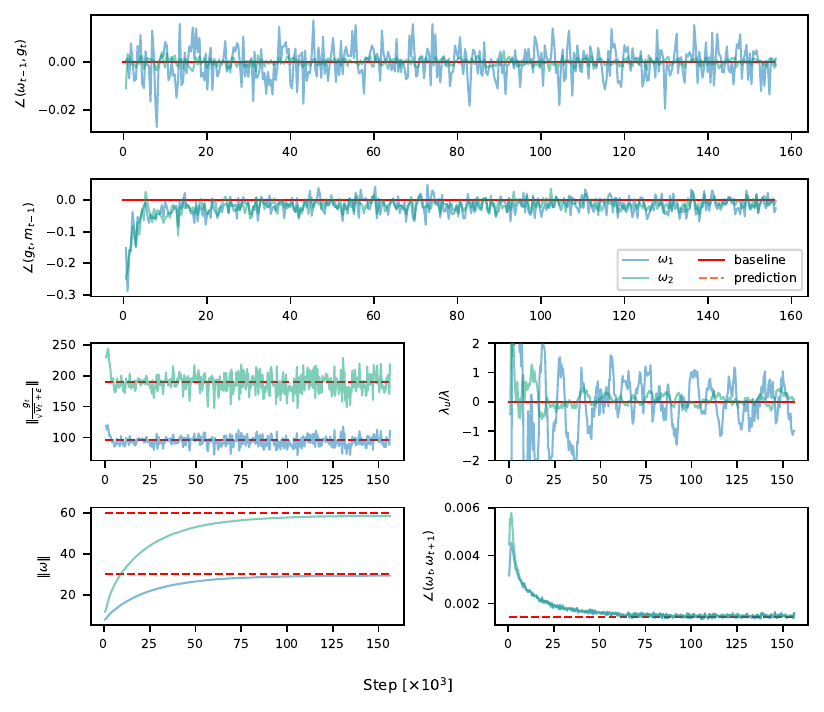}
    \caption{
    Measurement of how closely the random walk model approximates the training dynamics of two convolutional filters in an unnormalized ResNet-20 trained on CIFAR-10 using AdamW, with a constant learning rate of $\eta = 0.002$ and a weight decay of $\lambda = 0.01$. We observe similar overall behavior as in \cref{appx:fig:smd_cifar10}, but the gradients are not necessarily fully orthogonal to the weights as we would expect. There is a slight alignment between the gradients and weights, causing the effective weight decay to change.  Nevertheless, we observe that the predictions closely match the measurements after the initial transient phase in this setup.}
    \label{appx:fig:smd_unnorm_cifar10}
\end{figure}

\begin{figure}[tbh]
    \centering
    \includegraphics[width=\textwidth,keepaspectratio]{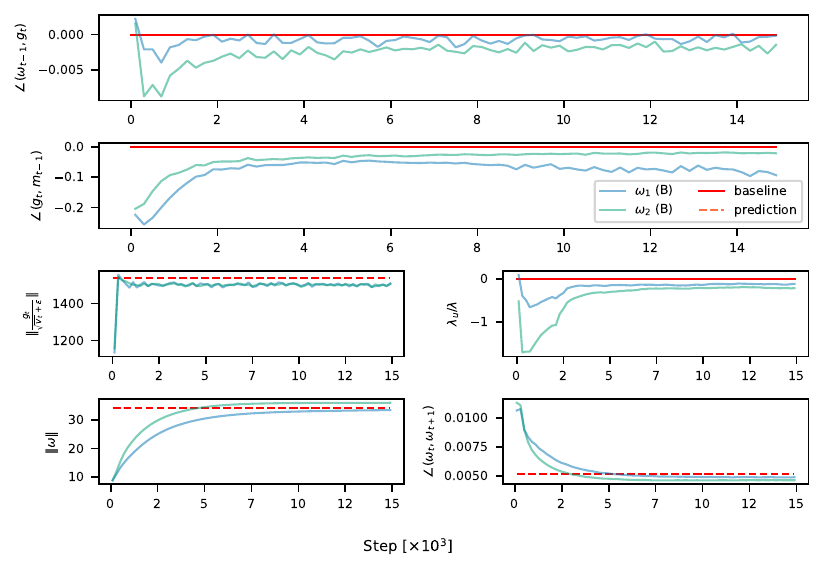}
    \caption{
     Measurement of how closely the random walk model approximates the training dynamics for a GPT2-124M model trained on Wikitext using AdamW, with a constant learning rate of $\eta = 0.0005$ and a weight decay of $\lambda = 0.5$.
     In this model, the weights are not fully scale-invariant.
     We observe a small but consistent negative alignment between the gradients and weights, resulting in a reduced weight decay effect. Additionally, there is a slight overestimation of the scaled gradient and a consistent negative alignment between the gradients and momentum. Despite these deviations, the random walk approximations yield reasonable predictions, as evidenced by the measurements in the last row.}
    \label{appx:fig:smd_llm}
\end{figure}

The random walk model we use assumes the gradient is dominated by the noise.
As discussed by \citet{mccandlish2018empirical}, the mini-batch gradient noise is inversely proportional to the batch size.
At the so called critical batch size, the noise term has a magnitude roughly equal to the true gradient.
For sufficiently small batch sizes relative to the critical batch size, which is often large in practice \citep{mccandlish2018empirical,shallue2019measuring}, the noise term thus dominates the mini-batch gradient.
In this case, the optimization trajectory will locally look like a random walk, but its properties can change over time.
Our analysis focuses on the simplified setting of a fully random walk, disregarding the influence of the true gradient on its dynamic behavior. This simplification allows us to make various assumptions that would not strictly hold for real neural network optimizations. Consequently, for real network optimization, we make several approximations that can affect the accuracy of the predictions in \cref{tab:equilibrium_summary}.
Here we evaluate how well the main simplifications from the random walk analysis of the AdamW optimizer in \cref{sec:analysis} apply to real neural network training tasks.
Specifically, we cover a standard and unnormalized CNN trained on CIFAR-10 (\cref{appx:fig:smd_cifar10} and \cref{appx:fig:smd_unnorm_cifar10}) and a Transformer model trained on Wikitext (\cref{appx:fig:smd_llm}).
Below we discuss the key quantities we approximate and how this affects the predicted equilibrium norm and rotation.

\textbf{Orthogonality Between the Weights and the Gradient $\angle(\omega_{t-1},\vg_{t})$:}
We use the orthogonality between the weights and the gradient as a measure of $\E[\langle \vu,\vomega \rangle]=0$.
If we measure a small positive or negative bias, the gradient contributes to the effective weight decay $\lambda_e$.
When the gradient $\vg_{t}$ is aligned with the weight $\omega_{t-1}$, it increases the weight decay.
Conversely, when the gradient $\vg_{t}$ is negatively aligned with the weight $\omega_{t-1}$ the weight decay decreases.

As a consequence, we tend to overestimate (or underestimate) the measured weight norm  $\|\vomega\|$ and underestimate (or overestimate) the expected angular update $\eta_r$.

By defining the error as a scaling factor of $\lambda$ (represented as $\lambda_{\textit{err}} = \frac{\lambda_e}{\lambda}$), we observe the following impact on our prediction
\begin{align}
    \label{appx:eq:orth_wg1} \eta_r &\approx  \sqrt{2 \eta (\lambda \cdot \lambda_{\textit{err}}) \frac{1-\beta_1}{1 + \beta_1}} =  \widehat{\eta}_r \cdot \sqrt{\lambda_{\textit{err}}}\\
    \label{appx:eq:orth_wg2}\|\vomega\| &\approx \sqrt{\frac{\eta C}{2(\lambda \cdot \lambda_{\textit{err}})}} = \widehat{\|\vomega\|} \cdot  \frac{1}{\sqrt{\lambda_{\textit{err}}}}
\end{align}

\textbf{Orthogonality Between the Momentum and the Gradient $\angle(g_{t},m_{t-1})$:}
We use the orthogonality between the momentum and the  gradient as an approximation of $\forall j \ne k: \E[\langle \vg_j,\vg_k \rangle]=0$.
It gives us information about the orthogonality of $\vg_t$ and the previous update directions. This means we have additional negative (or positive) terms in \cref{eq:rms_update2} and thus tend to overestimate (or underestimate) the approximated RMS update size $\eta_g$ in \cref{eq:rms_update3}.

\textbf{Scaled Gradient Norm $\|\frac{g_t}{\sqrt{v_t} + \epsilon}\|$:}
The scaled norm directly measures the assumption $\forall t,k: \E[\|\vg_t / \sqrt{\vv_k}\|] = \sqrt{C}$. If $\E[\|\vg_t / \sqrt{\vv_k}\|]$ is larger than $\sqrt{C}$, we underestimate the weight norm. If it is smaller, we overestimate the weight norm.

For an error $C_{\textit{err}}$ in the estimate, defined as $ \E[\|\vg_t / \sqrt{\vv_k}\|] = C \cdot C_{\textit{err}}$, we have:
\begin{align}
    \|\vomega\| &\approx \sqrt{\frac{\eta (C \cdot C_{\textit{err}})}{2\lambda}} = \widehat{\|\vomega\|}\cdot \sqrt{C_{\textit{err}}}
\end{align}

\textbf{Radial Gradient Component $\lambda_u / \lambda$:}
The radial gradient component approximates the impact of $\E[\vu_\parallel]$ relative to $\lambda$. It measures how much the scaled gradient component influences the effective weight decay $\lambda_e = \lambda + \lambda_u$. A large $\lambda_u$ affects our predictions similarly to what is described in \cref{appx:eq:orth_wg1} and \cref{appx:eq:orth_wg2}.
\vspace{30pt}

\FloatBarrier

\section{Rotational Dynamics of Scale-Sensitive 
Parameters}\label{appx:scale_sensitive_smd}

Most neural network architectures have some scale-sensitive parameters.
This commonly includes gains and biases as well as a final fully connected (FC) layer that is typically not followed by normalization.
In networks without normalization, with infrequent normalization, or poorly placed normalization, most weight vectors can be scale-sensitive.
The original, un-normalized, VGG~\citep{simonyan2014very} architecture is a good example of this.
VGG consists of a series of convolutional layers, with ReLUs and occasional pooling layers between them, and series of fully connected layers towards the end.
In this section we use it to investigate the rotational dynamics of scale-sensitive weights.

First we would like to note that the weight and gradient magnitude of scale-sensitive weights can also be largely arbitrary, similar to scale-invariant weights.
Although they can't be scaled directly without affecting the loss, we can often scale two of them without affecting the network output.
Consider two successive layers with a ReLU between them:
\begin{equation}
    f(\mX, \mW_1, \mW_2, \vb_1, \vb_2) = \text{ReLU}(\mX \mW_1 + \vb_1)\mW_2 + \vb_2
\end{equation}
where $\mW_1, \mW_2 \in \R^{C \times C}$ are weight matrices, $\vb_1, \vb_2 \in \R^{1 \times C}$ are vectors, $\mX \in \R^{B \times C}$ are inputs and we broadcast the operations.
Note that the ReLU is positively homogeneous, so for a positive scalar $r > 0$ we have:
\begin{equation}
    f(\mX, r\mW_1, r^{-1}\mW_2, r\vb_1, \vb_2) = \text{ReLU}(\mX \mW_1 r + \vb_1 r)\mW_2 r^{-1} + \vb_2 = f(\mX, \mW_1, \mW_2, \vb_1, \vb_2)
\end{equation}
Assuming the weights are scaled in-place (i.e.\ we don't modify the computation graph, only the weight values), this type of rescaling operation scales the relative update of $\mW_1$~by~$r^{-2}$ and $\mW_2$~by~$r^2$ when optimizing using SGD.
This can significantly affect the learning dynamics as studied in e.g.\ Path-SGD~\citep{neyshabur2015path}.

For a scale-sensitive weight $\vomega$, the gradient orthogonality \cref{eq:grad_perp} and inverse scaling \cref{eq:grad_inv} do not necessarily hold.
The inverse scaling holds in terms of rescaling operations like the ones mentioned above if they are applicable.
Generally, the gradient has some radial component in the direction of the weight.
The expected magnitude of this component depends on the average angle between the gradient and the weight as well as the expected gradient magnitude itself.
If we separate the gradient into radial and perpendicular components and view the radial component as a modification of the weights decay, we have a very similar setup to the one we analyzed for scale-invariant weights.
If a stable equilibrium exists, this could give rise to rotational dynamics which may vary from weight to weight based on the ``effective weight decay'' for each one.

We explore this with VGG-13 training on CIFAR-10 using SGDM.
We compare two versions, a standard unnormalized network and a variant where weight standardization is applied to every convolutional and fully connected layer.
For each setup, we measure the angular updates, the weight norms, and the relative radial gradient magnitude:
\begin{equation}
    \lambda_u = \E[\langle \vomega, \nabla_\vomega \mathscr{L} \rangle / \|\vomega\|^2] = E[\cos\left(\angle \left( \vomega, \nabla_\vomega \mathscr{L}\right)\right) \cdot \|\nabla_\vomega \mathscr{L}\|/\|\vomega\|]
\end{equation}
Note that we have written this in the case of no momentum by using $-\eta \nabla_\vomega \mathscr{L}$ instead of $\vu$, but for the standard implementation of SGDM the momentum magnifies both this version of $\lambda_u$ and the standard ``weight decay'' ($\ell_2$-regularization) term the same way so they are comparable.
The $\lambda_u$ term can therefore be viewed as modifying the weight decay, the effective weight decay parameter is $\lambda_e = \lambda + \lambda_u$ and accounts for the entire radial portion of a weight update.
We replace the standard $\lambda$ with $\lambda_e$ when showing predicted values for the unnormalized network.

\begin{figure}[tb]
    \centering
    \includegraphics[width=\textwidth]{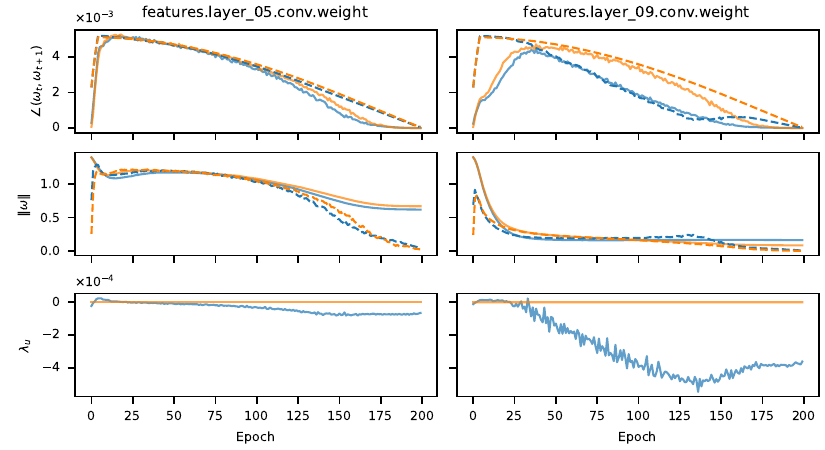}
    \caption{Measured (solid) and predicted equilibrium values (dashed) when training unnormalized (blue) and weight standardized (orange) variants of VGG-13 on CIFAR-10. The blue predictions account for the modified ``effective'' weight decay caused by the radial component of the gradient.}
    \label{appx:fig:vgg_dynamics}
\end{figure}

The results can be seen in \cref{appx:fig:vgg_dynamics} for two weights.
For the first one on the left, $\lambda_u$ is relatively small compared to $\lambda=5 \cdot 10^{-4}$ and the weight behaves similarly in both setups, with ``standard'' update dynamics in the unnormalized setup.
The equilibrium predictions match well early in training after the initial transient phase, but the weight falls out of equilibrium towards the end when it can't decay fast enough to keep up with the equilibrium weight magnitude.
For the second weight shown on the right, $\lambda_u$ is large causing a significant difference between the scale-invariant and scale-sensitive setups.
The modified equilibrium predictions using $\lambda_e$ capture the behavior well in the middle phase of training, after the initial transition before the weight falls out of equilibrium towards the end.
We note that in the unnormalized setup $\lambda_u$ changes over the course of training, starting out around 0 corresponding to an orthogonal gradient and growing larger in the later phases.
This is likely due to the cross-entropy loss used, which is minimized with large (infinite) output magnitudes once the network has learned to accurately classify (overfit) the training data.
This encourages the network to increase the output magnitude to fully overfit the data, resulting in a decreased effective weight decay and slower rotation.

Our results for VGG13 suggest that scale-sensitive weights can also have rotational dynamics in real neural networks.
The dynamics are less regular than in the normalized setup, with weights rotating at different speeds depending on the size of the radial gradient component.
The weight magnitude can also not vary freely like for scale-invariant weights, where we can trade off the weight decay and learning rate without affecting the dynamics much (once equilibrium is achieved).
Using large amounts of weight decay in unnormalized networks can bring the weight norms out of balance, resulting in issues like vanishing gradients or activations.
In unnormalized networks the magnitude of one weight matrix also affects the gradient magnitude of all others layers, further complicating the effect of weight decay.
Our rotational optimizer variants constrain the dynamics to match the equilibrium dynamics of weight standardized networks throughout training, eliminating some of these effects.

\section{Rotational Optimizer Wrapper}\label{appx:rotational_wrapper}

In this section, we provide further details on the algorithmic design choices used in our rotation optimizer wrapper, as shown in \cref{alg:rotational_wrapper}.
Note that the method can act as a wrapper around any given existing optimizer F with a known $\widehat{\eta_r}$. In cases where the true value is unknown or undesirable, we can also specify some different value of our choice.

\textbf{Rotational and Non-Rotational Updates:}
We use $\Omega$ to specify weights we apply rotational updates to, so a parameter $\vp$ is treated differently based on whether $\vp \in \Omega$ or not.
By default we consider each neuronal weight vector to be a separate vector in $\Omega$, but we could also apply the RVs at a coarser scale like whole layers (which is done in e.g.\ LARS).
The rotational wrapper leaves the update of non-rotational parameters unchanged.
Rotational parameters are rotated by $\widehat{\eta_r}$ on average and their magnitude is kept constant.
Some weights in $\Omega$ may be scale-sensitive meaning their magnitude can matter for efficient training.
Since the RVs constrain the weights we optionally introduce a learnable gain to allow the network to learn the right magnitude for these weights.
This gain can be absorbed into the weights for inference.

\textbf{Keeping the Weight Magnitude Constant:}
Alternatively, we could vary the weight magnitude according to our derived value for the equilibrium norm. However, with a learning rate schedule this value can become arbitrarily small causing numerical issues.
For scale-invariant weights the magnitude doesn't matter so we simply keep it constant. 
This has the added benefit of removing the inverse scaling effect of the weight norm on the gradient magnitude \cref{eq:grad_inv}, potentially making it a more informative metric over the course of training with a learning rate schedule.

\textbf{Controlling the Rotation Instead of the Relative Update:}
The rotation of a scale-invariant weight $\vomega$ is generally caused by both $\Delta_g \vomega$ and $\Delta_\lambda \vomega$ as can be seen in \cref{fig:smd_components}L~and~\ref{fig:smd_components}R.
In equilibrium, the sum of these components is roughly orthogonal to the weight vector.
We want to avoid having to apply the weight decay and our constrained magnitude is generally not equal to the equilibrium magnitude.
We therefore project $\Delta_g \vp$ to be orthogonal to $\vp$ and control the average size of this projected version of $\Delta_g \vp$ instead of the original $\vp$.
This lets us explicitly control the angular update, regardless of any radial component in $\Delta_g \vp$ that the weight decay would eliminate on average.
If we apply rotational updates to scale-sensitive weights, performing line~\ref{line:rms_update} after line~\ref{line:remove_projection} prevents any radial component in the gradient from affecting the rotational speed.

\textbf{Centering the Weights:}
Different normalization setups can result in slightly different SMD properties.
Layer Normalization typically makes an entire weight matrix scale-invariant whereas Batch Normalization makes individual filters (i.e.,\ rows or columns) independent.
The default form of the rotational wrapper corresponds to the rotational equilibrium dynamics obtained with Weight Standardization~\citep{qiao2019ws} also known as Centered Weight Normalization~\citep{huang2017centered}, where each filter is scale and shift invariant.
We remove the mean $\bar{\vp}=\frac{1}{C}\sum_{i=1}^{C} \evp_i$ of $\vp=[\evp_1,\ldots,\evp_C]$ since it is irrelevant in this setup.
This removal was also found to be beneficial in NF-Nets~\citep{brock2021signalpropagation,brock2021high}.
Note how we remove the mean component of the update before updating the RMS tracker.
This ensures that the average rotation is not decreased when there is a significant mean component in the update.

\textbf{Hyperparameters:}
The algorithm requires an $\epsilon$ value for numerical stability but otherwise only adds one hyperparameter, a decay factor $\beta$ similar to those in Adam.
It determines the rate at which we update our estimate of the average update magnitude (Line~\ref{line:rms_update}).
This in turn controls how much we let the rotation vary between steps.
We could potentially derive an analytical value for $\beta$ based on the convergence speed towards equilibrium.
For example $\beta$ should perhaps be roughly equal to $\sqrt{a}$ from \cref{eq:adamw_norm_convergence} for AdamW, when trying to match the dynamics exactly.
However, this rate may not be optimal and generally depends on the learning rate (which may vary according to a learning rate schedule).
We use a default of $\beta=0.9$ which should keep the expected angular update close to the equilibrium value over time, while still allowing some variation from step to step.
There is likely batch size dependence in the optimal value of $\beta$, with larger batches potentially benefiting from smaller values since balancing the average rotation within in each step could be sufficient in these cases.
An Adam-like bias correction is applied to the average update magnitude when it is used (Line~\ref{line:rotational_update}).

\textbf{Resource Requirements:}
We need to keep track of two scalars $\nu_\vp$ and $n_\vp$ for each rotational parameter.
Since $\vp$ is generally a vector, such as a row in a weight matrix, the memory requirement is negligible compared to quantities like momentum that store a scalar for every element of $\vp$.
The computational requirements in terms of floating-point operations are also relatively small, linear in the number of network (scalar) parameters like standard optimizers.
However, the rotational variants are not applied fully elementwise, making efficient (parallel) implementations slightly harder.

\section{Additional Experiments}\label{appx:sec:additional_experiments}

\textbf{Dynamics for a Random Walk:} In \cref{appx:simple_system} we measure the update dynamics in a simplified system undergoing a random walk matching our analytical setting. We observe that our analytical predictions from \cref{tab:equilibrium_summary} accurately describe the dynamics of this system for each of the optimizers we analyzed.

\begin{figure}[t]
    \centering
    \vspace{-6pt} %
    \includegraphics[width=0.9\textwidth]{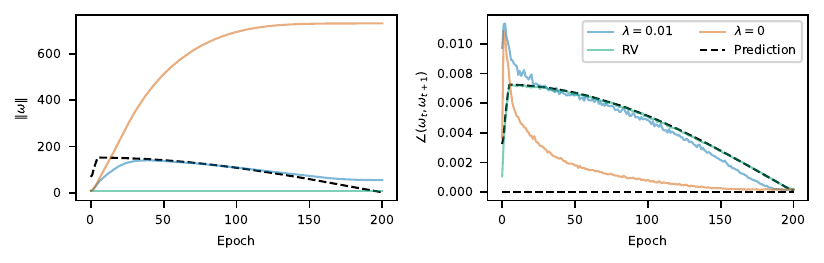}
    \vspace{-15pt} %
    \caption{Measurements of the weight norm $\|\vomega\|$ and angular updates $\eta_r$ for CIFAR10 ResNet20 training for AdamW and RV-AdamW compared to Adam without weight decay ($\lambda=0$). The black line represents our equilibrium and angular update size predictions from \cref{tab:equilibrium_summary}. Note how the angular updates without weight decay quickly approach zero, which would even happen with a constant learning rate schedule.}
    \label{appx:fig:adamw_vs_adam_dynamics}
\end{figure}

\textbf{Dynamics Without Weight Decay:}
In \cref{sec:experiments_measure_and_constrain}, we discussed that training without weight decay is similar to multiplying the learning rate schedule for $\eta_r$ with an exponentially decaying function.
This can be beneficial in some scenarios, e.g.\ \citet{loshchilov2018decoupled} found that for a constant learning rate schedule a weight decay of zero can be optimal, unlike when a cosine decay schedule is used.
The exponential decay in the angular updates without weight decay evident in \cref{appx:fig:adamw_vs_adam_dynamics}, where we present measurements of a ResNet-20 trained on CIFAR-10 with a learning rate of $\eta=0.05$ and no weight decay ($\lambda=0$), as well as the same setup with a weight decay of $\lambda=0.01$ for both AdamW and RV-Adam.
All experiments use a cosine learning rate schedule.
Note the equilibrium norm for RV-Adam is constant throughout training by design.

\begin{figure*}[tb]
    \centering
    \begin{subfigure}[b]{4.333in}
        \includegraphics[width=\textwidth]{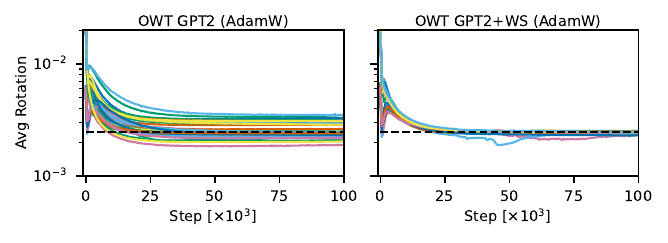}
    \end{subfigure}
    \begin{subfigure}[b]{2.166in}
        \includegraphics[width=\textwidth]{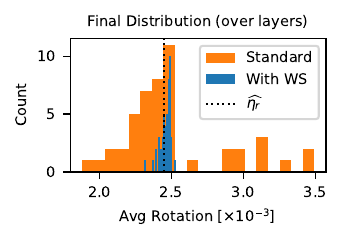}
    \end{subfigure}
    \caption{\textbf{Left:} The measured rotation of every linear layer of GPT2-124M, with and without Weight Standardization, for 100k steps on OpenWebText using a constant learning rate. The predicted equilibrium rotation is shown in dashed black (not adjusted for the scale-sensitivity). \textbf{Right:} The distribution of the average rotation of the layers at step 100k for each setting.}
    \label{fig:gpt_dynamics_scale_sensitive}
    \vspace{-10pt}
\end{figure*}

\textbf{Dynamics of Scale-Sensitive GPT Model:} In \S\ref{sec:experiments_measure_and_constrain} and \cref{fig:measurements_rn50_gpt2ws} we show the dynamics of a modified GPT2 model where Weight Standardization~(WS, \citet{qiao2019ws,huang2017centered}) is applied to all linear layers.
Although we don't show the results here, we found the Weight Standardized model to generally perform equally well or slightly better than the baseline.
WS makes individual neurons scale-invariant which in turn helps balance and regulate the dynamics as discussed in \S\ref{sec:experiments_balanced_rotation}.
Without scale-invariance, radial components in the gradient can modify the effective weight decay (see \S\ref{sec:scale_sensitive_dynamics}) leading to different equilibrium values for the magnitude and rotation.
We can predict modified values for SGDM (\cref{appx:scale_sensitive_smd}) based on measurements of the radial component, but have not attempted to do so for AdamW.

In \cref{fig:gpt_dynamics_scale_sensitive} we compare the angular updates of the Weight Standardized GPT2 model to a standard one without scale-invariance.
We show the predicted values based on the formula for scale-invariant layers.
As can be seen the unmodified GPT2 model has layers that deviate from this value although it is a reasonable approximation in most cases.
With Weight Standardization all layers end up very close to the predicted equilibrium value.
In fact we include the extra $\eta\lambda^2$ term from \cref{eq:adamw_equilibrium_norm} in the prediction here since it is significant compared to the width of the final distribution.
Two layers (in the first attention block) temporarily deviate slightly for reasons we are not fully sure of, but they coincide with significant changes in the gradient norms.
Effects such as dead neurons or a negative average alignment of successive gradients could also cause deviations from the random walk behavior our analysis assumes (see \cref{appx:real_vs_random_walk}).

\textbf{Constraining the Rotational Dynamics of Other Optimizers:}
In this section, we examine the performance of the Rotational Variants of SGDM and Lion.
\Cref{appx:tab:constraining_smd_sgdm_lion} shows that the RVs for both SGDM and Lion can match the baseline similar to what we observed for AdamW in \cref{tab:constraining_smd}.
In most cases no further hyperparameter tuning of the baseline values is needed (zero-shot) but otherwise light tuning (few-shot) suffices.
This shows that the RVs can replicate the benefits of weight decay while simplifying the training dynamics by removing the transient phase and fully balancing rotation.

\setlength{\tabcolsep}{4pt}
\begin{table}[bt]
  \centering
  \caption{Test set performance for baseline optimizers SGDM and Lion and their RVs. Zero-shot results for RVs use the baseline hyparparameters, the few-shot is lightly tuned.}
  \label{appx:tab:constraining_smd_sgdm_lion}
  \begin{tabular}{@{}lllll|llll@{}}
    \toprule
    \textbf{Dataset} & \textbf{Model} & \textbf{Optimizer} & \textbf{Batch Size\!\!} & \textbf{Metric ($\updownarrows$)} & \textbf{Baseline} & \textbf{RV Zero-Shot} & \textbf{RV Few-Shot} \\
    \midrule
    CIFAR-10 & ResNet-20 & SGD & 128 & Top-1 Acc. ($\uparrow$) & 92.7$\pm$0.1 & 92.4$\pm$0.2 & N/A \\
    CIFAR-10 & ResNet-20 & SGD & 2048 & Top-1 Acc. ($\uparrow$) & 92.0$\pm$0.2 & 92.0$\pm$0.3 & N/A \\
    CIFAR-10 & ResNet-20 & Lion & 128 & Top-1 Acc. ($\uparrow$) & 92.1$\pm$0.2 & 91.9$\pm$0.2 & N/A \\
    CIFAR-10 & ResNet-20 & Lion & 2048 & Top-1 Acc. ($\uparrow$) & 91.8$\pm$0.3 & 91.5$\pm$0.3  &  91.8$\pm$0.2 \\
    Imagenet-1k & ResNet-50 & SGD & 256 & Top-1 Acc. ($\uparrow$) & 77.4 & 77.3 & N/A \\
    \bottomrule
  \end{tabular}
\end{table}

\begin{figure}[t]
    \centering
    \vspace{-6pt} %
    \includegraphics[width=0.9\textwidth]{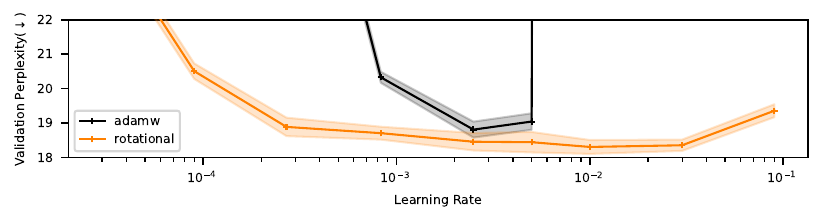}
    \vspace{-15pt} %
    \caption{
    Validation perplexity for GPT2-124M model on Wikitext for different learning rate, weight decay pairs with a constant product ($\eta\lambda=2.5\!\cdot\!10^{-3}$) resulting in a specific $\widehat{\eta_r}$ (\cref{tab:equilibrium_summary}).}
    \label{appx:fig:lr-wd-trade-off-llm}
\end{figure}

\textbf{Learning Rate vs Weight Decay for a Transformer model:}
Here we repeat the Learning Rate vs Weight Decay Experiment from \S\ref{sec:experiments_scheduling}, \cref{fig:lr-wd-trade-off}, for a GPT2-124M model trained on Wikitext.
We vary the learning rate $\eta$ while keeping the product of the weight decay $\lambda$ and learning rate, i.e.\ $\lambda\eta$ constant.
We include a learnable gain for each neuron when using the RV, as they are not scale-invariant and their magnitude may matter for learning.
The results are shown in \cref{appx:fig:lr-wd-trade-off-llm}.
Varying the learning rate while keeping the $\eta\lambda$ product constant only affects the updates of biases and gains in the RV as the rotational rate is constant.
For the baseline the relative size of the bias/gain updates compared to the angular updates is also affected but additional transient effects are also introduced.
We observe that the RV is less sensitive, displaying better results across a wide range of learning rates.
We believe this is primarily due to variations in the effective (rotational) step size schedule for the baseline, which can suffer from overly fast initial (transient) rotation at higher learning rates and an slow rotation for lower learning rates.

\begin{figure*}[tb]
    \centering
    \includegraphics[width=\textwidth]{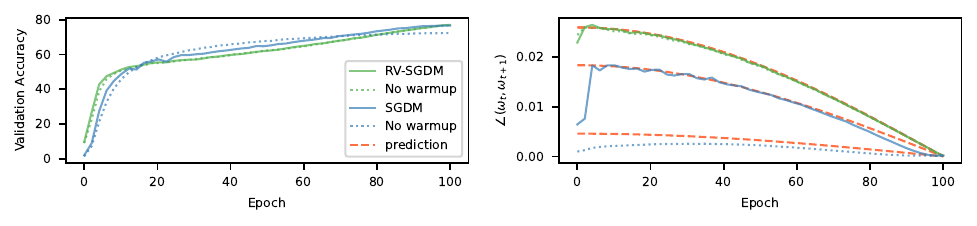}
    \caption{\textbf{Left:} Validation accuracy curves for SGDM and RV-SGDM with and without learning rate warmup for the best configuration of each based on \cref{fig:warmup}R. The final accuracy is 76.6\% for SGDM with warmup significantly higher than 72.3\% without warmup, compared to 77.1\% for RV-SGDM with warmup which is only slightly higher than 76.9\% without warmup. \textbf{Right:} Measured average angular updates averaged across all convolutional layers of the ResNet50. For both RV-SGDM runs, a learning rate of $\eta = 6.4$ was applied. For standard SGDM a learning rate of $\eta = 3.2$ compared to a significantly lower $\eta = 0.2$ was utilized for the SGDM without warmup. We believe the reduced learning rate for SGDM without warmup is necessary due to the instability encountered during the initial transient phase at higher learning rates.}
    \label{fig:resnet50_warmup_curves}
    \vspace{-10pt}
\end{figure*}

\textbf{Need for Learning Rate Warmup:}
In \cref{fig:warmup}R, we demonstrate that SGDM significantly benefits from a warmup period, while the RV-SGDM exhibits only marginal performance improvements.
For the learning rate that performed best in this experiment, we extended the run to span the full 100-epoch duration.
\Cref{fig:resnet50_warmup_curves} shows the validation accuracy curve over the course of training for each of these runs.
Notably, while SGDM without warmup exhibits significant performance gains over the course of the 100 epochs, it fails to achieve the performance of SGDM with warmup.
In contrast, training the ResNet50 with RV without a warmup period closely matches the performance with warmup.
This finding reinforces our belief that a learning rate warmup may aid in stabilizing the transient phase of training, an effect that could potentially be achieved by directly controlling the expected angular update.

\textbf{Balancing Adam+$\ell_2$:} Our analysis (\S\ref{sec:analysis_adam_vs_adamw}) shows that in Adam+$\ell_2$ the equilibrium rotation depends on the gradient norm unlike AdamW and the other optimizers listed in \cref{tab:equilibrium_summary}.
In our experiments (\S\ref{sec:experiments_balanced_rotation}) we found that this indeed holds for ResNet-18 on CIFAR-10, where we also reproduced a sweep from the original AdamW paper that shows a performance gap between the two.
Although we observe imbalanced rotation causing performance degradation in multiple settings, the experiment does not directly show that the imbalanced rotation causes the difference in this case.

The updates of two optimizer can only differ in two ways, in the magnitude of the update and/or the direction of the update.
We can construct a special RV that uses Adam+$\ell_2$ as the inner optimizer ($F$ in \cref{alg:rotational_wrapper}) but uses the rotational update size $\widehat{\eta}_r$ computed for AdamW.
If this RV performs similar to the standard RV-AdamW then the difference in the update direction should not be a significant factor.
Note however that the impact on the update direction depends on the weight norm (which is kept fixed at initialization in the RV but varies in standard training) as well as the strength of the $\lambda$ hyperparameter (which is much higher for the optimal AdamW configuration).
We expect this to result in a larger change in the direction of $\Delta_g \vp$ than would be observed along an Adam+$\ell_2$ optimization trajectory.
However, we do not validate this directly and expect the effect to vary over time and between network components (layers / neurons).

Running the two RVs at the optimal hyperparameter configuration from AdamW over 5 seeds gives 94.61$\pm$0.10\% test accuracy for RV-AdamW and 94.53$\pm$0.11\% for the special Adam+$\ell_2$ RV.
This is slightly lower than the 94.71$\pm$0.16\% for the baseline AdamW but around 0.5\% higher than the 94.08$\pm$0.16\% for Adam+$\ell_2$.
The RV-AdamW performance is likely slightly lower than AdamW due to rotational scheduling effects as is often the case for a zero-shot hyperparameter transfer.
The fact that the two RVs perform very similarly and noticeably better than the extensively tuned Adam+$\ell_2$ supports our conjecture that irregular rotation contributes to the lower performance of Adam+$\ell_2$.

\textbf{Imbalanced Rotation:} 
\begin{figure}[tb]
    \centering
    \vspace{-6pt} %
    \includegraphics[width=\columnwidth]{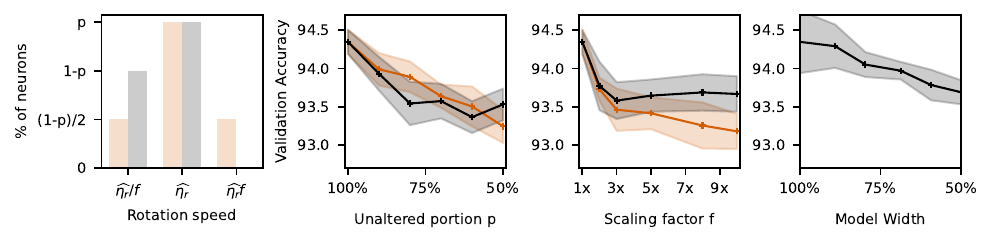}
    \vspace{-20pt} %
    \caption{
    The more imbalanced the rotation, the lower the final performance for ResNet-18 training on CIFAR-10. This happens even with extensive hyperparameter tuning of each setting, which does not compensate for the imbalance.
    \textbf{Left:}~Two artificially imbalanced angular update size distributions (black/orange). 
    A portion $1-p$ of the neurons is rotated $f$ times slower and/or faster than rest using a modified RV.
    \textbf{Right-1:}~Varying the portion $p$ for a fixed factor $f\!=\!10$.
    \textbf{Right-2:}~Varying the factor $f\!\in\![1, 10]$ for a fixed portion $p\!=\!50\%$.
    \textbf{Right-3:}~Reducing the network width unsurprisingly also degrades performance. Imbalanced rotation with $p\!=\!50\%$ and $f\!=\!3$ gives similar results as a network of half the width (93.7\%), suggesting we could be better off without neurons with even seemingly small deviations in the angular update size.
    }
    \label{fig:rotation_speed_control}
\end{figure}
In \cref{sec:experiments_balanced_rotation}, we explore the importance of balanced rotation in neural network training.
We find that training configurations that result in more balanced rotation perform better, both in the case of AdamW compared to Adam+$\ell_2$ and when comparing additional Weight Standardization to a baseline with only Layer Normalization.
However, it is possible that other differences between these setups also effect the performance.
To directly test the impact of imbalanced rotation, while eliminating as many confounding factors as possible, we construct a modified variant of RV-AdamW.
In this modified RV, we additionally scale the angular updates of some fraction $1-p$ of the neurons in each layer by a factor of $f$.
We consider two distributions shown in \cref{fig:rotation_speed_control}L, either rotating all affected neurons slower or rotating half of them faster and the other half of them slower.
For each configuration ($p$ and $f$) we tune the weight decay, which shifts the rotational distribution through $\eta_r$ while leaving the update size $\eta_g$ for biases and gains unaffected.

The results can be seen in \cref{fig:rotation_speed_control}R for a ResNet-18 trained on CIFAR-10.
We see that the performance degrades when increasing either the portion of affected neurons or the scaling factor for either artificially imbalanced rotational distribution.
The final panel shows how the previous imbalanced rotation compares with simply reducing the width of the network (completely removing neurons).
Interestingly, the second and third panels show that a scaling factor of around $3\times$ applied to $50\%$ of neurons results in performance comparable to a network with half the width.
This indicates that imbalanced rotation directly and significantly harms performance.
Why this happens is not entirely clear to us as we also discuss in \cref{appx:why_balanced_rotation}.
Neurons that rotate at a different speed may simply not learn effectively, which would work similarly to decreasing the number of neurons in the network.
Intuitively, slow moving neurons could also contribute to overfitting (which is more prominent with lower global learning rates).
We also speculate that fast rotating neurons may hinder the learning of other neurons, perhaps by limiting the maximum stable (global) learning rate or by changing the internal representations of the network too quickly.

\begin{figure}[tb]
  \begin{minipage}{0.41\textwidth}
    \resizebox{\linewidth}{!}{\begin{tabular}{@{}ll|ll@{}}
      \toprule
      \textbf{Dataset}     & \textbf{Model} & \textbf{Zero Mean} & \textbf{Invariance} \\
      \midrule
      \multirow{2}{*}{CIFAR-10} & \multirow{2}{*}{ResNet-18} & 94.5$\pm$0.4 (\textbf{y}n) & 94.5$\pm$0.4 (y\textbf{n})  \\
                                &  & 94.3$\pm$0.2 (\textbf{n}c) & 94.4$\pm$0.2 (y\textbf{t})\\
            \multirow{2}{*}{Wikitext} & \multirow{2}{*}{GPT2-124M} & 18.5 $\pm$0.42 (\textbf{y}c) & 18.5$\pm$0.42 (y\textbf{c})  \\
                                &  & 18.4 $\pm$ 0.4 (\textbf{n}c) & 18.4$\pm$0.15 (y\textbf{t})\\
      \bottomrule
    \end{tabular}}
  \end{minipage}
    \begin{minipage}{0.57\textwidth}
    \vspace{10pt}
    \includegraphics[width=0.9\textwidth]{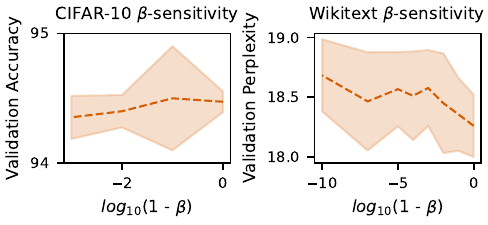}
  \end{minipage}
  \vspace{-1em}
  \caption{Experimental results for hyperparameter sensitivity. We report perplexity ($\downarrow$) on Wikitext validation dataset and top-1 Acc. ($\uparrow$) on a random validation split on CIFAR-10. In the default set up weight centering (y) and per neuron (n) scale-invariance is enabled.}
  \label{fig:hyperparameter_sensitivity}
\end{figure}

\textbf{Hyperparameter Sensitivity:}
The RVs introduce one significant hyperparameter, the decay rate $\beta$.
Further, we can decide whether to enable (y) or disable (n) centering of the weights (zero-mean) and whether to control the angular updates on the layer (l) or neuron (channel, c) level.
In this section we study the sensitivity of these choices in two different setups.
We train a ResNet-18 on a random train split of CIFAR-10 with the RV of SGDM and a GPT2-124M model on Wikitext with the RV of AdamW.
The results are shown in \cref{fig:hyperparameter_sensitivity}.
They indicate that the performance of the RVs remains relatively stable when the hyperparameters are varied.
Note that we tuned the effective update size for each configuration by varying the weight decay for the ResNet-18 experiments, but run the sensitivity study only for one learning rate, weight decay setting for the GPT2-124M model due to resource constraints.
Here we find little difference between the neuron and layer levels, but believe this difference may be larger in other settings that are more prone to imbalanced rotation on the neuron level.

\section{Experimental Details}\label{appx:experimental_details}
We perform our experiments on several popular datasets, i.e., CIFAR-10/100~\citep{Krizhevsky2009LearningML} and Imagenet-1k~\citep{imagenet15russakovsky} for image classification, IWSLT2014~\citep{cettolo14iwslt} for German-English translation, and Wikitext~\citep{merity2017pointer} and OpenWebText~\citep{radford2019language} for language modelling.
Our code utilizes the TIMM library~\citep{rw2019timm} for vision tasks, FairSeq~\citep{ott2019fairseq} for translation, and NanoGPT~\cite{nanoGPT} and LLM-Baselines~\cite{llm_baseline} for language modelling.

We train commonly used architectures, namely ResNet-20, ResNet-18, ResNet-50~\citep{he2016deep}, DeiT tiny~\citep{deit_tiny}, a small transformer, and a GPT2-124M network~\citep{radford2019language} from scratch.

\textbf{General Setup:}
For all experiments trained with SGDM we use a momentum of $0.9$, for experiments trained with AdamW we used $\beta_1 = 0.9$ and $\beta_2 = 0.999$ and for experiments trained with our RVs we used $\beta = 0.99$, unless otherwise stated.
For Lion we used $\beta_1 = 0.9$ and $\beta_2 = 0.999$ exclusively.
Most of the experiments are run on a single NVIDIA A100-SXM4-40GB GPU.

Note that we used default architectures for the baseline experiments, but used a learnable gain for DeiT trained on Imagenet-1k, Transformer-S trained on IWSLT2014 de-en and the GPT2-124M architecture on Wikitext and OpenWebText.
As mentioned in \cref{sec:rotational_optimizers} we do this since transformers are not scale-invariant so constraining the weight norms may be harmful.

Here we list additional details referenced in the \textbf{details}-column in \cref{appx:tab:constraining_smd} and \ref{appx:tab:hyperparameter_sensitivity}:
\begin{enumerate}[label=D-\arabic*,ref=D-\arabic*]
    \item \label{itm:1} We pre-process the data by normalizing it with mean $(0.4914 0.4822 0.4465)$ and std $(0.2023 0.1994 0.2010)$. For training we used simple data augmentation from~\citet{he2016deep}.

    \item \label{itm:2} We use the standard data augmentation from~\citet{he2016deep} for Imagenet.

    \item \label{itm:3} Analogously to~\citet{deit_tiny} we apply strong data augmentation. We use color jitter brightness up to $0.3$, auto-augmentation, random erase with probability $0.25$, drop path with probability $0.1$, mixup with probability $0.8$ and cutmix with probability $1.0$. Additionally, we use label smoothing of $0.1$.

    Note that in this experiment, we implemented a $\cos^2$ schedule for the RV few-shot training. We observed that the baseline training exhibited a smaller rotation towards the end. Consequently, we adjusted the effective update size for the RV training to more closely align with the baseline run's effective learning rate schedule.
    
    \item \label{itm:4} We use standard FairSeq library~\citep{ott2019fairseq} with dropout probability $0.3$.

    Note that additionally to using weight standardization with learnable gain, we set the weight decay to $0$ for scale-sensitive weights. This is done by default for the vision tasks in TIMM library~\citep{rw2019timm}, but not by FairSeq library~\citep{ott2019fairseq} we used for this task. We observed no difference in performance for the baseline model, yet this adjustment allowed to tune the effective update size for the scale-invariant weights, without affecting the learning rate of the scale-sensitive weights significantly.

    \item \label{itm:5} For this experiment we use the llm-baseline library ~\citep{llm_baseline}. For the GPT2-124M architecture, we use vocabulary size of $50304$, sequence length of $512$, embedding size of $768$. The model features $12$ repeated blocks, each comprising a self-attention block followed by a two-layer MLP block with hidden dimension $3072$. This results in a total of 124 million parameters. We use a drop probability of $0.2$ for dropout.

    \item \label{itm:6} For this experiment we use the nanoGPT library ~\citep{nanoGPT}. The details are the same as for \ref{itm:1}, except the sequence length is 1024 and no dropout is used. The model is trained for 5000 iterations which corresponds to roughly 20 tokens per parameter, inspired by Chinchilla~\citep{hoffmann2022training}.
    The learning rate schedule goes down to 0, which we found to be easier to combine with learning rate and weight decay sweeps without affecting the performance significantly in exploratory experiments.
\end{enumerate}

\textbf{Constraining the Rotational Dynamics:}\label{appx:constraining_smd}
The experimental details for the experiments reported in \cref{tab:constraining_smd,appx:tab:constraining_smd_sgdm_lion} can be found in \cref{appx:tab:constraining_smd}.

\textbf{Measuring the Update Dynamics \& Dynamics of Scale-Sensitive GPT Model:}
For the experiments in \cref{fig:measurements_rn50_gpt2ws}, we generally used the same settings for ResNet-50 on ImageNet-1k trained with SGDM and the GPT2-124M model trained with AdamW on OpenWebText. However, we use a constant learning rate without warmup, train for a 100k steps and use the default learning rate from NanoGPT, $6\mathrm{e}{-4}$ for the GPT2 model. We use Weight Standardization (with learnable gains) for GPT2 in \cref{fig:measurements_rn50_gpt2ws} but not \cref{fig:gpt_dynamics_scale_sensitive}.

\setlength{\tabcolsep}{4pt}
\begin{table}[tb]
  \centering
  \caption{Experimental set up (include training set and test set definition).}
  \label{appx:tab:constraining_smd}
  \resizebox{1.0\linewidth}{!}{\begin{tabular}{@{}llll|llllllll@{}}
    \toprule
     \multirow{2}{*}{\textbf{Dataset}} &  \multirow{2}{*}{\textbf{Model}} &  \multirow{2}{*}{\textbf{Optimizer}} &  \multirow{2}{*}{\textbf{Batch Size}} &  \multirow{2}{*}{\textbf{zero shot}} &  \multirow{2}{*}{\textbf{few shot}} &  \multirow{2}{*}{\textbf{details}} & \textbf{lr} & \multirow{2}{*}{\textbf{warmup}} & \textbf{epochs (e)/} & \textbf{train} & \multirow{2}{*}{\textbf{precision}} \\
     & & & & & &  & \textbf{schedule} & & \textbf{iteration (it)} & \textbf{duration} & \\
    \midrule
    \multirow{2}{*}{CIFAR-10} & \multirow{2}{*}{ResNet-20} & \multirow{2}{*}{SGD} & \multirow{2}{*}{128} & wd=$1\mathrm{e}{-4}$ & \multirow{2}{*}{N/A} & \multirow{2}{*}{(\ref{itm:1})} & \multirow{2}{*}{cosine} & lr=$1\mathrm{e}{-6}$ & \multirow{2}{*}{200 (e)} & \multirow{2}{*}{35min} & \multirow{2}{*}{float32} \\
    & & & & lr=0.5 & & & & 5 epochs & & & \\
    \midrule
    \multirow{2}{*}{CIFAR-10} & \multirow{2}{*}{ResNet-20} & \multirow{2}{*}{SGD} & \multirow{2}{*}{2048} & wd=$1\mathrm{e}{-4}$ & wd=$2\mathrm{e}{-4}$  & \multirow{2}{*}{(\ref{itm:1})} & \multirow{2}{*}{cosine} & lr=$1\mathrm{e}{-6}$  & \multirow{2}{*}{200 (e)} & \multirow{2}{*}{35min} & \multirow{2}{*}{float32} \\
    & & & & lr=4.8 & lr=4.8 & & & 5 epochs & & & \\
    \midrule
    \multirow{2}{*}{CIFAR-10} & \multirow{2}{*}{ResNet-20} & \multirow{2}{*}{AdamW} & \multirow{2}{*}{128} & wd=$1\mathrm{e}{-2}$ & \multirow{2}{*}{$\beta=0.9$} & \multirow{2}{*}{(\ref{itm:1})} & \multirow{2}{*}{cosine} & lr=$1\mathrm{e}{-6}$  & \multirow{2}{*}{200 (e)} & \multirow{2}{*}{35min} & \multirow{2}{*}{float32} \\
    & & & & lr=$5\mathrm{e}{-2}$ & & & &  5 epochs & & & \\
    \midrule
    \multirow{2}{*}{CIFAR-10} & \multirow{2}{*}{ResNet-20} & \multirow{2}{*}{AdamW} & \multirow{2}{*}{2048} & wd=$1\mathrm{e}{-2}$ & wd=$8\mathrm{e}{-2}$ & \multirow{2}{*}{(\ref{itm:1})} & \multirow{2}{*}{cosine} & lr=$1\mathrm{e}{-6}$ & \multirow{2}{*}{200 (e)} & \multirow{2}{*}{35min} & \multirow{2}{*}{float32} \\
    & & & & lr=$1.6\mathrm{e}{-1}$ & lr=$1.6\mathrm{e}{-1}$ & & &  5 epochs & & & \\
    \midrule
    \multirow{2}{*}{CIFAR-10} & \multirow{2}{*}{ResNet-20} & \multirow{2}{*}{Lion} & \multirow{2}{*}{128} & wd=1.0 & \multirow{2}{*}{N/A} & \multirow{2}{*}{(\ref{itm:1})} & \multirow{2}{*}{cosine} & lr=$1\mathrm{e}{-6}$  & \multirow{2}{*}{200 (e)} & \multirow{2}{*}{35min} & \multirow{2}{*}{float32}\\
    & & & & lr=$5\mathrm{e}{-4}$ & & & & 5 epochs & & & \\
    \midrule
    \multirow{2}{*}{CIFAR-10} & \multirow{2}{*}{ResNet-20} & \multirow{2}{*}{Lion} & \multirow{2}{*}{2048} & wd=1.0 & wd=2.0 & \multirow{2}{*}{(\ref{itm:1})} & \multirow{2}{*}{cosine} & lr=$1\mathrm{e}{-6}$  & \multirow{2}{*}{200 (e)} & \multirow{2}{*}{35min} & \multirow{2}{*}{float32} \\
    & & & & lr=$1.6\mathrm{e}{-2}$ & lr=$1.6\mathrm{e}{-2}$ & & &  5 epochs & & & \\
    \midrule
    \multirow{2}{*}{Imagenet-1k} & \multirow{2}{*}{ResNet-50} & \multirow{2}{*}{SGD} & \multirow{2}{*}{256} & wd=$1\mathrm{e}{-4}$ & \multirow{2}{*}{N/A} & \multirow{2}{*}{(\ref{itm:2})} & \multirow{2}{*}{cosine} & lr=$1\mathrm{e}{-6}$  & \multirow{2}{*}{90 (e)} & \multirow{2}{*}{30h} & \multirow{2}{*}{float16}  \\
    & & & & lr=$1\mathrm{e}{-1}$ & & & & 5 epochs & & &  \\
    \midrule
    \multirow{2}{*}{Imagenet-1k} & \multirow{2}{*}{DeiT tiny} & \multirow{2}{*}{AdamW} & \multirow{2}{*}{1024} & wd=$5\mathrm{e}{-2}$ & wd=$2\mathrm{e}{-1}$ & \multirow{2}{*}{(\ref{itm:3})} & \multirow{2}{*}{cosine} & lr=$1\mathrm{e}{-6}$  & \multirow{2}{*}{300 (e)} & \multirow{2}{*}{70h} & \multirow{2}{*}{float16}\\
    & & & & lr=$5\mathrm{e}{-4}$ & lr=$5\mathrm{e}{-4}$ & & & 5 epochs & & & \\
    \midrule
    \multirow{3}{*}{IWSLT2014 de-en\!\!} & \multirow{3}{*}{Transformer-S\!\!} & \multirow{3}{*}{AdamW} & \multirow{3}{*}{4096} & wd=$1\mathrm{e}{-4}$  & wd=$4\mathrm{e}{-1}$ & \multirow{3}{*}{(\ref{itm:4})} & \multirow{3}{*}{cosine} & \multirow{3}{*}{4000 (it)} & \multirow{3}{*}{22021 (it)} & 
    \multirow{3}{*}{50min} & \multirow{3}{*}{float16}  \\
    & & & & lr=$5\mathrm{e}{-4}$ & lr=$5\mathrm{e}{-4}$ & & & & & & \\
    & & & & $\beta_2=0.98$ & $\beta_2=0.98$ & & &  & & & \\
    \midrule
    \multirow{2}{*}{Wikitext} & \multirow{2}{*}{GPT2-124M} & \multirow{2}{*}{AdamW} & \multirow{2}{*}{55 $\times$ 3} & wd=$0.5$ & \multirow{3}{*}{N/A} & \multirow{3}{*}{(\ref{itm:5})} & cosine & \multirow{3}{*}{$2\mathrm{e}{-2}$\,(\%)} & \multirow{3}{*}{15000 (it)} & \multirow{3}{*}{3h} & \multirow{3}{*}{bfloat16}\\
    & & & & lr=$5\mathrm{e}{-3}$ & & & div\_f=$1\mathrm{e}{2}$ & & & & \\
    & & & & $\beta_2=0.95$ & & & final\_div\_f=$1\mathrm{e}{4}$ & & & & \\
    \midrule
    \multirow{2}{*}{OpenWebText} & \multirow{2}{*}{GPT2-124M} & \multirow{2}{*}{AdamW} & \multirow{2}{*}{12 $\times$ 40} & wd=$0.1$ & \multirow{3}{*}{N/A} & \multirow{3}{*}{(\ref{itm:6})} & cosine & \multirow{3}{*}{250 (it)} & \multirow{3}{*}{5000 (it)} & \multirow{3}{*}{4h} & \multirow{3}{*}{bfloat16}\\
    & & & & lr=$4.8\mathrm{e}{-3}$ & & &  min\_lr=$0$ & & & & \\
    & & & & $\beta_2=0.95$ & & &  & & & & \\
    \bottomrule
  \end{tabular}}
\end{table}

\textbf{Learning Rate vs Weight Decay:}
For the ResNet-20 experiment on CIFAR-10 and language model task on Wikitext with a GPT2-124M model we use the few shot (zero shot) setting reported in \cref{appx:tab:constraining_smd} as default. We then sweep over the learning rate keeping $\eta\lambda=5\!\cdot\!10^{-4}$,  $\eta\lambda=2.5\!\cdot\!10^{-3}$ respectively, constant.

\textbf{Transient Effects:}
In the experiment described in the preceding paragraph, we monitored $\|\vomega\|$, $\angle(\vomega_{t},\vomega_{t+1})$ during training for one of the runs with the chosen learning rates: $1\!\cdot\!10^{-4}$, $8.3\!\cdot\!10^{-3}$, and $3\!\cdot\!10^{-1}$.

\textbf{Need for Learning Rate Warmup:} For the ResNet-50 experiment on ImageNet-1k we follow the same base setup as we report in \cref{appx:tab:constraining_smd}.
We train for a total of 10 epochs using a cosine decay schedule (applied stepwise) and no warmup.
We use $32$ local accumulations on top of batches of size 256 to emulate a batch size of 8'192 and sweep the learning $\lambda$ in the range $2^i \cdot 0.1$, with $i \in [0, \ldots, 9]$.
The GPT2 experiments follow the setup in \cref{appx:tab:constraining_smd} with the learning rate swept over $\{6 \cdot 10^{-4} \cdot 2^{i}\}$ for $i \in [-1, \ldots, 5]$. %

\textbf{Adam vs AdamW:} For the sweep we train a ResNet-18 on a 90/10 train/val split from the original train set.
We use a step-wise cosine schedule and train for 200 epochs without warmup.
In this sweep we try to reproduce the results in figure~2 from \cite{loshchilov2018decoupled}, albeit with a slightly different network and training for 200 epochs instead of 100.
The best configuration for Adam was $\eta=7.813\cdot 10^{-4}$, $\lambda=1.250\cdot 10^{-4}$ resulting in a validation set accuracy of $93.919\%$.
The best configuration for AdamW was $\eta=1.25 \cdot 10^{-2}$, $\lambda=8.0\cdot 10^{-2}$ with a validation accuracy of $94.319\%$.
On the test set we run each configuration over 5 different seeds, obtaining test accuracies of ${94.08 \pm 0.16\%}$ for Adam+$\ell_2$ and ${94.74 \pm 0.14\%}$ for AdamW.

\textbf{Benefit of Weight Standardization:}
We train a ResNet-18 with layer normalization and weight standardization on top of layer normalization on CIFAR-100 using the same augmentation, learning rate schedule and base hyperparameters as for the ResNet-20 on CIFAR-10 experiments, unless otherwise noted below.
We train on a random subset containing 90\% of the train set and use the remaining 10\% for validation which we report.
The inputs are normalized for mean \verb|(0.5071, 0.4867, 0.4408)| and std \verb|(0.2675, 0.2565, 0.2761)|.
We use a weight decay of $5 \cdot 10^{-4}$ and a batch size of 256.
The layer normalization is implemented with a Group Normalization~\citep{wu2018group} using a single group.

\textbf{Imbalanced Rotation:} We trained a ResNet-18 on a 90/10 train/val split from the original
train set with learning rate $\eta = 0.1$ , batch size 256 and varying weight decay to tune the effective updated size.
For each rotation speed $f$, portion $p$ and network width scaling we report the average performance over the best configuration of this sweep.
All other settings are equivalent to the settings reported in \cref{appx:tab:constraining_smd}.

\textbf{Hyperparameter Sensitivity:}
The experimental details for the experiments reported in \cref{fig:hyperparameter_sensitivity} can be found in  \cref{appx:tab:hyperparameter_sensitivity}. Note, that we don't specify the weight decay used for the ResNet-18 experiment, since we tuned the weight decay for each configuration with $2^i \cdot \lambda, i \in [-64, 32]$. We use a batch size of 256 and a batch size of 55 with 3 accumulation steps for the GPT2-124M model.

\setlength{\tabcolsep}{4pt}
\begin{table}[tb]
  \centering
  \caption{Experimental details for hyperparameter senstivity study.}
  \label{appx:tab:hyperparameter_sensitivity}
  \resizebox{1.0\linewidth}{!}{\begin{tabular}{@{}lll|lllllllll@{}}
    \toprule
    \multirow{2}{*}{\textbf{Dataset}} & \multirow{2}{*}{\textbf{Model}} & \multirow{2}{*}{\textbf{Optimizer}} & \textbf{training} & \textbf{validation}  & \textbf{hyper-} &  \multirow{2}{*}{\textbf{details}} &\textbf{lr} & \multirow{2}{*}{\textbf{warmup}} & \textbf{epochs (e)/} & \textbf{train} & \multirow{2}{*}{\textbf{precision}}  \\
    & & & \textbf{dataset} & \textbf{dataset} & \textbf{parameters} & & \textbf{schedule} &  & \textbf{iterations (it)} & \textbf{duration} & \\
    \midrule
    \multirow{2}{*}{CIFAR-10} & \multirow{2}{*}{ResNet-18} & \multirow{2}{*}{SGD} & \multirow{2}{*}{90\% Train} & \multirow{2}{*}{10\% Train} & wd=N/A & \multirow{2}{*}{(\ref{itm:1})} & \multirow{2}{*}{cosine} & lr=$1\mathrm{e}{-6}$ & \multirow{2}{*}{200 (e)} & \multirow{2}{*}{35min} & \multirow{2}{*}{float32}\\
     & & & & & lr=$1.0$ & & & 5 epochs & & \\
    \midrule
    \multirow{3}{*}{Wikitext} & \multirow{3}{*}{GPT2-124M} & \multirow{3}{*}{AdamW} & \multirow{3}{*}{Train} & \multirow{3}{*}{Validation} & wd=$0.5$ & \multirow{3}{*}{(\ref{itm:5})} & cosine & \multirow{3}{*}{$2\mathrm{e}{-2}$\,(\%)} & \multirow{3}{*}{15000 (it)} & \multirow{3}{*}{3h} & \multirow{3}{*}{bfloat16}\\
    & & & & & lr=$4\mathrm{e}{-3}$ & & div\_f=$1\mathrm{e}{2}$ & & & & \\
    & & & & & $\beta_2=0.95$ & & final\_div\_f=$1\mathrm{e}{4}$ & & & & \\
    \bottomrule
  \end{tabular}}
\end{table}

\newpage
\section{Update Size vs Learning Rate}\label{appx:update_size_vs_lr}
An update size, such as the average angular update $\eta_r$ or RMS update $\eta_g$, specifically refers to a measure of the size of individual updates.
This does not necessarily correlate directly to the net change over a longer period of multiple steps, e.g.\ an epoch of training.
The longer term change is determined by the shape of the optimization trajectory which depends on the size and direction of the individual steps.
For a fixed update size, a more consistent direction in the updates will cause a larger net change over a time period.
For this reason we prefer to use the term ``effective update size'' rather than ``effective learning rate'' which is sometimes used to refer to measures of long term changes.
We note that the update sizes are easier to measure and control, although the long term changes may be more informative for hyperparameter tuning and adjustment.

The difference is particularly important when using momentum which has a somewhat unintuitive effect in the random walk setting, e.g.\ for SGDM and AdamW.
As seen in \cref{tab:equilibrium_summary} higher momentum coefficients  decrease the average rotation in each step but we know they will also cause additional correlation between successive steps.
This can be viewed as smoothing the optimization trajectory, it will have smaller random fluctuations in each step but the averaged ``velocity'' is not necessarily smaller.
Comparing the average rotation per step between optimizers with different amounts of momentum will therefore not necessarily correlate with measures of how fast parameters are being updated over longer time intervals.

In the random walk setting the long term change is likely to be proportional to the total update contribution $\|\vu\|$ rather than the size of a single update step $\|\Delta_g \vp\|$ (see definitions in \S\ref{sec:equilibrium_geometric_model}).
Analogously, the total rotational contribution from a single gradient sample would be roughly $\|\vu\|/\inlinevomegaeqnorm$ rather then $\|\Delta_g \vp\|/\inlinevomegaeqnorm$ used in \S\ref{sec:adamw_equilibrium}.
These total update contributions may then add up orthogonally assuming successive gradients behave like in the random walk (i.e.\ that they are independent and zero mean in expectation).
We refer to this longer term speed as \textbf{diffusion rates} based on a loose analogy with random walks like Brownian Motion.
\Cref{tab:equilibrium_diffusion} shows the modified rates for SGDM and AdamW.
The diffusion rates are more consistent with the way learning rate is typically scaled with the momentum for SGDM, i.e.\ keeping $\eta/(1-\alpha)$ constant rather than $\eta/(1+\alpha)$, see e.g.\ \citet{chiley2019online} and \citet{fu2023momentum}.

\begin{table*}[tb]
\centering
\caption{
An extension of \cref{tab:equilibrium_summary} showing values for diffusion rates based on the expected Total Update Contribution (TUC) of the gradients from a single timestep, corresponding to the update sizes. The TUC attempts to capture the rate of longer-term changes rather than the instantaneous update sizes, accounting for the increased alignment from momentum.
}
\label{tab:equilibrium_diffusion}
\begin{tabular}{r|c|c|c}
\toprule
Measure & Definition & SGDM \cref{eq:sgdm_update} & AdamW \cref{eq:adamw_update} \\
\midrule
Equilibrium norm $\widehat{\|\vomega\|}$ & $\sqrt{\E[\|\vomega\|^2]}$ & $\sqrt[4]{\frac{\eta \E[\|\tilde{\vg}\|^2]}{2\lambda \cdot (1-\alpha)}}$ & $\sqrt{\frac{\eta C}{2\lambda}}$ \\
\midrule
RMS update size $\widehat{\eta_g}$ & $\sqrt{\E[\|\Delta_g \vp\|^2]}$ & $\eta \sqrt{\frac{\E[\|\vg\|^2]}{1-\alpha^2}}$ & $\eta \sqrt{C\frac{1-\beta_1}{1+\beta_1}}$ \\
\midrule
RMS diffusion rate $\widehat{\tau_g}$ & $\sqrt{\E[\|\vu\|^2]}$ & $\frac{\eta}{{1-\alpha}} \sqrt{\E[\|\vg\|^2]}$ & $\eta \sqrt{C}$ \\
\midrule
Expected rotation $\widehat{\eta_r}$ & ${\sqrt{\E[\|\Delta_g \vp\|^2]}}/{\widehat{\|\vomega\|}}$ & $\sqrt{\frac{2 \eta \lambda}{1 + \alpha}}$ & $\sqrt{2 \eta \lambda \frac{1-\beta_1}{1 + \beta_1}}$ \\
\midrule
Rotational diffusion rate $\widehat{\tau_r}$ & ${\sqrt{\E[\|\vu\|^2]}}/{\widehat{\|\vomega\|}}$ & $\sqrt{\frac{2 \eta \lambda}{1 - \alpha}}$ & $\sqrt{2\eta\lambda}$ \\
\bottomrule
\end{tabular}
\end{table*}

\newpage
\section{Why Balanced Rotation Might Work}\label{appx:why_balanced_rotation}
We empirically observe that balanced rotation seems to perform better than the imbalanced rotation resulting from various methods.
The reasons for this are not fully clear to us, but we believe that balanced rotation is a heuristic that ensures that all neurons and layers are updated at a reasonable speed.
For example, we intuitively expect that a layer which is updated very slowly, perhaps barely changing through the training process, will unlikely contribute optimally to the final model resulting in worse performance and wasted compute.
Conversely, a rapidly changing layer may cause instability, limiting the maximum stable learning rate and preventing other layers from learning effectively.
This suggests that sufficiently (and adversarially) imbalanced rates are not optimal.
At the same time it seems unlikely that exactly balanced rates are optimal in all settings, exhaustively tuning the rotation speed of each component would likely result in some specific optimal, imbalanced, rotation for a given problem.

Modern neural networks have a complex structure which can give rise to various effects that scale the gradient of one component relative to another in arbitrary ways.
\citet{neyshabur2015path} gives an example of this for ReLU networks (without branches), where one layer can be scaled up by a positive factor and another down.
This preserves the encoded function but will change the optimization trajectory and can significantly degrade performance.
Scale-invariance is another example, where scaling the weights of a layer will not affect the network outputs but changes the angular updates and therefore the optimization trajectory for standard optimizers.
Finally we can change the network architecture by e.g.\ inserting a fixed scaling factor into the computational graph that scales a given parameter (whose value is adjusted to compensate) which in turns scales the gradients changing the optimization trajectory.

These examples show that the relative gradient magnitude of one component compared to another may simply not be very meaningful quantity in neural networks.
Perhaps higher order information like curvature could allow us to determine what the relative update size should be.
Without this information, normalizing the update magnitude through methods like Adam, Sign-SGD or Rotational Optimizer Variants may act as heuristics that are better than using the arbitrary gradient magnitude.
Proving this might be infeasible, these heuristics may only help in specific cases dependent on the network architecture, initialization, dataset and so on, or it may at least be possible to construct adversarial scenarios where they hinder training.
We do not pursue this further here and simply treat balanced rotation as an empirically useful heuristic that seems to play a role in the effectiveness of many influential methods in deep learning.

\end{document}